\documentclass[lettersize,journal]{IEEEtran}
\usepackage{amsmath,amsfonts}
\usepackage{algorithmic}
\usepackage{algorithm}
\usepackage{array}
\usepackage[caption=false,font=normalsize,labelfont=sf,textfont=sf]{subfig}
\usepackage{textcomp}
\usepackage{stfloats}
\usepackage{url}
\usepackage{verbatim}
\usepackage{graphicx}
\usepackage{cite}
\usepackage{subfig}
\usepackage{hyperref}
\usepackage{cleveref}
\usepackage{tabularx}
\usepackage{enumitem}
\usepackage{multirow}
\usepackage{longtable}
\usepackage{multicol}
\usepackage{tabularray}
\usepackage{amsthm}
\newtheorem{definition}{Definition}

\graphicspath{ {./images/} }

\begin{document}

\title{
Analyzing State of Operators and the Impact of AI-Enhanced Decision Support in Control Rooms: A Human-in-the-Loop Specialized Reinforcement Learning Framework
}

\author{{Ammar N. Abbas*, Chidera W. Amazu, Joseph Mietkiewicz, Houda Briwa, Andres Alonzo Perez, Gabriele Baldissone, Micaela Demichela, Georgios G. Chasparis, John D. Kelleher, and Maria Chiara Leva \\ 

    \textit{Collaborative Intelligence for Safety-Critical systems (CISC) \\ \href{https://www.ciscproject.eu/}{ciscproject.eu}}}

    \thanks{*Corresponding author: ammar.abbas@scch.at}
    \thanks{This publication is the result of the research done along the Collaborative Intelligence for Safety-Critical systems (CISC) project that has received funding from the European Union’s Horizon 2020 Research and Innovation Programme under the Marie Skłodowska-Curie grant agreement no. 955901.}
    }

\markboth{Journal of \LaTeX\ Class Files,~Vol.~XX, No.~X, December~2023}%
{Shell \MakeLowercase{\textit{Abbas, Ammar N. et al.}}: AI-Based Decision Support System for Control Rooms}


\maketitle

\begin{abstract}
In complex industrial and chemical process control rooms, effective decision-making is crucial for safety and efficiency. The experiments in this paper evaluate the impact and applications of an AI-based decision support system integrated into an improved human-machine interface, using dynamic influence diagrams, a hidden Markov model, and deep reinforcement learning. The enhanced support system aims to reduce operator workload, improve situational awareness, and provide different intervention strategies to the operator adapted to the current state of both the system and human performance. Such a system can be particularly useful in cases of information overload when many alarms and inputs are presented all within the same time window, or for junior operators during training. A comprehensive cross-data analysis was conducted, involving 47 participants and a diverse range of data sources such as smartwatch metrics, eye-tracking data, process logs, and responses from questionnaires. The results indicate interesting insights regarding the effectiveness of the approach in aiding decision-making, decreasing perceived workload, and increasing situational awareness for the scenarios considered. Additionally, the results provide valuable insights to compare differences between styles of information gathering when using the system by individual participants. These findings are particularly relevant when predicting the overall performance of the individual participant and their capacity to successfully handle a plant upset and the alarms connected to it using process and human-machine interaction logs in real-time. These predictions enable the development of more effective intervention strategies.
\end{abstract}


\section{Introduction}
\IEEEPARstart{I}{n} today's complex industrial setting and chemical process control rooms, operators frequently encounter complex situations demanding rapid and precise decision-making. The Human-Machine Interface (HMI) can overwhelm operators with excessive information, leading to information overload and potentially compromising their ability to respond effectively, thus increasing the likelihood of human errors. To address this challenge, there is a need for a decision support framework to assist operators in detecting and responding to potential safety incidents. In this context, we present the results of an experimental study in this paper to assess the effectiveness of an improved AI-based recommendation system in addressing information overload and mitigating process abnormalities.

\begin{figure}[!t]
    \centering
        \includegraphics[width=\linewidth]{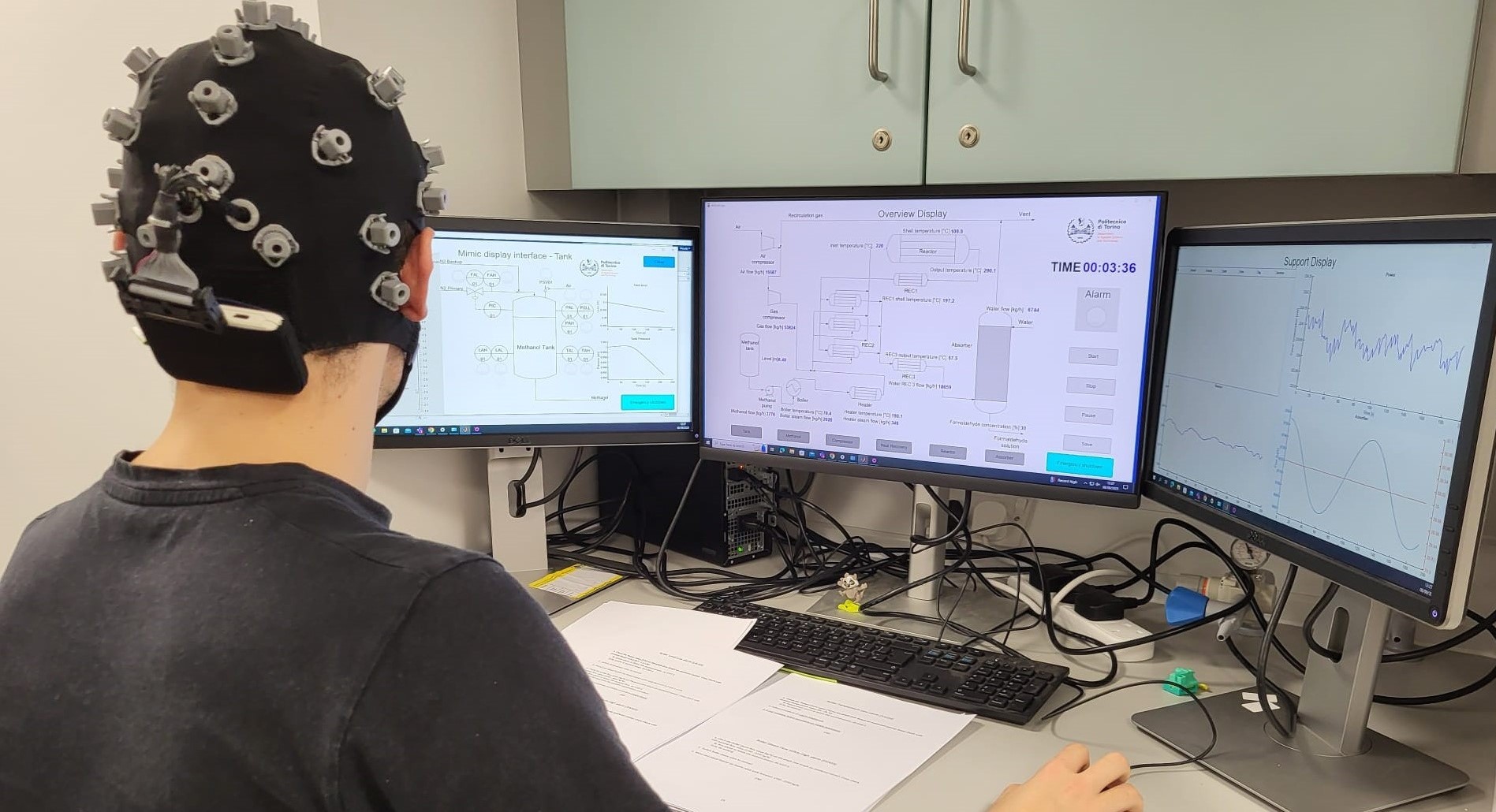}
        \caption{Case study: simulated control room environment of a process industry for a living lab.}
        \label{fig: participant}
\end{figure}

One of the objectives of the experiment is to ascertain whether the integration of the enhanced HMI that incorporates screen-based procedures along with an AI-based recommendation system can reduce operators' workload and enhance their situational awareness. This recommendation system utilizes a dynamic influence diagram in combination with reinforcement learning. The system is designed to identify anomalies and provide operators with not only information about the deviation but also specific procedures to follow, which are dynamically updated to reflect the current state of the system. To test this framework, we employ a simulation of formaldehyde production and present various deviation scenarios to participants. These participants are tasked with responding to these scenarios, with or without the assistance of the recommendation system. The aim is to compare the operator groups that have the aid of the AI-enhanced recommendation system with the ones that do not. The effectiveness of the recommendation system is then assessed based on its ability to aid participants in handling different scenarios. To assess the system we use various data sources involving quantitative, qualitative, and physiological measurements such as questionnaires, an eye tracker, a smartwatch, and simulator logs as shown in \cref{fig: participant}. Initial findings indicate the potential of an enhanced AI-based recommendation system to support operators in decision-making during challenging situations and to improve safety in industrial processes.

The presentation of procedures to operators is a critical aspect of this research. Providing operators with appropriately tailored and simplified procedures that adapt to the current state of the system has shown promising results in assisting their work. This system's interpretability and basis in expert knowledge are enhanced through the collaboration between the influence diagram and reinforcement learning. Such a system holds particular advantages for operators dealing with information overload and junior operators. Furthermore, to gain insights into human behavior and their intervention approach the analysis is performed within the participants of the group that is provided with the AI-enhanced decision support system. The cross-analysis compares the perception level and intervention ability of the participant based on the source they target to gain information about the system such as the screen procedures, AI-based decision support, or both.

Furthermore, the outcomes from these experiments and a survey \cite{amazu2023decision}, which involved experts and stakeholders who might engage with the developed support system led to the development of an extension of the dynamic influence diagram-based reinforcement learning framework. The focus is on proposing a Human-in-the-Loop (HITL) hierarchical framework for interpretable, specialized, and safe Deep Reinforcement Learning (DRL) that can be employed in real-world safety-critical industries. This extended framework aims to predict the plant's state during process abnormalities from human-machine interaction and process logs, enabling improved intervention strategies. Therefore, it can be used as a decision support tool during the training of junior operators, but also as a co-pilot to support even experienced operators in situations where there are too many alarms all arriving in the same time window. The tool can further also take up the role of taking over control and requesting approval to automate the recovery process. 

The paper is structured as follows: In \cref{sec:rel_lit}, we delve into the related work, outlining the contributions of this paper. Moving on to \cref{sec:frame}, we define the framework used for the AI-enhanced decision support system in the experiments and its extended version for future enhancements. In \cref{sec:case_study}, we present the formulated case study, which serves as the basis for data collection and analysis. The data collection process and details of the collected data are discussed in \cref{sec:data_coll}. Subsequently, \cref{sec:eda} conducts an in-depth analysis of the data, making comparisons between groups and within participants for the AI-supported group. Finally, in \cref{sec:hmm_pred}, we validate the extended framework using results from the experiments and elucidate its application capabilities.

\section{Related Work and Contributions}
\label{sec:rel_lit}
The review article by \cite{sethu2023application} emphasizes AI's potential to assist operators in making precise and swift decisions, thereby enhancing the safety of nuclear energy production. They also investigate various causes of human errors in nuclear power plants and assess how AI has been integrated into various operator support systems to address these errors. Eight specific types of support systems are examined, including decision support, sensor fault detection, operation validation, operator monitoring, autonomous control, predictive maintenance, automated text analysis, and safety assessment systems. Highlighting the significance of human-autonomous system interactions in ensuring plant system performance and reliability, the review addresses various human factors-related issues identified in the literature. The authors argue that a crucial gap exists in integrating the Human-in-the-Loop (HITL) strategy with both black-box models, such as Deep Neural Networks (DNN), and the white-box approach involving probabilistic modeling. 

The authors in \cite{lee2007development} also discuss the importance of preventing human errors in nuclear power plants. They propose an enhanced control room interface design and decision support system to enhance operational performance. Their methodology involves analyzing operators' cognitive activities, resulting in the development of two decision support systems for fault diagnosis and operation validation. 

The authors in \cite{kang2022concept} propose a framework, designed to improve initial emergency responses in nuclear power plants. The framework aims for agile, dynamic, and intuitive operation, seeking to reduce response time and operator workload through automation and real-time risk assessment. Scenario tests demonstrate a 95\% reduction in tasks and improved efficiency. Despite these positive results, the authors acknowledge limitations such as the reliance on rule-based logic, the importance of rigorous effectiveness verification, and the system's narrow focus on early-stage emergencies. As the system is framed on rule-based logic, therefore, it does not take into account the uncertainties involved in the environment and needs to reiterate the rules over time to adapt to the changing environment. 

The experiments presented in \cite{Hsieh2012} focus on a decision support system designed for identifying abnormal operating procedures in nuclear power plants. The study involved 32 graduate students with backgrounds similar to new nuclear power plant operators. After undergoing training and qualification tests, participants utilized the support system in a formal experiment. The results showed that the support system significantly reduced errors by 25\%, decreased decision-making time by 25\%, and increased decision accuracy by 18\%. Operators using the system made fewer erroneous decisions, experienced reduced mental workload, and demonstrated a preference for the support system. The study also highlighted the importance of avoiding information overload to maintain decision quality. As a recommendation, the developed decision support system is suggested as a valuable training tool, offering enhanced performance and reducing mental burden for operators. 

An experimental study was performed in \cite{balaji2023act}, where the authors discussed the challenges faced by human operators in process industries and explored the potential of using a digital twin to enhance their performance. The authors introduced a cognitive architecture using eye tracker data that can be used to create a human digital twin and explained how it could be applied to process industries. They evaluated its performance by comparing it to control room operators in a disturbance rejection task, involving 11 participants. The authors collected process data, operator actions, and eye-tracking data. The results showed that the human digital twin's performance was generally in agreement with that of human operators. The digital twin successfully diagnosed the cause of the abnormality and initiated necessary control actions. It focused on areas directly related to the disturbance and employed a proactive monitoring strategy using the trend of the process variables. 

\subsection*{Contributions}
This paper contributes by conducting an exploratory data analysis on experiments carried out within a simulated control room environment. The specific focus is on comparing two groups: one utilizing an AI-enhanced decision support system based on dynamic influence diagrams and deep reinforcement learning, and the other not using such a system. Additionally, we delve into the analysis of human decision-making preferences for the group using the AI-enhanced decision support system and their associated consequences. Moreover, Our paper addresses a significant gap in the existing literature by investigating the synergies among various models and the impact of a Human-in-the-Loop (HITL) AI-enhanced decision support framework, specifically integrating dynamic influence diagrams, a hidden Markov model, and Deep Reinforcement Learning (DRL). We leverage DRL for its adept handling of uncertainties and adaptability to dynamic environments. The objective is to incorporate human states and actions to enhance decision-making interventions. In contrast to prior research, our proposed framework uniquely emphasizes automated intervention strategies and decision-support controls during process abnormalities. This automation is triggered when the system identifies increased task loads for humans or predicts a high probability of human failure during abnormal situations.  While our current research paper does not explicitly delve into the creation of a human digital twin, it lays the groundwork for such advancements. The results presented here serve as a foundation for further enhancements in the framework.

\section{Framework}
\label{sec:frame}

    \subsection{Preliminaries}
    The AI-based recommendation system leverages Deep Reinforcement Learning (DRL) and a dynamic influence diagram. DRL is employed in an online setting, where it learns through interaction with the environment and observation of process behavior based on its actions. The dynamic influence diagram is constructed using expert knowledge, which includes the physical equations governing the behavior of the system components. These two models are combined to detect deviations in the process and provide the operator with precise recommendations, along with specific values tailored to the current circumstances.

        \vspace{0.1in}
        \subsubsection{Dynamic Influence Diagram (DID)}
        A dynamic influence diagram (DID) \cite{howard2005influence, tatman1990dynamic} is a graphical decision analysis tool that extends traditional influence diagrams by incorporating a time dimension. DIDs represent decision problems through nodes such as decision, chance, and value nodes, connected by directed arcs that illustrate causal relationships and dependencies. The inclusion of a time element allows the modeling of evolving systems. Key components include utility functions, decision rules, and the application of probabilistic links to account for uncertainty. DIDs enable scenario and sensitivity analysis, offering insights into decision-makers' strategies and the robustness of outcomes. The use of a dynamic influence diagram within this framework and experiment is described in detail by authors in \cite{mietkiewicz2023dynamic, mietkiewicz2024enhancing}. The definition of an influence diagram is as follows:

        \begin{definition}{Discrete Limited Memory Influence Diagram\cite{kjaerulff2013bayesian}} \label{def:DLMI}
            Given an influence diagram denoted as $N=(X, G, P, U)$, it comprises the following components:
            \begin{enumerate}[label=(\roman*)]
                \item A Directed Acyclic Graph (DAG) $G=(V, E)$, where $V$ represents the set of nodes and $E$ is the set of directed edges, indicating dependency relations and information flow.
                \item A collection of discrete random variables $X_C$ and decision variables $X_D$, such that the total set of variables $X$ is the union of these two sets, i.e., $X = X_D \cup X_C$. These variables are represented by the nodes in $G$.
                \item A set of conditional probability distributions $P$, where each distribution $P(X_v | X_{pa(v)})$ is associated with a discrete random variable $X_v$ given its parent variables $X_{pa(v)}$ in the graph.
                \item A set of utility functions $U$, with each utility function $u(X_{pa(v)})$ associated with a node $v$ in the subset of utility nodes $V_U \subseteq V$.
            \end{enumerate}
        \end{definition}

        To determine the decision option with the highest expected utility, we calculate the expected utility for each decision alternative. Let $A$ be a decision variable with options $a_1, \dots, a_m$, $H$ a hypothesis with states $h_1, \dots, h_n$, and $\epsilon$ a set of observations as evidence. The probability of each hypothesis outcome $h_j$ and the expected utility of each action $a_i$ can be computed. The utility for an outcome $(a_i, h_j)$ is denoted as $U(a_i, h_j)$, where $U(\cdot)$ is the utility function. The expected utility for action $a_i$ is given by:
        
        \begin{equation}
        EU(a_i) = \sum_{j=1}^n U(a_i, h_j) \times P(h_j | \epsilon)
        \end{equation}
        
        Here, $P(\cdot)$ represents the belief in hypothesis $H$ given evidence $\epsilon$. The utility function $U(\cdot)$ quantifies the decision maker's preferences numerically.
        
        The optimal decision is made using the principle of maximum expected utility, selecting an option $a^*$ such that:
        
        \begin{equation}
        a^* = \mathrm{argmax}_{a_i \in A} \, EU(a_i)
        \end{equation}
        
        Dynamic influence diagrams extend traditional influence diagrams by incorporating discrete time elements, effectively creating a time-sliced model. This approach involves replicating a static network structure across multiple time slices, where each slice represents the system at a specific point in time. The progression of the system over time is captured through connections between variables across these different time slices. Essentially, a dynamic model can be visualized as a series of static models placed sequentially, each depicting the system at a distinct time step. The links between these time steps illustrate the impact of the system's past state on its present state. In our experiment, we utilized a finite horizon dynamic influence diagram, which means our model was designed to consider a specific, limited number of time steps into the future.
                
        \vspace{0.1in}
        \subsubsection{Deep Reinforcement Learning (DRL)}
        Deep Reinforcement Learning (DRL) \cite{franccois2018introduction} combines deep neural networks with reinforcement learning, offering an approach for training to make sequential decisions in complex and dynamic settings. In Process control and its optimization, where traditional control techniques may struggle to address the complexities and uncertainties inherent in real-world processes, DRL has proven to be beneficial \cite{spielberg2020deep}. Mathematically, reinforcement learning involves Markov Decision Processes (MDPs), where an agent interacts with an environment by taking actions based on its current state, receiving rewards, and updating its policy to maximize cumulative future rewards. The Q-value function, denoted as Q(s, a), represents the expected cumulative reward of taking action 'a' in state 's' and following the optimal policy thereafter. The Bellman equation, a fundamental concept in reinforcement learning, expresses the recursive relationship between Q-values. Authors in \cite{abbas2022deep} have delved into the application of DRL in the process control environment.
        
        \begin{equation}
        Q(s, a) = \mathbb{E}[r + \gamma \max_{a'} Q(s', a') | s, a]
        \end{equation}
        
            \paragraph{Twin Delayed Deep Deterministic Policy Gradient (TD3) Architecture}
            Twin Delayed DDPG (TD3) architecture \cite{fujimoto2018addressing}, proposed to tackle challenges like overestimation bias. TD3 serves as the foundational DRL architecture in our framework, incorporating twin Q-value estimators to alleviate overestimation errors and a delayed policy update mechanism for stabilizing the learning process. The TD3 algorithm follows the actor-critic approach, where the actor decides and the critic evaluates the policy.

            \paragraph{Critic Update}
            \begin{equation}
                \mathcal{L}(\theta_Q) = \mathbb{E}_{(s, a, r, s') \sim \mathcal{D}}\left[\frac{1}{2}(Q_{\text{1-target}}(s, a) - y)^2\right]
            \end{equation}
            
            \begin{equation}
                y = r + \gamma(1 - d) \min_{i=1,2} Q_{i_{\text{-target}}}(s', \pi_{\text{target}}(s'))
            \end{equation}

            \paragraph{Actor Update}
            \begin{equation}
                \mathcal{L}(\theta_{\pi}) = -\mathbb{E}_{s \sim \mathcal{D}}[Q_{\text{1}}(s, \pi(s))]
            \end{equation}
            
            \paragraph{State}
            Within the framework of a Partially Observable Markov Decision Process, relying solely on the observed state may be insufficient due to inherent partial observability constraints. To overcome this challenge, our research employs a tuple consisting of the history of expert actions concatenated with the history of process variables. This history extends up to a length denoted as "l," as illustrated in \cref{eq:state}. The selected historical information includes the current state at time "t" and the preceding trajectory at time "t - 1".
            
            \begin{equation}
            s_t:=\left\langle\left(y_{t-l}, a_{t-l-1}^E\right), \ldots,\left(y_t, a_{t-1}^E\right)\right\rangle
            \label{eq:state}
            \end{equation}

            \paragraph{Action}
            In the context of the TD3 architecture, the actor-network outputs a continuous action parameterized by a neural network. The actor's output is denoted as:
            
            \begin{equation}
            \label{eq: drl_actor}
                \pi_{\theta}(s) = \mu(s)
            \end{equation}
            
            Here, \( \mu(s) \) is the deterministic policy function, representing the mean of the distribution over the continuous action space.

            \paragraph{Reward}
            In a disturbance rejection scenario, the agent seeks to determine an optimal policy that minimizes tracking error and stabilizes the process while deviating minimally from the optimal set point. This objective is incorporated into the DRL agent through a reward function (r) or a cost function (-r), such as the negative l1-norm of the set-point error, as expressed in \cref{eq:reward}.
            
            \begin{equation}
            r\left(s_t, a_t^A, s_{t+1}\right)=-\sum_{i=1}^{m_y}\left|y_{i, t}-y_{i, \mathrm{sp}}\right|
            \label{eq:reward}
            \end{equation}        

        \vspace{0.1in}
        \subsubsection{Specialized Reinforcement Learning Agent}
\begin{figure}[!t]
            \centering
            \includegraphics[scale=0.25]{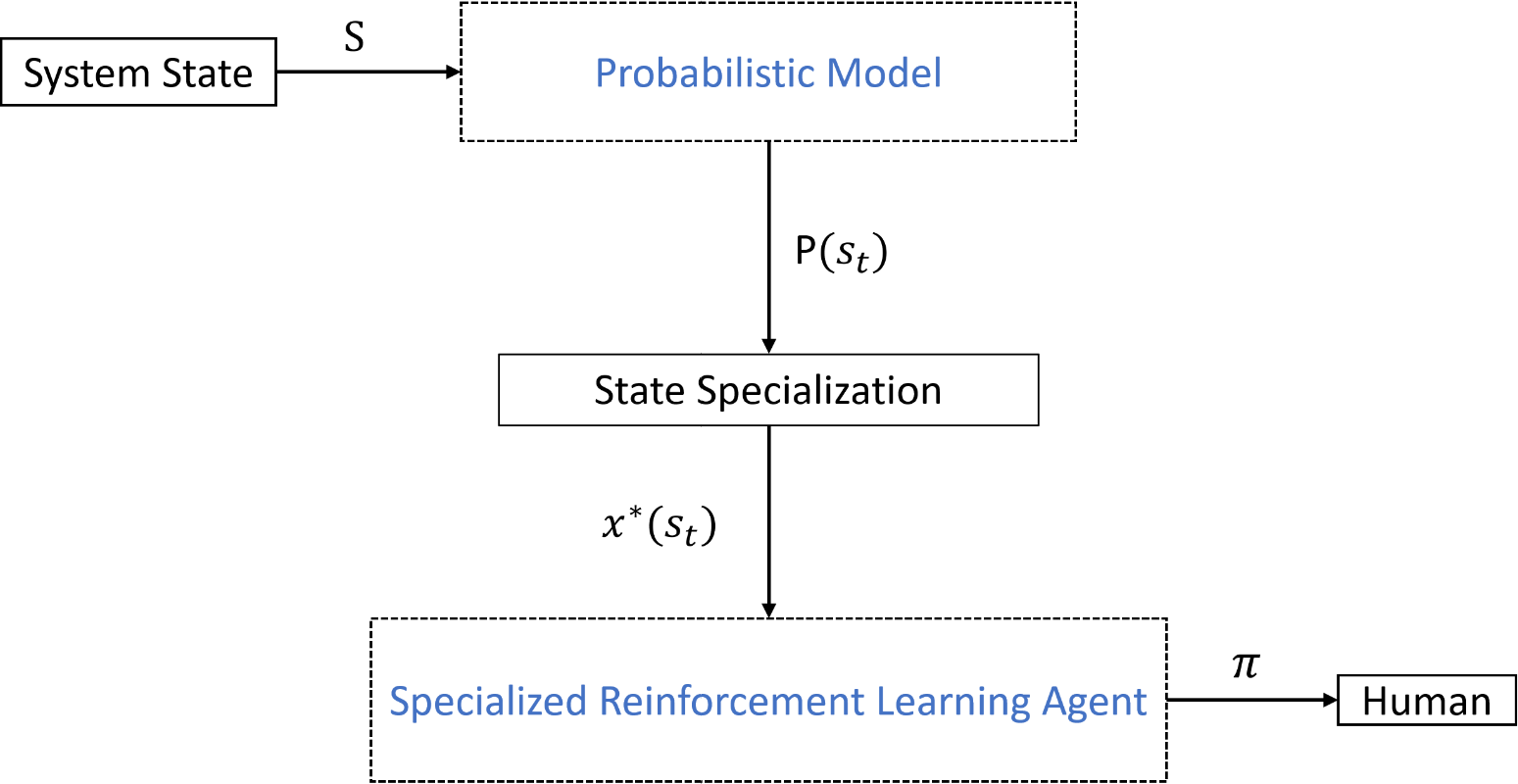}
            \caption{Specialized Reinforcement Learning Agent (SRLA). \\\hspace{\textwidth} Source: \cite{abbas2024hierarchical}} 
            \label{fig: srla}
        \end{figure}

        A Specialized Reinforcement Learning Agent (SRLA) integrates the strengths of probabilistic modeling and Deep Reinforcement Learning (DRL) as shown in \cref{fig: srla}, proposed in \cite{abbas2024hierarchical}. SRLA enables the DRL agent to specialize in specific scenarios within the environment, particularly in cases involving process abnormalities. This specialization enhances training efficiency and reduces the need for excessive data. In the associated figure, $P(s_t)$ denotes the probability of a particular state, $x^*(s_t)$ signifies the specialized state where the DRL agent is activated, and the system state $S$ is filtered to extract information about that specific state. The recommended optimal control strategy $\pi$ is then presented to the operator. An instantiation of the framework was adapted for a case study in process control and optimization, as presented in \cite{abbas2023specialized}. The actor and critic updates are modified as shown in \cref{eq: srla} where $s^*=$ desired state identified through the probabilistic model.

        \begin{equation}
            \begin{aligned}
            & \mathcal{L}_A^*\left(\theta_\pi\right)=-Q\left(s^*,\left(a^E+a^A\left(s^* \mid \theta_\pi\right)\right) \mid \theta_Q\right) \\
            & \mathcal{L}_Q^*\left(\theta_Q\right)=\frac{1}{2}\left(R-Q\left(s^*,\left(a^E+a^A\left(s^* \mid \theta_\pi\right)\right) \mid \theta_Q\right)\right)^2
            \label{eq: srla}
        \end{aligned}
        \end{equation}
        
        Furthermore, the framework was extended as an AI-enhanced recommendation system for process control, where a Multi-Specialized Reinforcement Learning Agent (M-SRLA) configuration was employed. In this setting, multiple agents operate independently, with only a specific agent activated to offer the optimal control strategy when a process abnormality is detected through the influence diagram as presented by \cite{mietkiewicz2023dynamic}.

        \vspace{0.1in}
        \subsubsection{Hidden Markov Model (HMM)}
        Hidden Markov Models (HMMs) \cite{rabiner1989tutorial} are probabilistic models widely used for modeling sequential data in diverse fields. They consist of a set of hidden states, each associated with a probability distribution over observable outcomes. The transitions between hidden states are governed by probabilities, and at each state, an observation is emitted based on another probability distribution. HMMs are characterized by their ability to capture temporal dependencies. A similar hybrid approach of combining HMM with DRL was used by authors in \cite{abbas2022interpretable} and \cite{abbas2023specialized}, where the sole purpose of HMM was to provide separate the state space into normal and abnormal states as well as to provide interpretations for the root cause of failure. In our proposed framework HMMs \cite{hmmlearn} are used in a similar context to predict the human failure in process abnormality states.

            \paragraph{State Transitions}
            \begin{equation}
                P(q_t | q_{t-1})
            \end{equation}
            where $P(q_t | q_{t-1})$ represents the transition probability from state $q_{t-1}$ to $q_t$.
            
            \paragraph{Emission Probabilities}
            \begin{equation}
                P(x_t | q_t) 
            \end{equation}
            where $P(x_t | q_t)$ represents the emission probability of observation $x_t$ given state $q_t$.

            \paragraph{Hidden State Prediction}
            Given the current state $q_{t-1}$, the observation sequence $x_1, x_2, \ldots, x_{t-1}$ up to time $t-1$, and the model parameters $\theta$ (initial, transition, and emission probabilities)  the predicted probability distribution for the next state $q_t$ is given by:  
            \begin{equation}
                \label{eq: hmm}
                P(q_t | x_1, x_2, \ldots, x_{t-1}, \theta)
            \end{equation}

    \subsection{Human-Centered Specialized Reinforcement Learning Decision Support Framework for Safety-Critical Systems}

\begin{figure}[!t]
        \centering
        \includegraphics[scale=0.23]{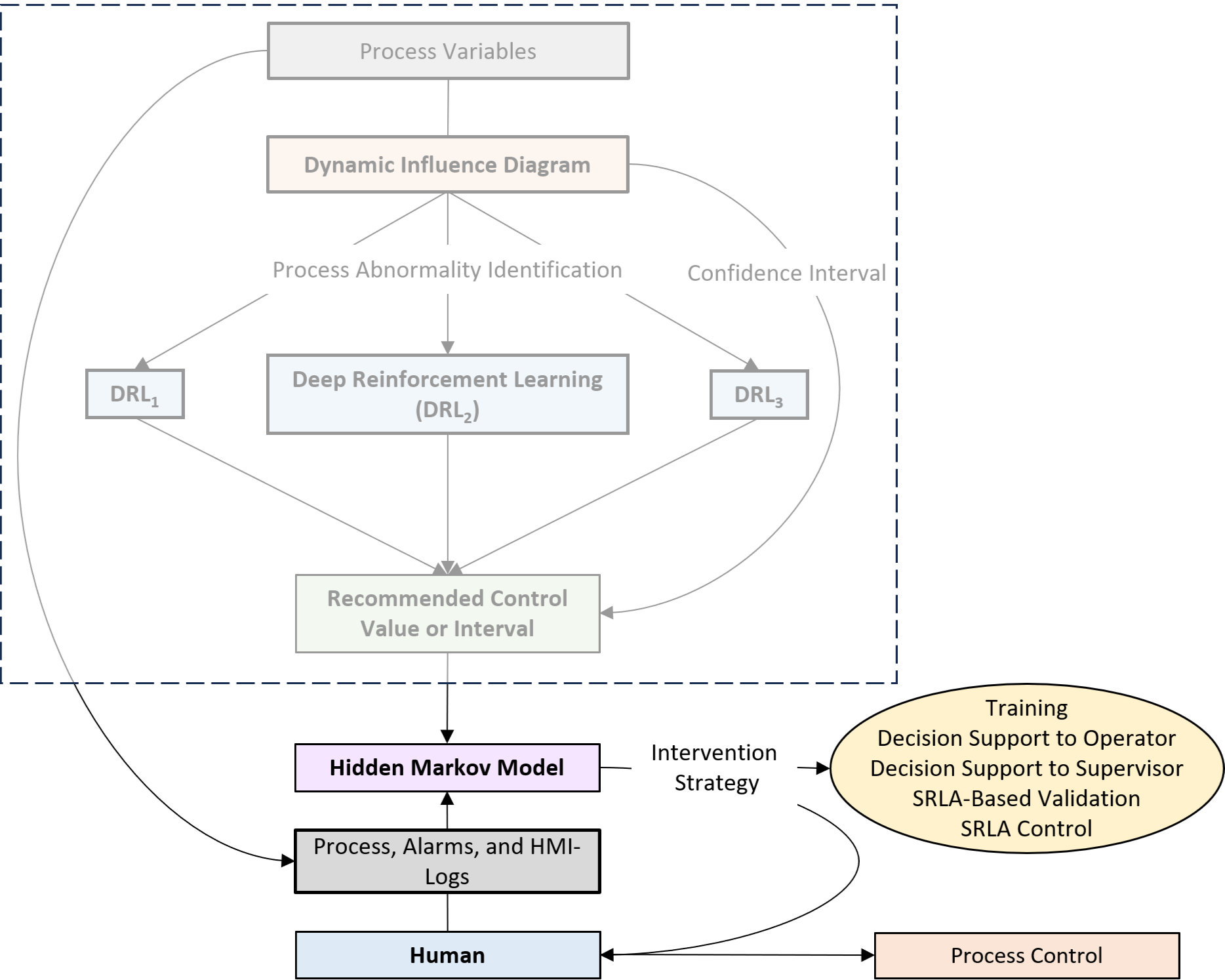}
        \caption{Human-Centered Specialized Reinforcement Learning Agent for Safety-Critical Systems. The dashed area represents the framework developed in \cite{mietkiewicz2023dynamic,mietkiewicz2024enhancing}.} 
        \label{fig: hc_srla}
    \end{figure}

    In this paper we build upon the previous framework \cite{mietkiewicz2023dynamic} and introduce an extended version as shown in \cref{fig: hc_srla} that incorporates a Human-in-the-Loop (HITL) setup. The proposed framework additionally captures the human state prediction in real-time by the use of the Hidden Markov Model (HMM) on the data from process, alarms, and HMI-logs, which provide the information of the process variables as well as the human interaction with the process. This framework further increases the capabilities in terms of the intervention strategies such as using it during the operator's training, decision support for the operator or supervisor, operator action validation, and requesting automating of the intervention through DRL.

    The methodology involves utilizing an influence diagram to locate specific failures within the system globally. Once the failure-associated procedure is identified, the relevant step of the procedure is presented to the operator based on the current situation. In cases where the procedure requires the manual adjustment of a controller with continuous values, locally specialized deep reinforcement learning is activated to determine the appropriate value, considering the current state of the system. One additional safety barrier that we introduce for using such a black-box system in safety-critical industries is that the exact continuous value is only suggested to the operator if it is found within the confidence interval of the DID. The influence diagram, due to data discretization, provides a value in the form of an interval. If the value derived from deep reinforcement learning falls outside this interval, only the procedure, along with the interval, is presented to the operator. This precaution is taken to ensure the interpretability and safety of the overall system. In summary, the influence diagram serves to model the system globally, while reinforcement learning precisely addresses local issues.

    Furthermore, the HMM is introduced at the final layer between the human and the recommended control value that determines different intervention strategies based on the state of both the system and the human as derived from the real-time process logs, which includes the data of current process variables, alarm information, and human-system interactions. The HMM predicts if the human state will be able to handle the situation given the time and circumstances and based on this it suggests the operator to either adjust the control manually or allow the system to automate certain processes. HMM can also be used to validate the actions of the operator according to the expert standards and prompt to avoid human errors. Moreover, the proposed framework is versatile enough to be used in real situations as well as during the training of the operator to enhance the guidance. 
    
        \subsubsection*{Algorithm}

        The algorithm is defined in \cref{alg: hc-srla} and is available in the project  \href{https://github.com/ammar-n-abbas/drl-based-decision-support}{repository}.
        
        \begin{algorithm}
            \caption{Human-Centered Specialized Reinforcement Learning Agent (HC-SRLA)}
            \label{alg: hc-srla}
            \begin{algorithmic}
            
                \STATE {\bfseries \emph{STEP I:}} Dynamic Influence Diagram Monitoring
                \STATE \quad {\bfseries Input:}
                \STATE \quad \quad $y$: process variables
                \STATE \quad \quad \textit{DID}: trained model
                \STATE \quad {\bfseries Output:}
                \STATE \quad \quad $s^*$: specific event of interest (such as abnormality)
                \STATE \quad \quad \textit{recommendation:} pruned procedure \newline
        
                \STATE {\bfseries \emph{STEP II:}} Deep Reinforcement Learning Inference
                \STATE \quad {\bfseries Input:}
                \STATE \quad \quad $s^*$: specific event of interest (such as abnormality)
                \STATE \quad \quad $y$: process variables
                \STATE \quad \quad $\pi_{\theta}(s)$: trained specialized actor
                \STATE \quad \quad \quad Safety control confidence interval check
                \STATE \quad {\bfseries Output:}
                \STATE \quad \quad $\mu(s)$: recommended control value (\cref{eq: drl_actor} or interval \newline
        
                \STATE {\bfseries \emph{STEP III:}} Hidden Markov Model Monitoring
                \STATE \quad {\bfseries Input:}
                \STATE \quad \quad $logs$: process, alarms, and HMI-logs 
                \STATE \quad \quad \textit{HMM}: trained model parameters
                \STATE \quad {\bfseries Output:}
                \STATE \quad \quad $q_t$: hidden state predition from \cref{eq: hmm} \newline
        
                \STATE {\bfseries \emph{STEP IV:}} Decision Support and Human Intervention
                \STATE \quad {\bfseries Input:}
                \STATE \quad \quad $q_t$: hidden state interpretation (such as human failure)
                \STATE \quad {\bfseries Output:}
                \STATE \quad \quad \textit{intervention}: intervention strategy suggestion
                \STATE \quad \quad \textit{control}: human intervention
            \end{algorithmic}
        \end{algorithm}

\section{Case Study}
\label{sec:case_study}

The case study, conducted in collaboration with Politecnico Di Torino and Technological University Dublin, involves a simulated chemical plant dedicated to formaldehyde production. The plant, which produces a 30\% formaldehyde solution at a rate of 10,000 kg/h through partial oxidation of methanol with air, consists of three main sections: the feed section (comprising various systems such as nitrogen flow, methanol tank, pumps, boiler, compressors, heaters, piping, controllers, safety valves, and indicators), the heat and recovery section (housing three heat exchangers), and the reaction and separation section (featuring a reactor, controllers, alarms, sensors, rupture disk, absorber, and piping).

The study focuses on hazardous events or process safety occurrences, including depressurization of the methanol tank, air ingress into the methanol tank leading to the formation of a flammable atmosphere, and reactor overheating. These scenarios are categorized based on complexities (normal and abnormal situations) within different plant sections. The goal is to vary the task load as a variable and observe its impact on performance, analyzing it alongside other variables to enhance the understanding of the system's behavior.

The experiment aims to investigate the impact of various interactions between variables in human-machine and automation interactions on operator performance and its implications for process safety. The goal is to facilitate comprehensive risk-based decision-making for real-time support adaptation, process control optimization, and management of change. The key objectives include developing a real-time model to assess human performance in human-machine interaction environments, modeling safety data for Human-in loop configurations in process control, exploring optimal decision-making processes in safety-critical systems using Reinforcement Learning with human-in-the-loop, and creating Bayesian networks to assist human operators in recommending adaptive automation strategies.

    \subsection{Process Control Room Simulation Environment}

\begin{figure*}[!t]
        \centering
        \includegraphics[scale=0.12]{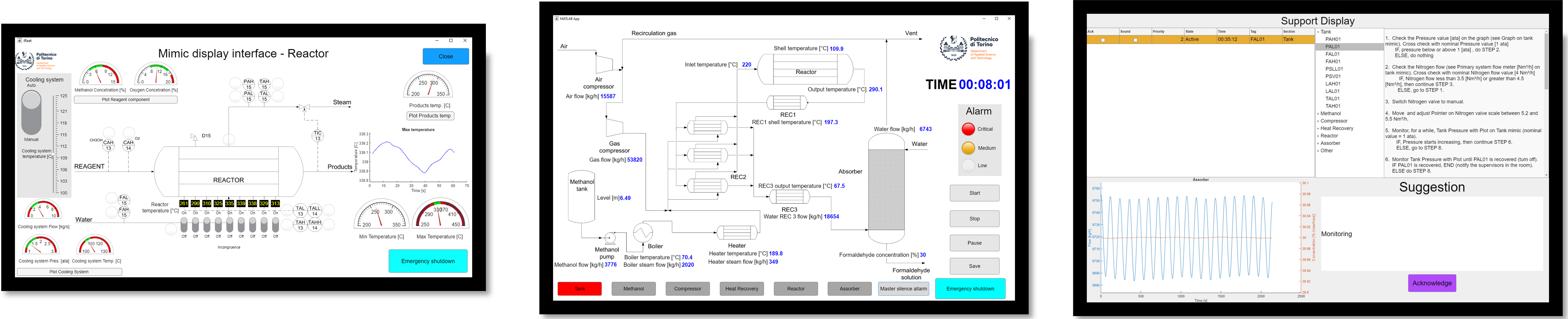}
        \caption{Simulator screens, (left) mimic subprocess, (middle) overview of the plant, and (right) support panel.} 
        \label{fig: screens}
    \end{figure*}

    Our experiment is based on a simulator created for formaldehyde production \cite{demichela2017risk}, which simulates a control room environment. A significant enhancement to this simulator is the "support panel" converting it into a comprehensive control room simulation with features like graphical production monitoring, an alarm list, a procedure list, and a suggestion box as shown in \cref{fig: screens}. 

    \subsection{Groups}
    For the comparative evaluation, the participants were divided into two groups, the one with the AI support system and the one without it. 
    
        \subsubsection{GroupN (Without AI System)}
\begin{figure}[!t]
            \centering
            \includegraphics[scale=0.225]{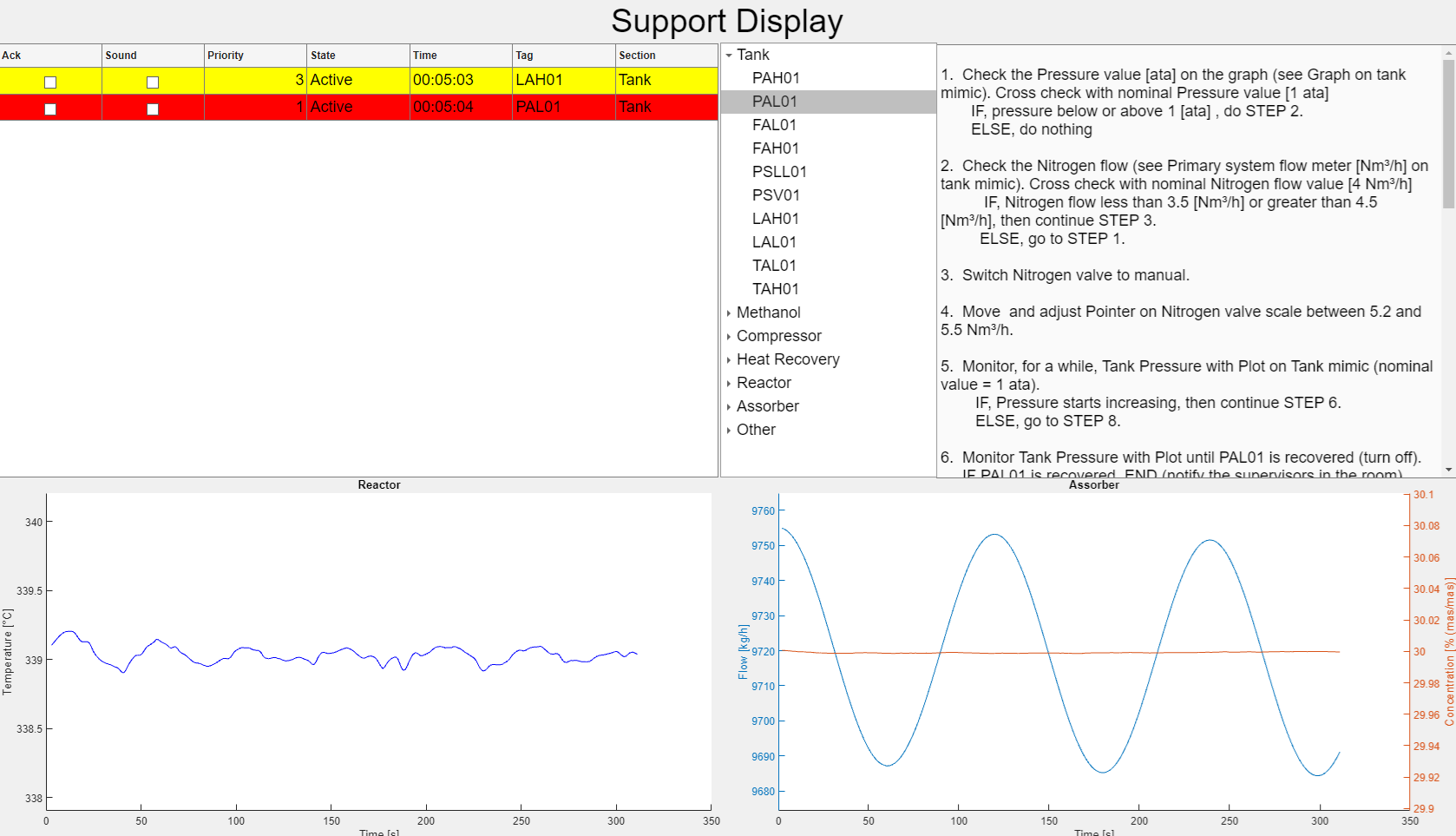}
            \caption{GroupN support panel including screen-based procedures.} 
            \label{fig: groupN_support}
        \end{figure}

        The group without the AI system has a screen-based procedure panel to manually go through the intervention procedure and had the difference in the last section of the support panel that does not include any suggestion box as shown in \cref{fig: groupN_support}. 
        
        \paragraph{Screen-Based Procedure Panel}
        The screen-based procedures provided the participants with the intervention procedures for all the alarms per every sub-section of the plant. The participant has to click on the specific alarm which has to be recovered and follow the procedure accordingly.
        
        \subsubsection{GroupAI (With AI System)}
\begin{figure}[!t]
            \centering
            \includegraphics[scale=0.18]{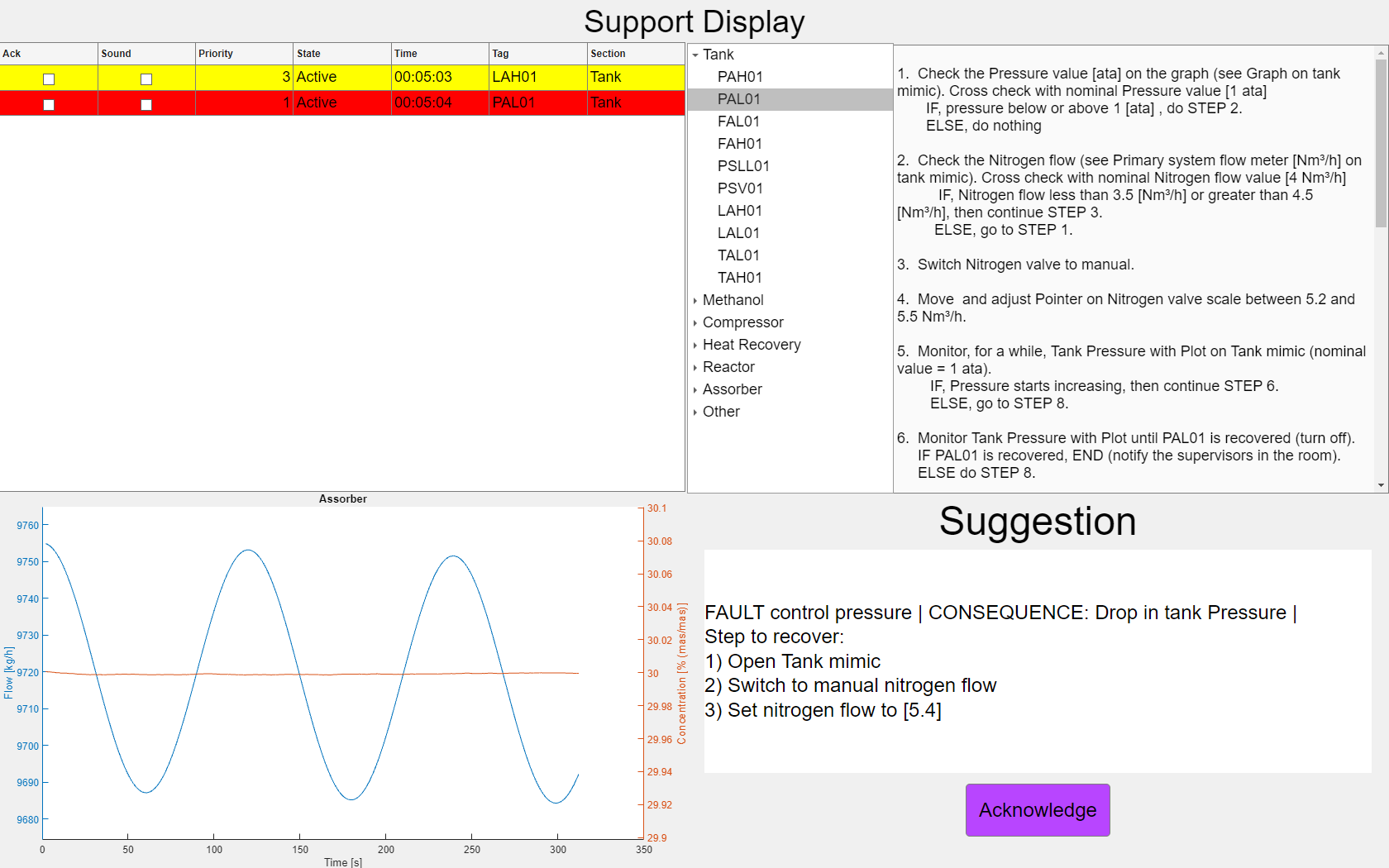}
            \caption{GroupAI with support panel including screen-based procedures and AI-based suggestion box.} 
            \label{fig: groupAI_support}
        \end{figure}

        The only difference between GroupAI from GroupN is an additional panel of suggestion boxes as shown in \cref{fig: groupAI_support}.
        
        \paragraph{AI-Enhanced Decision Support Panel}
        The AI-enhanced decision support panel introduces a concise representation of the intervention procedure to be followed and gives the root cause of the predicted failure even before the alarm. The probabilistic model predicts the failure and deep reinforcement learning provides the exact analog value for the controller to be manually configured. If the participant agrees to follow the suggestions by the AI system they have to acknowledge and then follow the guidelines. Furthermore, to gain the attention of the participant the acknowledge box starts blinking if there is any change in the suggestion so that the operator (participant) can monitor that change and perform the suggested action.
        
    \subsection{Scenarios}

    To evaluate the effectiveness of our decision support, we devised three scenarios, and further analysis of the scenarios was divided according to the Time of Interest (TOI):
    
        \subsubsection{Baseline Overview}
        This TOI refers to the condition where the operator just observes the overall process the time before the occurrence of the first critical alarm. This created a baseline performance analysis of the participant in terms of physiological measures such as an eye tracker.

        \subsubsection{Critical Alarm}
        This TOI refers to the start of the first critical alarm to the time when it is either fully recovered or the end of the experiment. The three scenarios included within this TOI are as follows:
        
            \paragraph{Pressure indicator control failure} The automatic pressure management system in the tank malfunctions, requiring the operator to manually adjust nitrogen inflow to maintain pressure. The interruption of nitrogen flow leads to a pressure drop as the pump continues channeling nitrogen into the plant.
        
            \paragraph{Nitrogen valve primary source failure} Similar to the first scenario, the primary nitrogen source fails, prompting the operator to switch to a backup system. While the backup system starts slowly, the operator regulates pump power to slow down the pressure drop in the tank.
    
            \paragraph{Temperature indicator control failure in the Heat Recovery section} The operator attempts to resolve the issue by adjusting the cooling water flow in the absorber manually.
        
        \subsubsection{Scenario 3: Alarm Overflow}
        This TOI refers to the start of the beginning of the alarm overflow (second critical alarm) to the end of the experiment. Scenario 3 included within this TOI is as follows:
                
            \paragraph{Temperature indicator control failure in the Heat Recovery section} The operator attempts to resolve the issue by adjusting the cooling water flow in the absorber manually. However, this proves ineffective, and the operator contacts the supervisor. The supervisor advises that the problem exceeds control room resolution and requires a field operator's intervention. While the field operator addresses the issue on-site, the control room operator manages the reactor's temperature, with a focus on preventing potential reactor issues.

    As the focus of this analysis is not the comparison across the scenarios, therefore, the results from different scenarios were normalized across the scenarios and aggregated as an average for each participant, creating a single data point for each participant per group to reduce comparison complexity. Furthermore, to normalize metric within its corresponding baseline TOI (where applicable) \cite{mathot2018safe}, the following \cref{eq: eye_norm} was applied:

    \begin{equation}
        \text{\textit{Metric}} = \frac{\text{\textit{Metric in TOI}}}{\text{\textit{Metric in Baseline TOI}}}
        \label{eq: eye_norm}
    \end{equation}
    
\section{Data Collection} 
\label{sec:data_coll}

The dataset encompasses measurements obtained from diverse data sources. They incorporate both objective and subjective metrics commonly employed in assessing cognitive states related to workload, situational awareness, stress, and fatigue. Various data collection tools such as health monitoring watch, eye tracking, process and HMI logs, operational metrics (response time, reaction time, performance, etc), NASA Task Load Index (NASA-TLX), Situation Awareness Rating Technique (SART), a think-aloud Situation Presence Assessment Method (SPAM), AI support questions, and AI\_vs\_human error were utilized. A detailed explanation of the dataset is provided in this data article \cite{AMAZU2024110170} and the dataset is available in this \href{https://zenodo.org/doi/10.5281/zenodo.10569181}{repository} \cite{ammar_n_abbas_2024_10600674}.

Furthermore, the datasets encompass information derived from a simulated formaldehyde production plant, involving participant interaction within a controlled experimental setting resembling a control room. The human-in-the-loop scenario included tasks like Monitoring, Alarm Handling, Recovery planning, and intervention (Troubleshooting, Control, and Evaluation). Data collection involved 47 participants divided into two groups, each undergoing the specified task flow. Participants tested three scenarios lasting 15–18 minutes, with breaks and survey completion periods in between, utilizing different combinations of decision support tools. The decision support tools varied across groups, encompassing factors of digitized screen-based procedures and the inclusion of an AI recommendation system.

\begin{figure}[!t]
    \centering
        \includegraphics[width=0.95\linewidth]{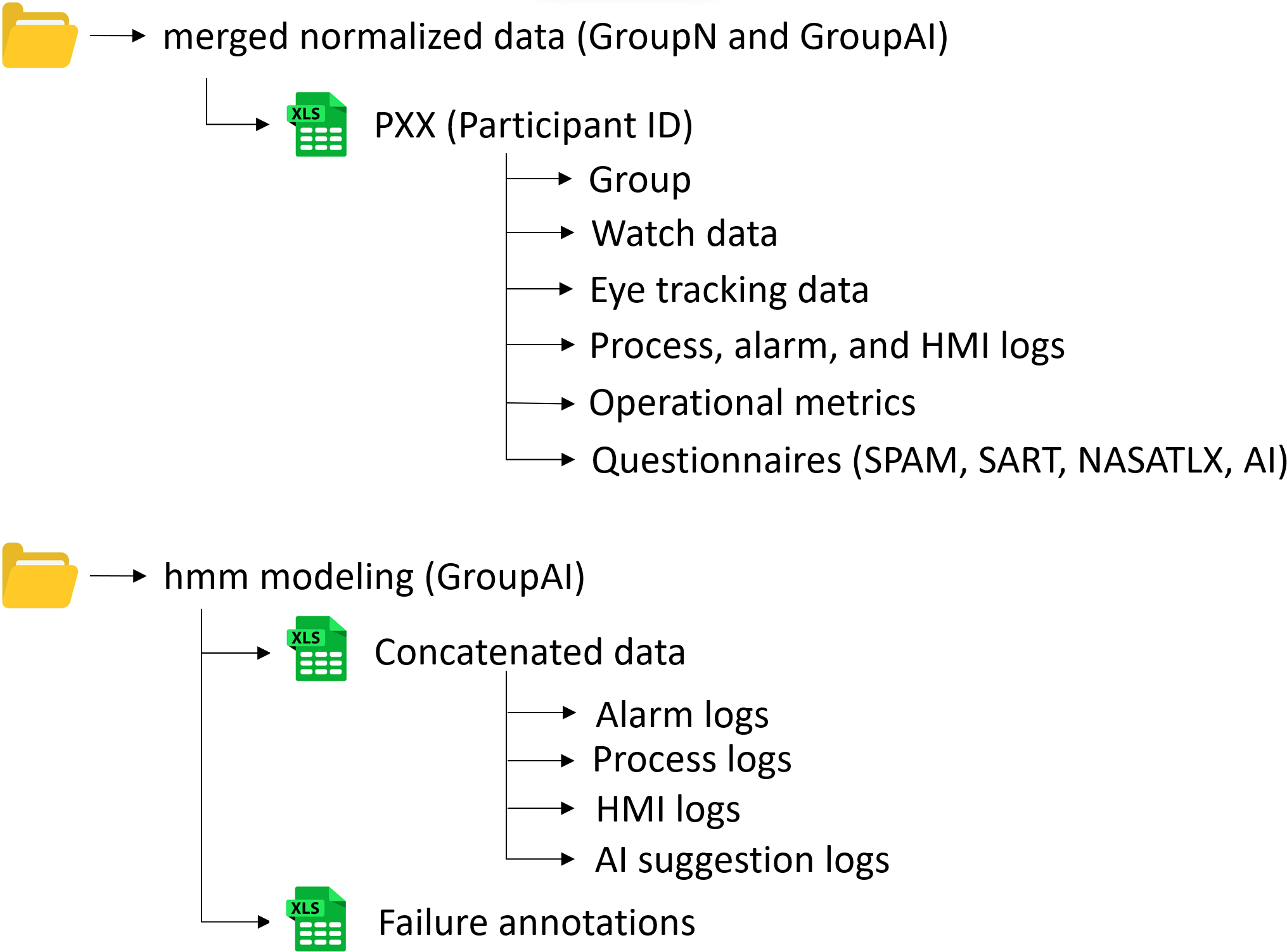}
        \caption{Dataset file hierarchy}
        \label{fig: data-hierarchy}
\end{figure}

The significance of this research lies in its relevance to comparing current industry practices and their impact on operators' performance and safety. It is also applicable for validating proposed solutions within the industry. The dataset is utilized for statistical analysis to compare outcomes among different groups. These datasets have potential applications for decision-makers involved in control room design and optimization, process safety engineers, system engineers, human factors engineers in process industries, and researchers in related domains. The hierarchy of the dataset is shown in \cref{fig: data-hierarchy} and the processed dataset can be found in this \href{https://github.com/ammar-n-abbas/drl-based-decision-support}{repository} \cite{ammar_n_abbas_2024_10641061}.

The collected raw data was processed particularly for the analyses in this paper. The data from individual participants was concatenated and merged in a single xls file for further evaluation. The data used for comparison between the GroupN and GroupAI is presented in the merged normalized data folder. The xls file contains the data points for each participant per row and each column represents the data and sub-data collected from various sources. As the focus of this analysis is not between the scenarios, therefore, the data is normalized across scenarios to avoid the effect of the scenarios in the analysis and is averaged for every participant to acquire a single vector of data for each participant.

In the folder of "hmm modeling" the concatenated data represent the time-series data of the process, alarms, and HMI logs for every participant into a single file as is required by the HMM python library \cite{hmmlearn}. Further, a separate file is included that provides the labels for participants who failed during the task based on various factors such as the consequence of plant shutdown or reactor overheating and overall performance.

    \subsection{Ethics Statement}
    \emph{This research study was conducted following the ethical guidelines set forth by the Technological University Dublin Ethical Review Committee. Ethical approval for this study was obtained from the Technological University Dublin Ethical Review Committee (REC Approval Number: REC-20-52A). All participants provided informed consent before participating in the study, and their confidentiality and privacy were strictly maintained throughout the research process. Any potential risks to participants were minimized, and steps were taken to ensure the well-being of all individuals involved in the study.}
    
    \subsection{Value of the Data}
    
    \subsubsection{Optimizing Human-AI Interaction} The dataset provides an opportunity to study the integration of human-in-the-loop configurations with AI systems in safety-critical industries. By examining the data, researchers can identify the factors necessary for successful collaboration between humans and AI. This knowledge can lead to the development of optimized interaction mechanisms, ensuring that the strengths of both humans and AI are leveraged effectively to enhance decision-making in critical scenarios. 
            
    \subsubsection{Evaluation of AI-Enhanced Decision Support System} The dataset allows for qualifying and quantifying the performance and effectiveness of the AI-enhanced decision support system incorporating Deep Reinforcement Learning (DRL). By analyzing the data, researchers can assess how well the system performs in safety-critical process industries with human-in-the-loop configurations, which is rarely observed. This evaluation can provide insights into the potential benefits, scope, and limitations of utilizing DRL in such contexts. 

    \subsection{Data Explanation}
    In this section, a brief description of each variable in the dataset has been provided. 

      \vspace{0.1in}
        \subsubsection{Biometric Measures}
        The biometric data for this study is described in \cite{amazu2023experiment}. This includes data on the pulse rate, electrodermal activity, and temperature of participants during the test. A brief description of each measure is also given below.
           
            \paragraph{Electrodermal Activity (EDA) or Galvanic Skin Response (GSR)} 
             It measures the skin’s electrical conductance of the skin. It is influenced by sweat gland activity, for example, the skin's moisture level.

            \paragraph{Pulse Rate or Heart Rate} 
             It defines the number of heartbeats per unit time (bpm).

            \paragraph{Temperature} 
            The body temperature here refers to the degree of coldness or hotness of the body.
        
        \vspace{0.1in}
        \subsubsection{Eye Tracker}
        Tobii Pro Glasses 3 \cite{tobii_pro_glasses_3} and Tobii Pro Lab \cite{tobii_pro_lab} analysis software were used in this experiment for eye tracking and extracting useful metrics within the defined Time of Interest (TOI), allowing for better evaluation of visual attention dynamics:

                \begin{itemize}
                \item Baseline Overview (pre-alarm occurrence): From the start of the experiment to the start of the first critical alarm (scenarios 1, 2, and 3).
                \item Critical Alarm: From the start of the first critical alarm to the end of it (scenarios 1, 2, and 3).
                \item Alarm Overflow: From the start of the second critical alarm to the end of the experiment (scenario 3).
                \end{itemize}
       
            All the eye tracker metrics were categorized based on these TOIs for evaluating visual attention across distinct phases of the experiment. To normalize each eye tracker metric within its corresponding TOI, \cref{eq: eye_norm} was applied. This normalization process ensured that the metrics were evaluated relative to the baseline, facilitating meaningful comparisons and insights.
            
            \paragraph{Fixation}   
            Fixation is the stable gaze or sustained focus on a specific point in the visual field. During fixation, the eyes remain relatively stationary, allowing detailed processing of visual information at that location.
                \begin{itemize}
                    \item Duration: Fixation duration is the amount of time individuals concentrate on a particular point of interest. This measurement helps identify elements that capture extended attention, enhancing our understanding of information processing and cognitive involvement.
                    \item Pupil Diameter: Pupil diameter during fixation serves as a vital metric indicating shifts in cognitive load and emotional arousal. Analyzing changes in pupil size allows us to comprehend the cognitive effort and emotional reactions linked to particular visual stimuli.
                \end{itemize}

            \paragraph{Saccade} 
            It refers to rapid eye movements that occur between periods of fixation. It allows us to evaluate visual perception, information processing, and the dynamics of decision-making. Further characteristics during saccades include:
                \begin{itemize}
                    \item Amplitude: The distance covered during a saccade.
                    \item Velocity: The speed of the eye movement.
                    \item Peak Velocity: The maximum speed reached during a saccade.
                    \item Duration: The time taken for the completion of a saccade.
                \end{itemize}
                
            \paragraph{Heat Map}
            The heat map visually displays regions that attracted participants' visual attention. Brighter areas on the heat map signal a higher concentration of fixations, providing information about key focal points within the visual stimuli. This understanding is especially valuable when comparing the heat map with TOIs, revealing how visual attention changes during different phases of the experiment. 
            
        \vspace{0.1in}
        \subsubsection{Process, Alarms, and HMI Logs (online)}
        The performance and behavioral measures derived from the online logs and how they were derived are detailed in \cite{amazu2023operationallogs}. A brief description of what they mean is shown below.
            
            \paragraph{Alarms}
                \begin{itemize}
                \item Number of alarms annunciated.
                \item Number of alarms silenced.
                \item Number of alarms acknowledged.
                \end{itemize}
            
            \paragraph{No. of Procedures} The number of procedures opened during the duration of each scenario.

            \paragraph{No. of Mimics Opened} The number of mimics\footnote{Mimics: Graphical interfaces that represent the layout and components of the sub-processes. These mimics allow the user to analyze variables associated with the sub-process and to be able to manually control if necessary} opened during the duration of each scenario.
       
            \paragraph{AI Acknowledgement} The number of times the AI acknowledgment button was pressed during the duration of each scenario.

\begin{figure}[!t]
                \centering
                \subfloat[]{\includegraphics[width=0.95\linewidth]{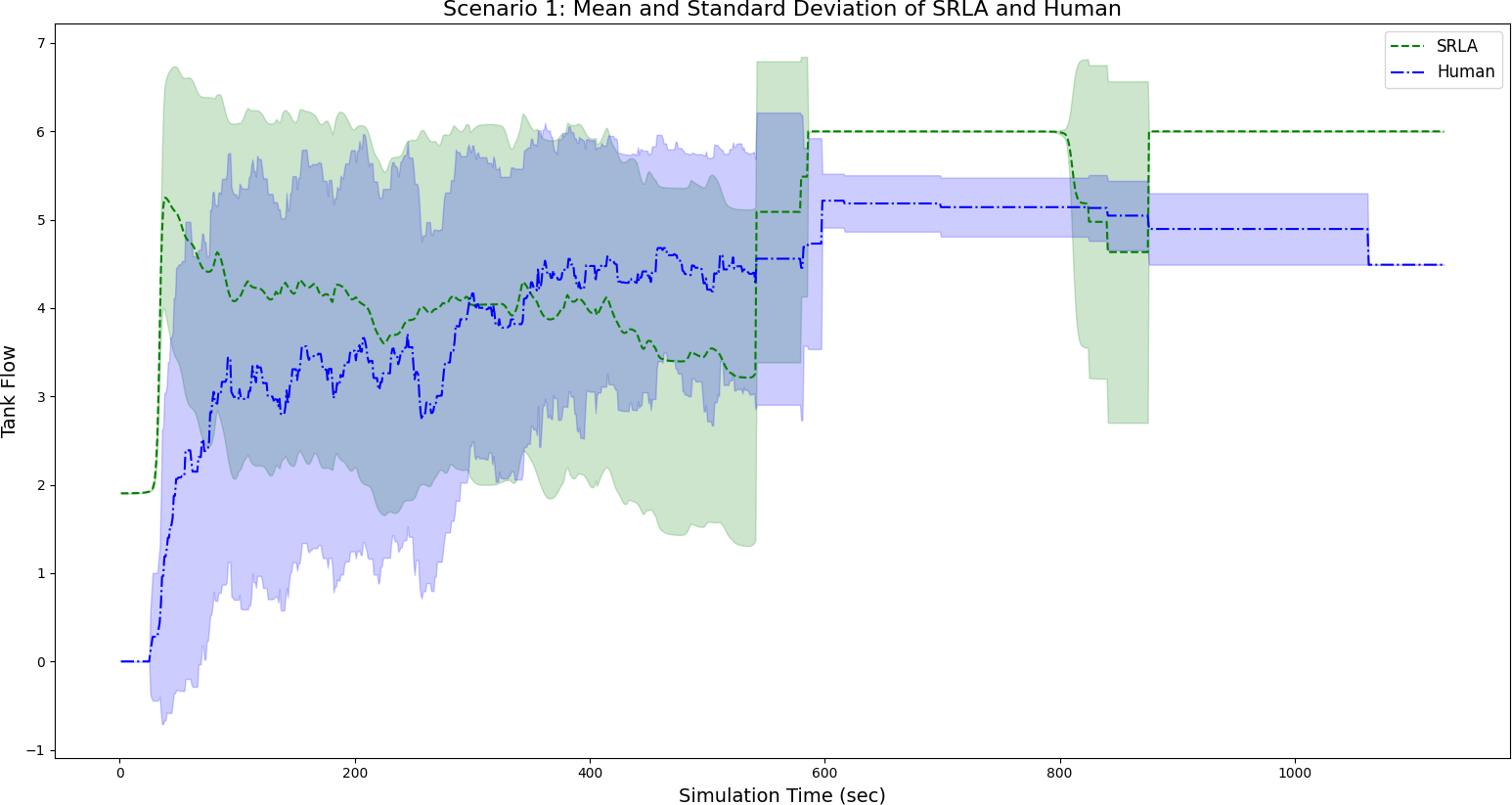}}
                    \\
                \subfloat[]{\includegraphics[width=0.95\linewidth]{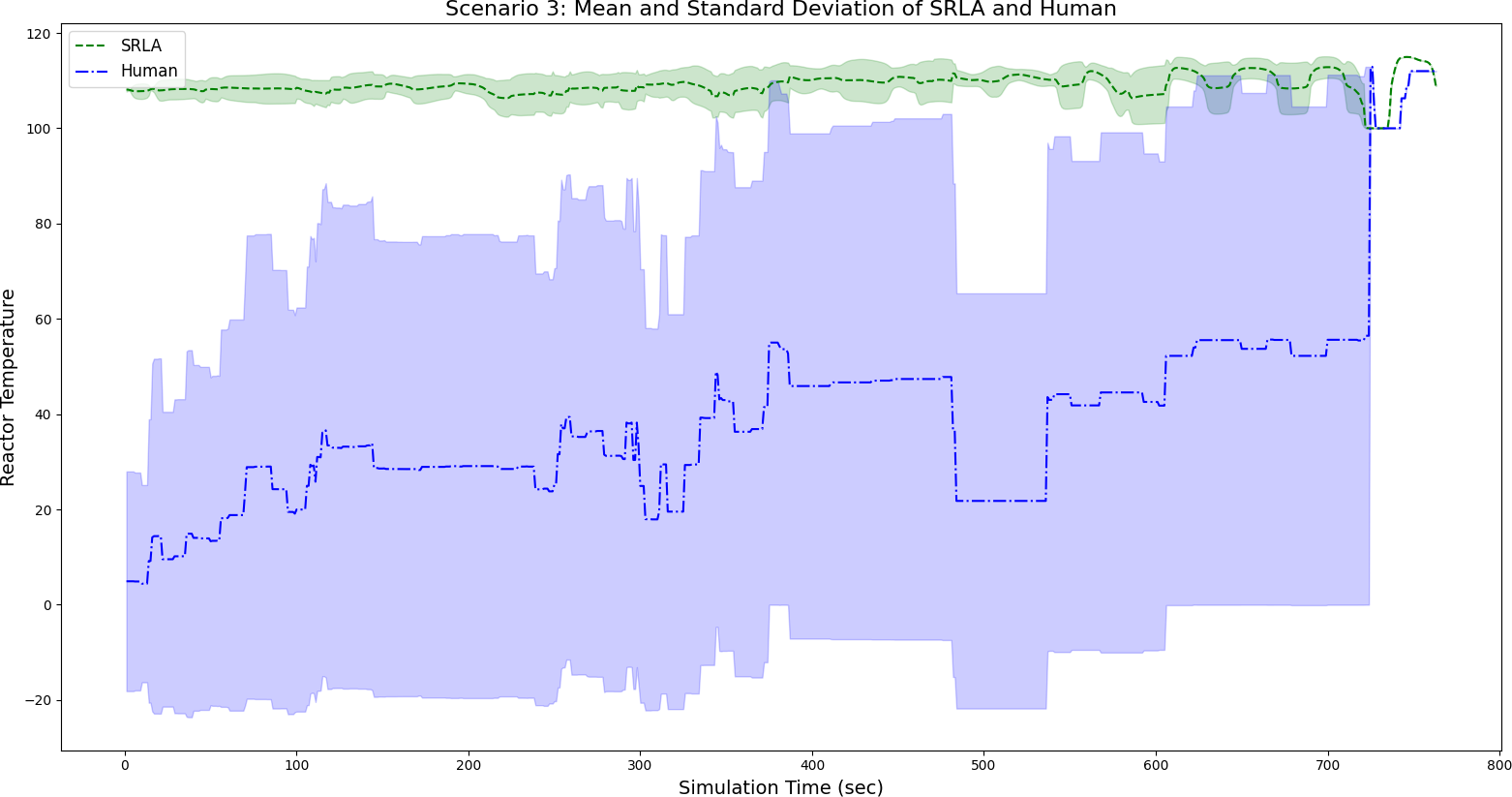}}
                \caption{Average DRL vs human response for 23 participants during (a) critical scenario: tank flow error and (b) overflow scenario: reactor temperature error.}
                \label{fig: srla-vs-human}
            \end{figure}

            \paragraph{AI vs Human Response}      
            Deviations and preliminary analysis of decisions taken by the human participant and suggestions by AI/DRL agent were analyzed and the mean error was calculated. \Cref{fig: srla-vs-human} illustrates the aggregated mean and standard deviation of the human vs AI (DID + SRLA) control for all the participants in GroupAI. It includes the scenarios of (a) critical alarm and (b) alarm overflow. The DRL control suggestion is based on the current state of the process and is considered to be ideally optimal in the case of this experiment. The optimality of the DRL suggestions is verified through a carefully controlled experimental design, barring any unforeseen deviations by the human participant. This can be further validated in the correlation analysis conducted later in \cref{sec:corr_anal}, revealing a positive correlation between higher errors and more adverse consequences. Therefore, the AI\_vs\_human measure was used to evaluate the performance of participants in terms of optimally following the suggestions.
    
        \vspace{0.1in}
        \subsubsection{Operational Measures (offline)}
        These metrics are derived from the online process logs and can only be measured once the experiment has been completed, hence these are computed offline. It gives the overall indication of the performance of the participant.
        
            \paragraph{Performance Measures}
                \begin{itemize}
                \item Recovery Status: Classified into optimal, good, and poor based on the participant's ability to manage the situation without an alarm annunciation (optimal, ability to recover the alarm even if annunciated (good), and failure to recover the alarm (poor).
                \item Accuracy: This metric quantifies the mean square error between the executed action and the prescribed control action (expert response, as specified in the operating procedure manual). Notably, the expert response over which the ``Accuracy" is measured is a constant average baseline and is not adapted to the current state of the system, unlike the AI\_vs\_Human response. Furthermore, accuracy is a common factor between both groups for easy comparison, however, the AI\_vs\_Human response is just for GroupAI.
                \item Consequence: Broken down into different levels depending on the event. The possible consequences are impurity of air in the tank atmosphere, plant shutdown, reactor overheating, and safe state.
                \item Overall Performance: Classified into optimal, good, and poor performance based on percentiles of the recovery time.
                \end{itemize}
            
            \paragraph{Behavioral Measures}  
                \begin{itemize}
                \item Recovery Time: The time to recover the critical alarm.
                \item Reaction Time: The time of the first control action.
                \item Response Time: The time it takes to Perform the last control action on the expected area of interest. 
                \end{itemize} 
            
        \vspace{0.1in}        
        \subsubsection{Questionnaires}
        Some questionnaire-based measures below have been detailed in a previous study by \cite{amazu2023impact}. A brief description is provided for each.
        
            \paragraph{Task Load} 
            Questions on how the participants perceived the complexity of the task were asked at the end of each scenario.   
            
            \paragraph{NASA-TLX, SART, and SPAM Indexes}
            The Task Load Index (TLX) and Situation Presence Assessment Method (SPAM) indexes are calculated as the average of their six and three dimensions, respectively. Situation Awareness Rating Technique (SART) is based on \cref{eq: sart}, as previously detailed in \cite{amazu2023impact} \footnote{SART Demand: Sum of the first three dimensions, SART Supply: Sum of dimensions 4, 5, 6, and 7, SART Understanding: Sum of the last three dimensions. Dimensions are orderly presented in \cref{tab: indexes}}. The thematic breakdown of the questions asked is shown in \cref{tab: indexes}.
            
            \begin{equation}
                \text{SART Understanding} = \text{SART Demand} - \text{SART Supply}
                \label{eq: sart}
            \end{equation}        
            
            \begin{table}[h]
                \centering
                \caption{Questions that were asked to evaluate each of these themes.}
                \begin{tabular}{|l||l||l|}
                    \hline
                    \textbf{NASA-TLX Index} & \textbf{SART Index} & \textbf{SPAM Index} \\
                    \hline
                    Mental demand & Instability & Monitoring \\
                    Physical demand & Variability & Planning \\
                    Temporal demand & Complexity & Intervention \\
                    Performance & Arousal & \\
                    Effort & Spare capacity & \\
                    Frustration & Concentration & \\
                    & Attention Division &\\
                    & Quantity &\\
                    & Quality &\\
                    & Familiarity &\\
                    \hline
                \end{tabular}
                \label{tab: indexes}
            \end{table}
            
            \paragraph{Alarm Prioritization Support}: this entails the perceived support of the participants on how well the alarm prioritization supported them during the scenario. 
            
            \paragraph{Procedures Support}: this refers to the perception of the participants on how well the procedures supported them during the task.

            \paragraph{AI Support Questions}
            These questions were asked after the end of the entire experiment (all 3 scenarios).
                \begin{itemize}
                \item Level of explainability of the AI suggestion.
                \item Level of trust for the AI suggestion.
                \item Helpfulness of the AI suggestion.
                \item Additional workload imposed by the AI suggestion.
                \item The tradeoff between the benefits of the AI system vs the additional workload.
                \item Importance of validating the AI suggestion by manually going through the screen-based procedures.
            \end{itemize}

\begin{figure}[!t]
                \centering
                    \includegraphics[width=0.7\linewidth]{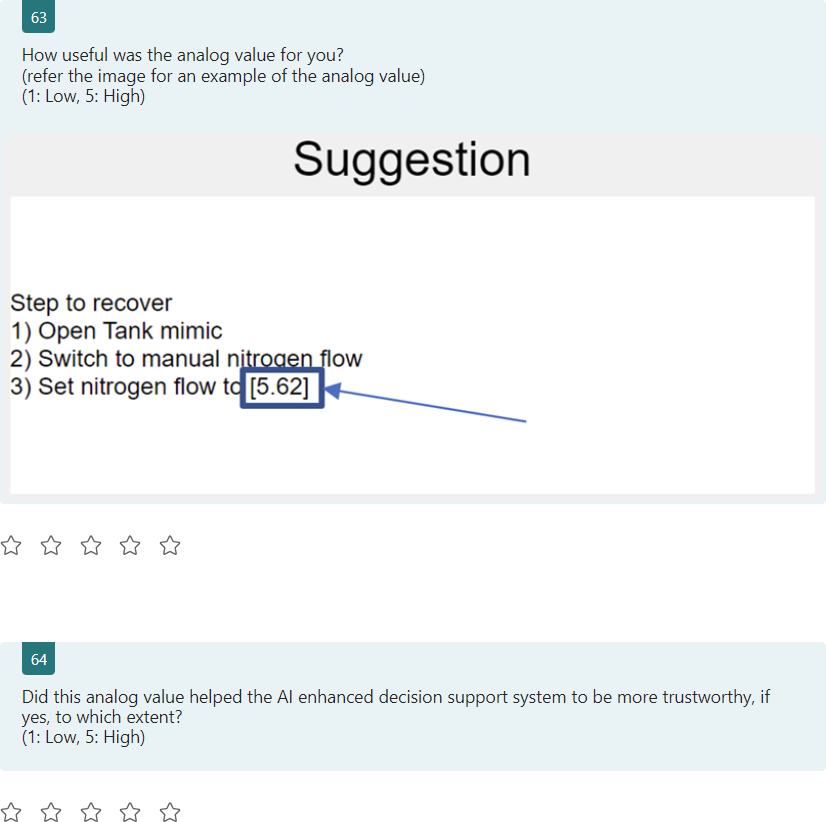}
                    \caption{Questions related to deep reinforcement learning suggestions}
                    \label{fig: drl-questions}
            \end{figure}

            \paragraph{Questions Related to DRL}
            Some specific questionnaires were asked for the DRL analog value suggestions provided to the human participant as shown in \cref{fig: drl-questions} 

                \begin{itemize}
                \item Importance of the DRL (analog) value in the AI suggestion.	
                \item Increase in trust due to the DRL (analog) value (if any).
                \end{itemize}
            
The data analysis is divided into two main sections. In \Cref{sec:eda}, we compare GroupN and GroupAI, and within GroupAI, we focus on participants. The objective is to assess the impact of the AI-enhanced decision support framework, specifically, the dynamic influence diagram-based deep reinforcement learning outlined in \cite{mietkiewicz2023dynamic, mietkiewicz2024enhancing}. This analysis lays the groundwork for understanding the dataset, recognizing patterns, and informing the development of an extended framework for real-world implementation. Moving on to \Cref{sec:hmm_pred}, this section validates the extended framework. It offers a detailed analysis of observations and explores potential future applications in industries. Specifically, it delves into the human-in-the-loop specialized deep reinforcement learning framework that utilizes a hidden Markov model for real-time system failure prediction based on human and process states.

\section{Analysis I: Exploratory Data Analysis}
\label{sec:eda}

An experimental study was conducted to evaluate the performance of the recommendation system and the importance of the Deep Reinforcement Learning (DRL) agent. GroupN is the group without the AI system and GroupAI refers to the group with the aid of the AI recommendation system. There are a total of 47 participants with 23 participants in GroupN and 25 participants in GroupAI. The data was normalized using a MinMax scalar with values ranging from 0.0 to 1.0. The normalization was applied across grouping by scenario to minimize the effect of the scenarios on all the participants from both groups. As the focus of this paper was not to analyze the differences between the scenarios, therefore, the normalized values were then averaged for the scenarios per participant so that each participant had now a single data point for each variable. 

In this section, we cross-correlate different performance measures to evaluate the overall performance between groups as well as within participants in GroupAI for their different choices of preferences. Furthermore, this cross-analysis will help in identifying the relationship of the process, alarm, and HMI logs as well as the DRL vs AI response with other important factors such as situational awareness, task load, trust in the system, and overall performance, etc to be able to validate the real-time prediction of operator state.

    \subsection{Human Performance vs System Performance}
\begin{figure}[!t]
        \centering
            \includegraphics[width=0.95\linewidth]{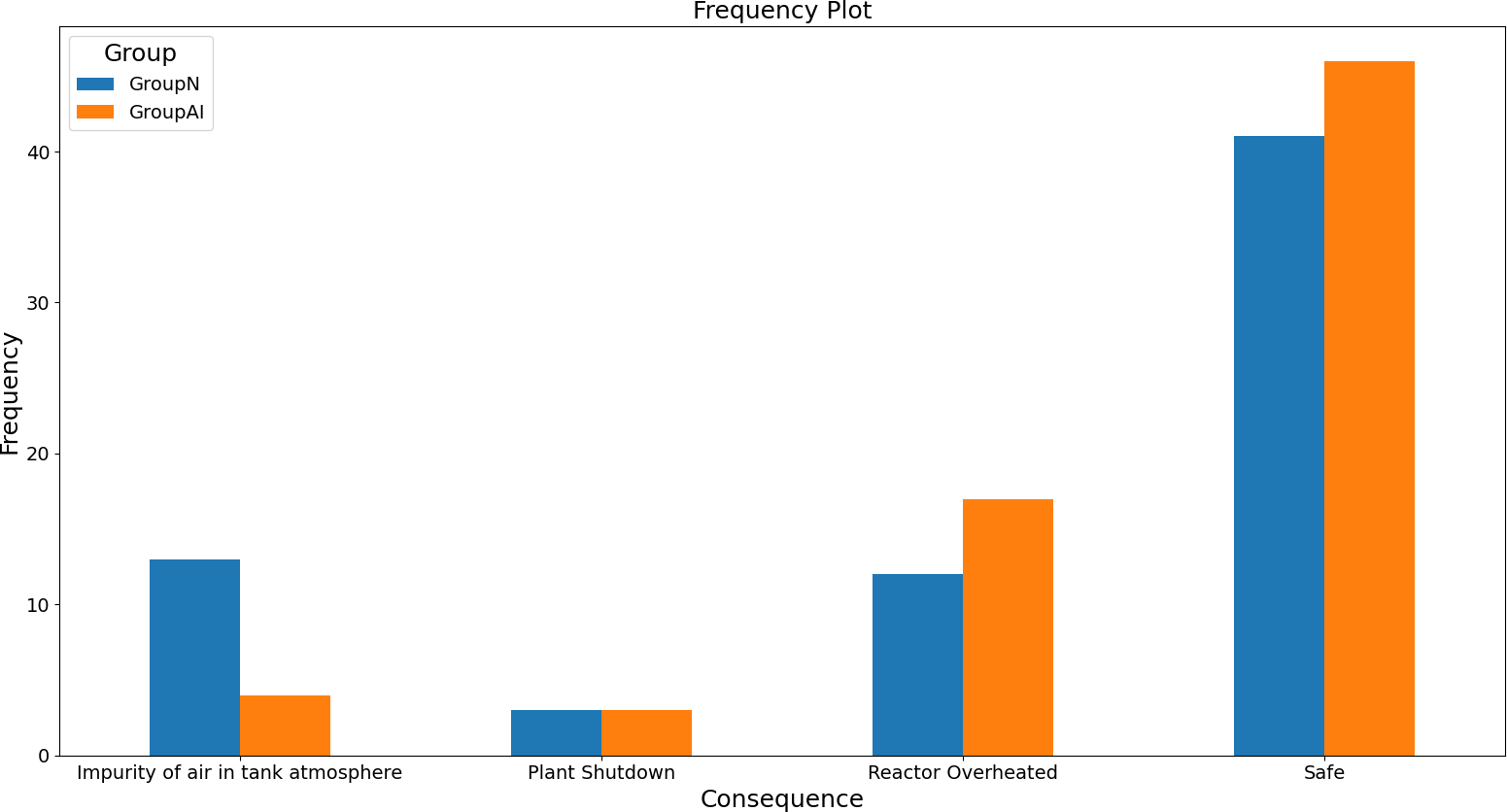}
            \caption{Frequency plot of consequences}
            \label{fig: freq_conseq}
    \end{figure}

    The metrics for analysis were divided in such a way that it can capture the performance of the human as the operator and decision-making agent as well as the performance of the system that would be in the case of GroupAI the AI and DRL recommendation system. From the frequency plot of the overall consequences incurred during the simulation as shown in \cref{fig: freq_conseq}, it can be observed that with the AI system in alarm overflow situations (a problem in the reactor), it becomes difficult to recover the plant and to focus on the recommendation of the DRL precisely. 

    \subsection{GroupN and GroupAI}

        \subsubsection{Heat Map}

\begin{figure}[!t]
            \centering
            \subfloat[]{
                \includegraphics[width=0.95\linewidth]{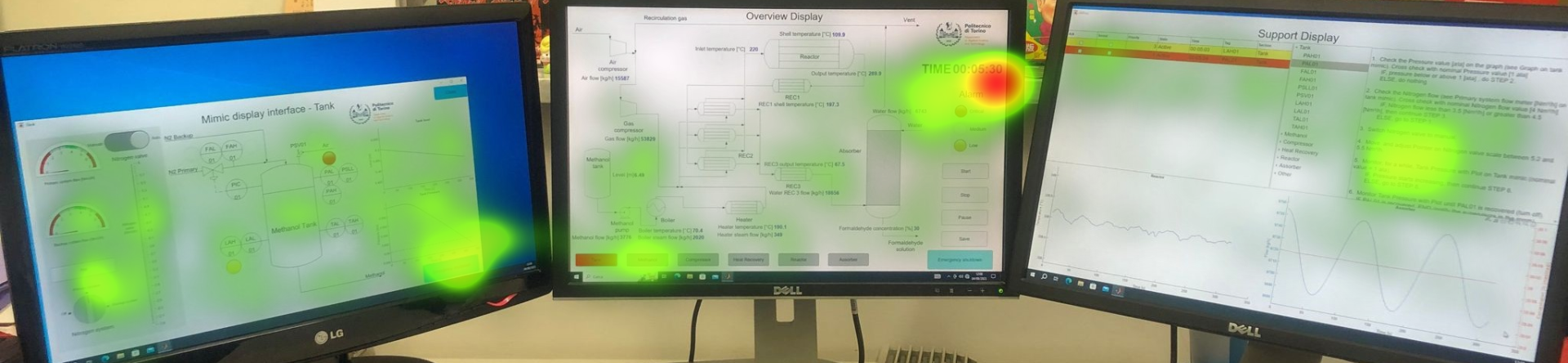}
                \label{fig:subfig1}}
                \\
            \subfloat[]{
                \includegraphics[width=0.95\linewidth]{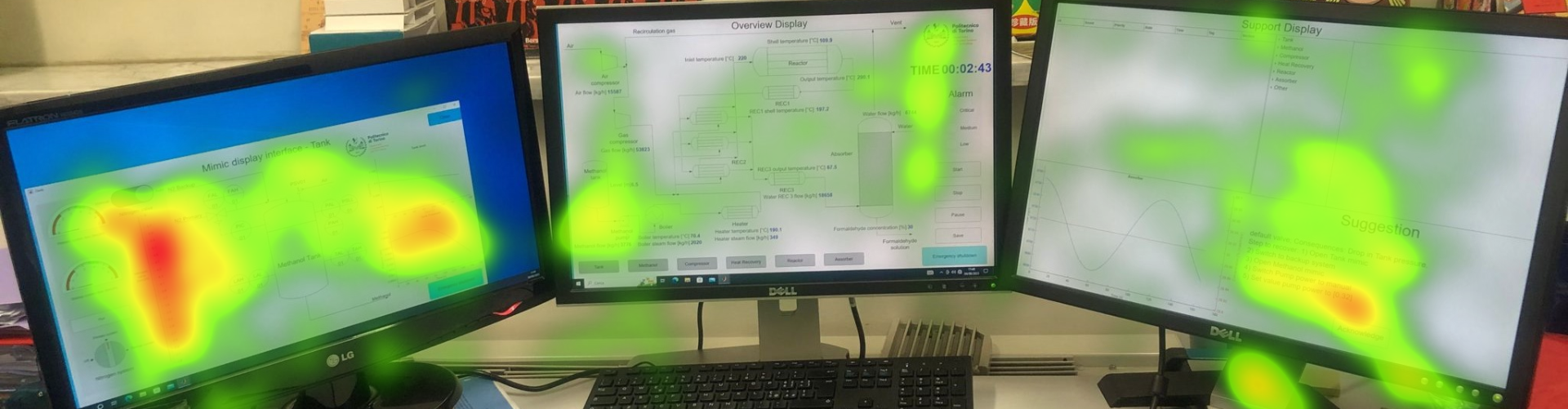}
                \label{fig:subfig2}}
            \caption{Heat maps for participants in (a) GroupN (without AI support) and (b) GroupAI (with AI support).}
            \label{fig: heat-map-group}
        \end{figure}

        \Cref{fig: heat-map-group} shows the heat map of the participants' data (17 participants) for scenario 1 and the group with and without AI support for the entire recording. 
            
\begin{figure}[!t]
                \centering
                \subfloat[]{
                    \includegraphics[width=0.95\linewidth]{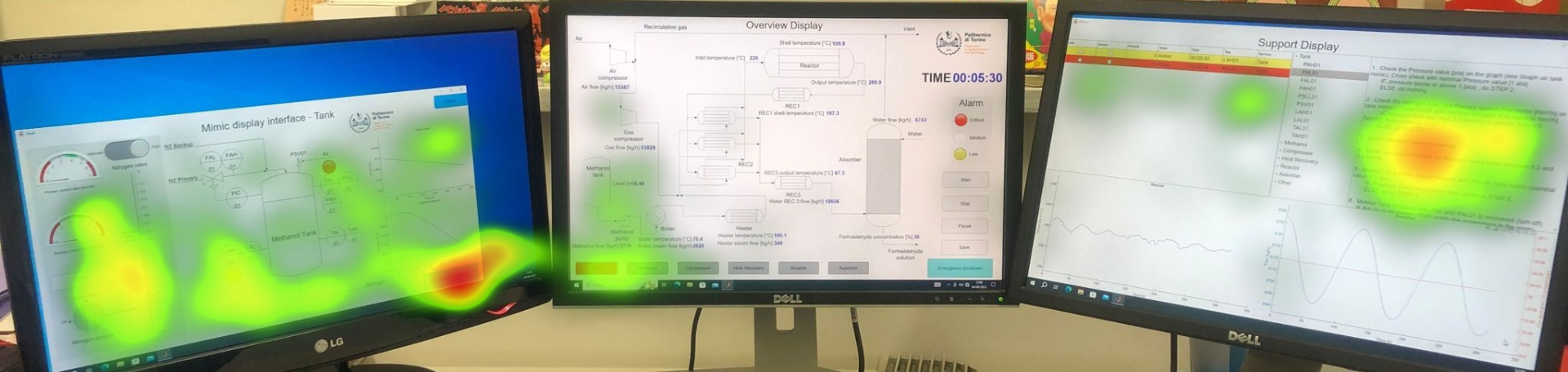}
                    \label{fig:subfig1}}
                    \\
                \subfloat[]{
                    \includegraphics[width=0.95\linewidth]{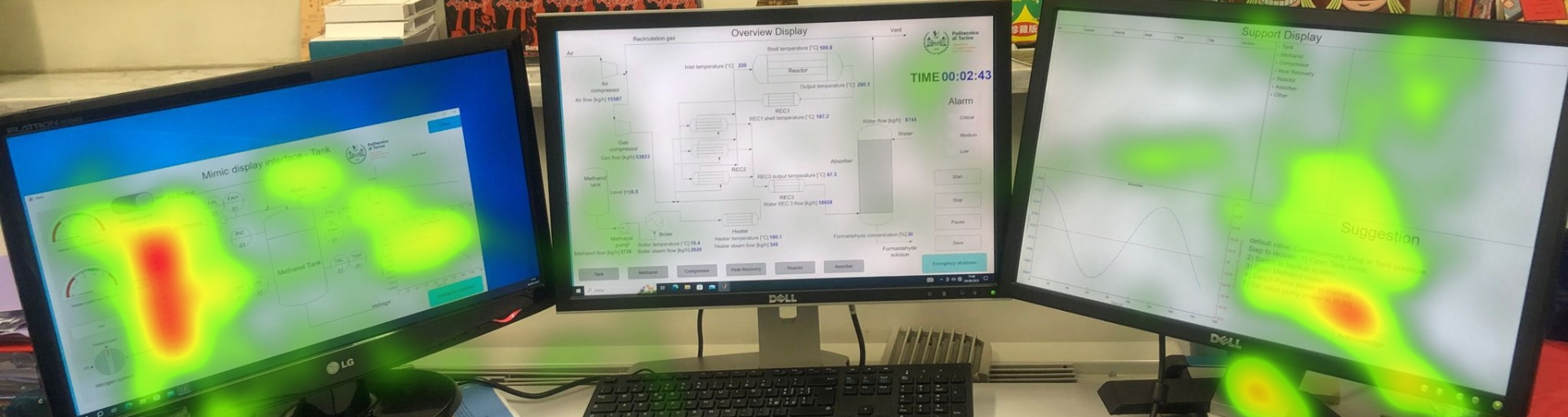}
                    \label{fig:subfig2}}
                \caption{Heat maps for participants in (a) GroupN (without AI support) and (b) GroupAI (with AI support) within the Time of Interest (TOI) of critical alarms.}
                \label{fig: heat-map-group-toi}
            \end{figure}

            \paragraph{Time of Interest: Critical Alarm}
            \Cref{fig: heat-map-group-toi} shows the heat map for the specific Time of Interest (TOI) of critical alarm. As can be seen for GroupAI (with AI support), people focus more on the AI suggestion rather than the procedure, and the Deep Reinforcement Learning (DRL) analog suggestion value and the control task to retrieve the plant to normal condition as soon as possible as compared to GroupN (without AI support). 
            
        \vspace{0.1in}
        \subsubsection{Radar Plot}
\begin{figure}[!t]
            \centering
                \includegraphics[width=1.0\linewidth]{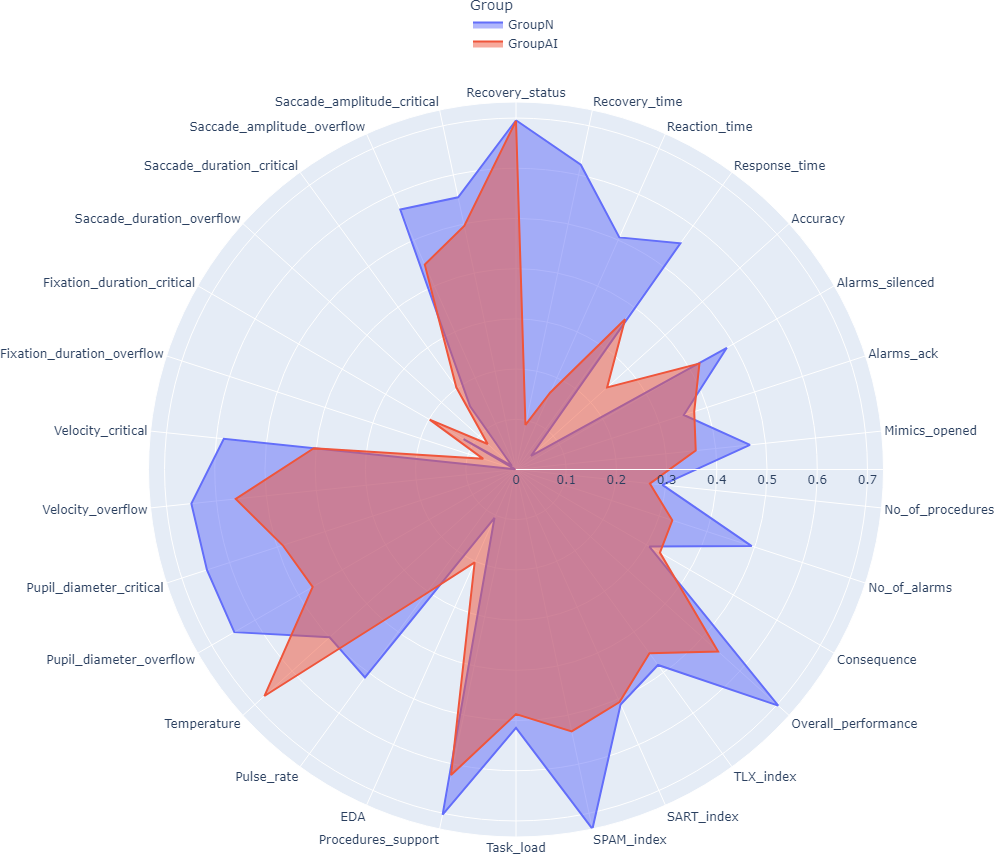}
                \caption{Radar plot between GroupN (blue) and GroupAI (red)}
                \label{fig: radar_group}
        \end{figure}

        As seen from \cref{fig: radar_group}, we can cross-evaluate several different factors.

            \paragraph{Recovery Status and Consequence} These factors are nearly the same for both groups, indicating the performance based on the overall recovery of the plant does not vary.

            \paragraph{Recovery, Reaction, and Response Time} All of these factors are lower in GroupAI as compared to GroupN indicating the performance in terms of reduced information processing time.

            \paragraph{Accuracy} It is higher for GroupAI which shows the ability to follow the operational guidelines accurately.

            \paragraph{Mimics Opened and Number of Alarms} Both of these factors were observed to be lower in the case of GroupAI, indicating the preventive and proactive ability provided by the AI system to the participants and minimizing the task-solving complexities.

            \paragraph{Overall Performance} It was measured greater for GroupN as compared to GroupAI which was evaluated based on how correctly the overall system was stabilized.
            
            \paragraph{Task Load, SART, TLX, and SPAM Index} All the questionnaires indicated lower values for GroupAI compared to GroupN. Despite GroupAI having a reduced task load, their situational awareness is effectively lower.

            \paragraph{EDA, Temperature, and Pulse Rate} In examining these measures, it was observed that GroupAI exhibited higher levels of EDA and temperature, while their pulse rate was lower. The autonomic nervous system (ANS) is responsible for regulating various physiological responses \cite{ghiasi2020assessing}, including heart rate and electrodermal activity. While EDA and temperature may indicate increased sympathetic nervous system activity, heart rate is also influenced by both sympathetic and parasympathetic branches. The group with higher EDA and temperature may be experiencing a specific pattern of autonomic response that involves decreased parasympathetic activity, leading to lower heart rates.

            \paragraph{Eye Tracking Metrics}
            GroupAI exhibited longer fixation and saccade durations, smaller and slower eye movements (lower saccade amplitude and velocity), and reduced pupil diameter compared to GroupN. These findings suggest that GroupAI employs a more focused and deliberate cognitive processing strategy, potentially perceiving the task as less demanding or engaging than GroupN. 

        \vspace{0.1in}        
        \subsubsection{Correlation Graph}
        \label{sec:corr_anal}
        Correlation analysis was conducted on participant data from both groups, followed by filtering for visualizations that exceeded a specified threshold (0.4) as shown in \cref{fig:correlation_group}. The subsequent analysis revealed several noteworthy characteristics: 
        
        \begin{figure*}[!t]
            \centering
                \includegraphics[width=1.0\linewidth]{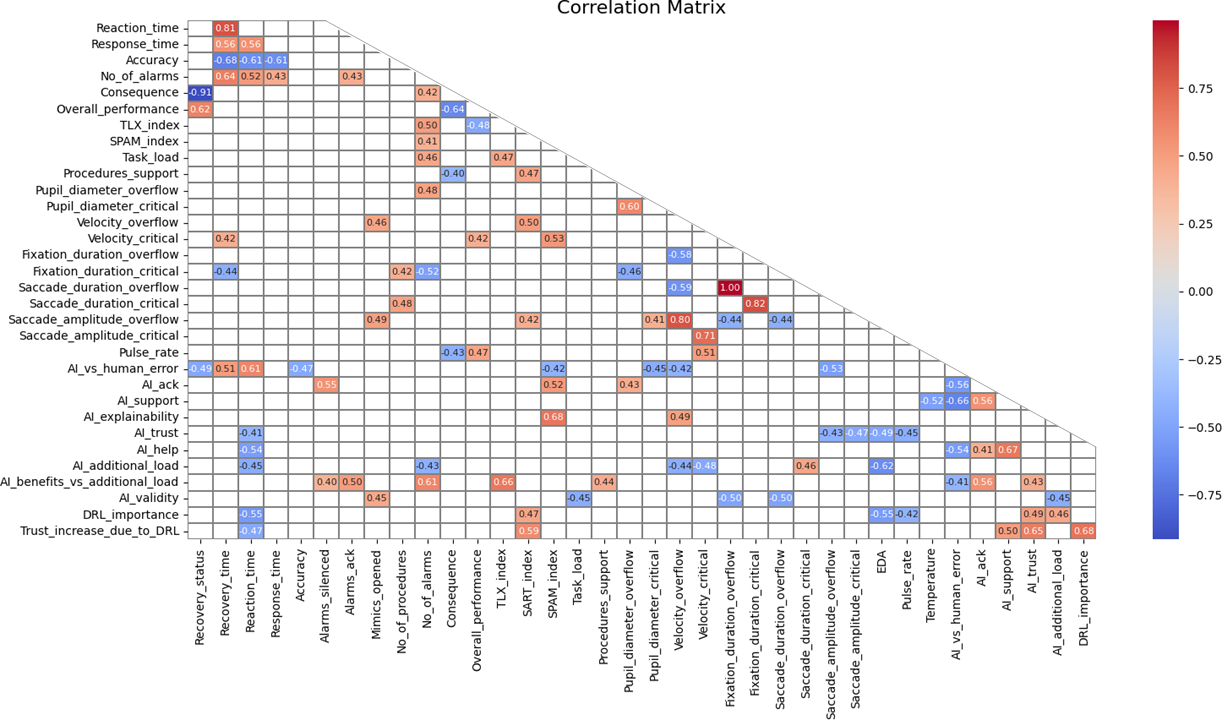}
                \caption{Radar plot between GroupN (blue) and GroupAI (red)}
                \label{fig:correlation_group}
        \end{figure*}

            \paragraph{Recovery Time} Reaction time, response time, and number of alarms are directly correlated with the recovery time and inversely correlated with the accuracy. 

            \paragraph{Number of Alarms} It is directly correlated with the TLX index, task load, SPAM index, average pupil diameter, and consequence.

            \paragraph{Procedures Support} It is directly correlated with the SART index and inversely correlated with the consequence.

            \paragraph{Veloctity} Velocity is directly correlated with recovery time, mimics opened, overall performance, SART, and SPAM index.

            \paragraph{Fixation Duration} It shows an inverse correlation between recovery time and the number of alarms. It is directly correlated with the number of procedures opened.

            \paragraph{AI (DRL) vs Human Response} This metric shows a direct correlation with recovery and reaction time. On the other hand, it is inversely correlated with recovery status, accuracy, SPAM index, pupil diameter, velocity, saccade amplitude, AI acknowledgment, AI support, and AI help. As the deviation from the suggested value increases, there is a corresponding rise in the time needed to attain process stabilization, leading to a decline in recovery performance. The escalation of errors also signifies a reduction in situational awareness (SPAM). Additionally, it is noteworthy that a lower error in adhering to the DRL recommendations correlates with higher participant ratings for the help and support offered by the overall AI system, along with an increased frequency of acknowledging and accepting the AI suggestions. Furthermore, higher error also results in lower pupil diameter, saccade velocity, and amplitude which may suggest reduced arousal, slower eye movements, and more restricted visual processing.

            \paragraph{AI Acknowledgement} The degree to which a participant concentrates on adhering to the AI's recommendation directly correlates with heightened situational awareness (SPAM), enhanced perception of AI assistance, and support.

            \paragraph{AI Explanability} A heightened understanding of the AI system, particularly in terms of explainability for the participant, yields increased situational awareness (SPAM). 

            \paragraph{AI Trust} The participant's response to the trust in the AI system is inversely proportional to the reaction time, saccade amplitude, EDA, and pulse rate. It suggests that as individuals develop a higher level of trust in the AI system, they exhibit faster reaction times, reduced eye movements, decreased electrodermal activity, and a lower pulse rate. These observed physiological and behavioral changes likely signify a more relaxed and less stressed response in participants who trust the AI system.

            \paragraph{AI as Additional Work Load} Participants tend to rate higher scores for perceiving the AI system as an additional workload. Surprisingly, this perception is associated with a decrease in reaction time, number of alarms, velocity, and electrodermal activity (EDA), which are typically indicative of lower workload. However, there is a positive correlation with saccade duration, suggesting that participants may spend more time making eye movements when perceiving the AI system as a higher workload. The contradictory nature of these findings suggests a need for further investigation to understand if there are any unexpected factors influencing these outcomes or if there might be nuances in how participants perceive and respond to the AI system's workload.

            \paragraph{AI Benefits vs Additional Workload} It shows a positive correlation with alarms silenced, alarms acknowledgments, number of alarms, TLX index, procedures support, AI acknowledgment, and AI trust. With AI vs human error, a negative correlation can be observed. The favorable view that participants hold toward the benefits of AI vs the additional workload is linked to heightened trust, following AI suggestions by acknowledging it. The inverse relationship with human error implies pervasive confidence in AI as a dependable and error-reducing solution.
            \paragraph{AI Validity} The participants who were more likely to consider the importance of validating the AI system opened more mimics and had a higher fixation and saccade duration. Longer saccades may indicate increased cognitive load or visual search difficulty, while prolonged fixations suggest in-depth processing, interest, or confusion \cite{unema2005time, fadardi2022post, mahanama2022eye, stuyven2000effect}.
            \paragraph{DRL Importance} We observed a positive correlation between the importance of DRL as a response by participants to SART, AI trust, and AI as an additional workload. Further, it was negatively correlated with reaction time, EDA, and pulse rate. This implies that as the perceived importance of DRL increases, participants tend to respond more to the task, trust AI more, and perceive AI as a greater workload, but also exhibit faster reaction times and reduced physiological responses.            
            \paragraph{Trust Increase due to DRL} It is directly correlated with SART, AI support, AI trust, and DRL importance, and inversely correlated with the reaction time. It indicates the increase in overall AI support and trust with the DRL's analog value and faster reaction responses.
      
        \vspace{0.1in}
        \subsubsection{Factor Analysis}

\begin{figure}[!t]
        \centering
            \includegraphics[width=0.9\linewidth]{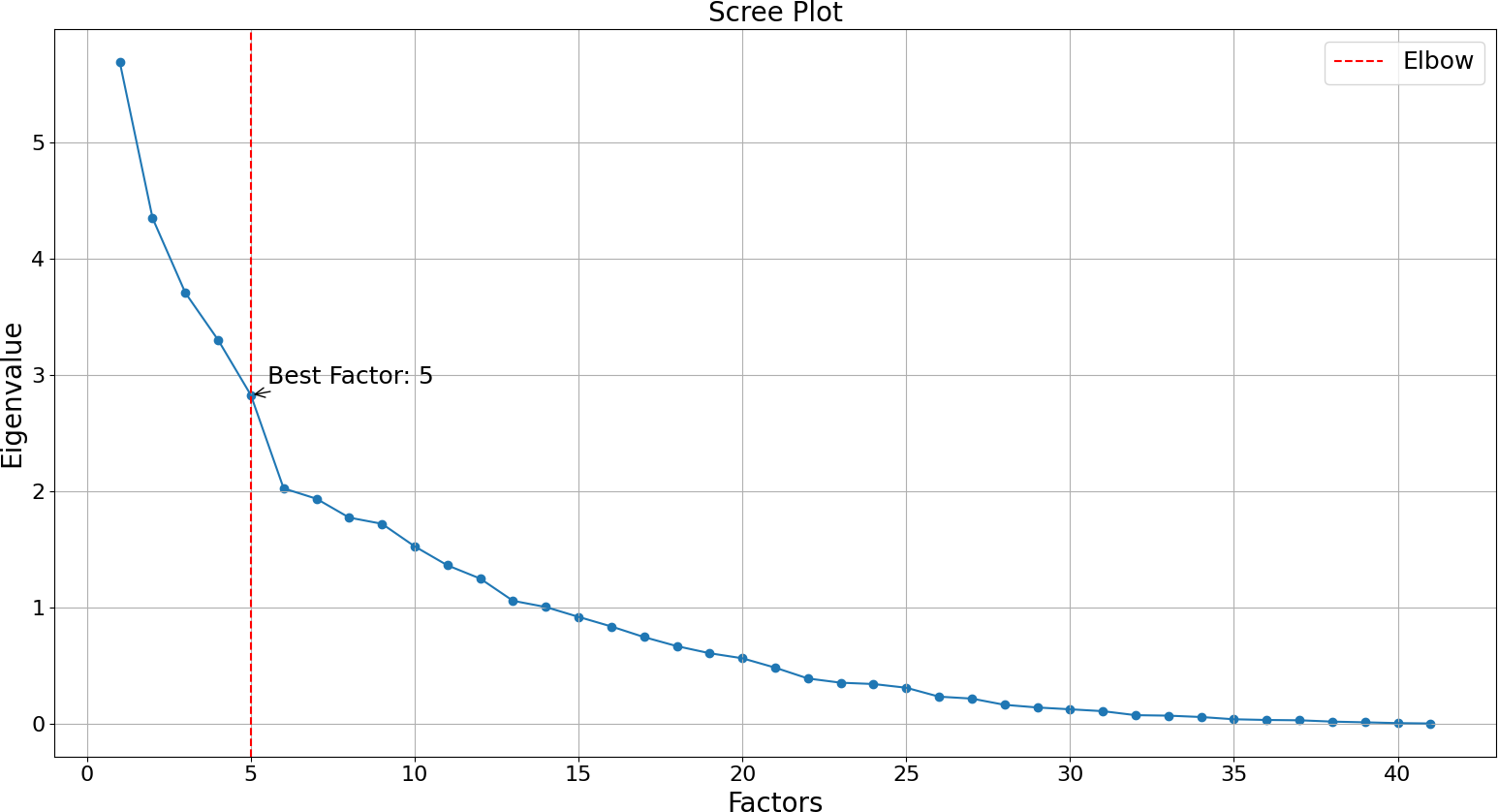}
            \caption{Scree plot and the best possible value of the factors.}
            \label{fig: factor_analysis}
        \end{figure}

\begin{figure}[!t]
        \centering
            \includegraphics[width=0.9\linewidth]{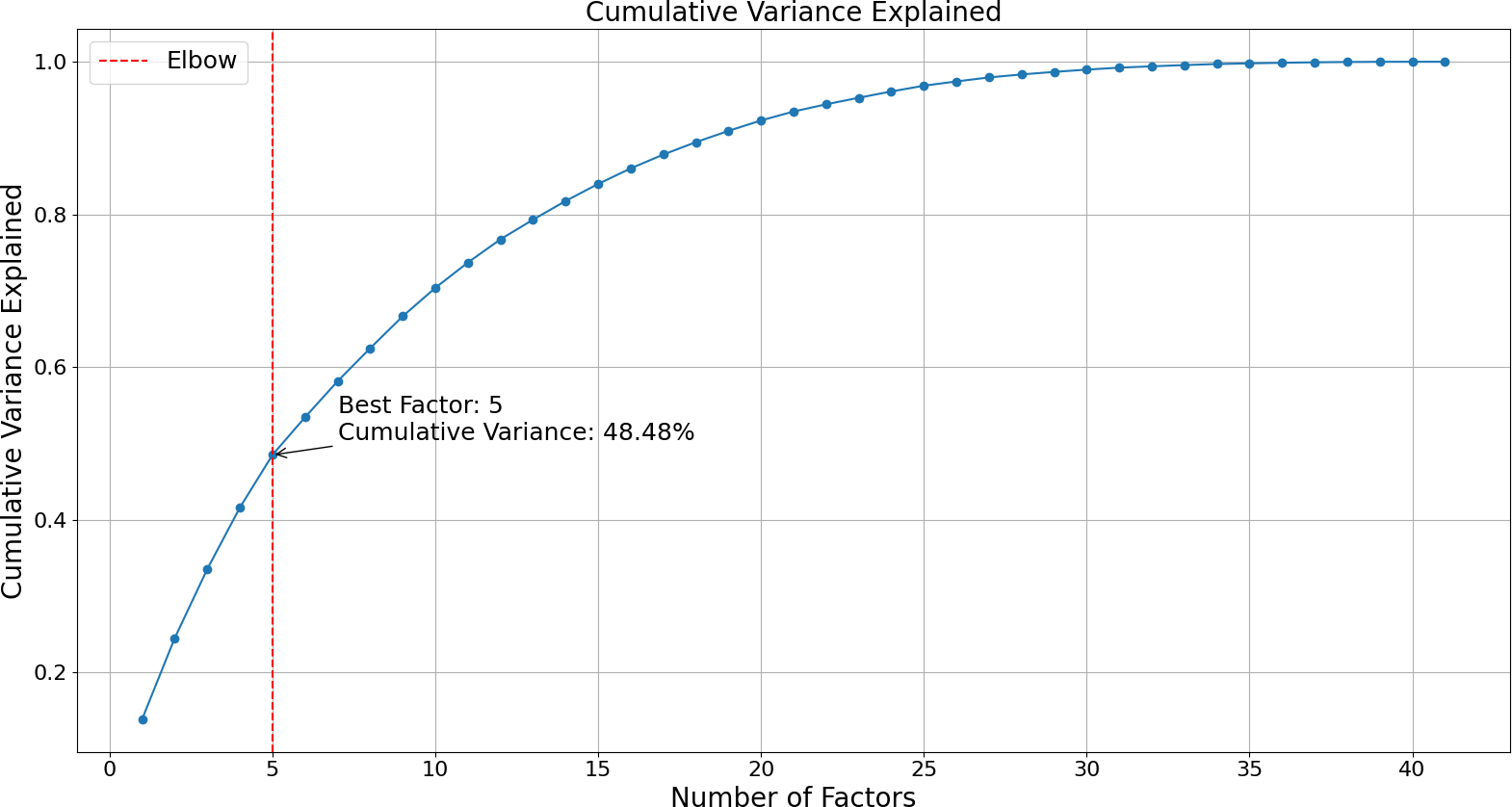}
            \caption{Cumulative explained variance per each factor.}
            \label{fig: cum_var}
        \end{figure}

        Factor analysis is a statistical method used to explore relationships among observed variables by postulating underlying latent factors. In the factor model equation, observed variables (\(X\)) are expressed as linear combinations of latent factors (\(F\)) and unique factors (\(U\)), represented by the factor loading matrix (\(\Lambda\)). The covariance matrix (\(\Sigma\)) can be decomposed into the product of \(\Lambda\) and its transpose, plus a diagonal matrix (\(\Psi\)) of unique variances. To determine the optimal number of factors, a scree plot is generated by plotting the eigenvalues of the covariance matrix, and the elbow joint, indicative of the optimal number of factors, is identified by analyzing the second difference in the cumulative variance explained with a threshold of $0.05$ as shown in \cref{fig: factor_analysis,fig: cum_var}. The following statistical modeling can be decomposed as follows:
        
        \begin{equation}
        \begin{gathered}
            X = \Lambda F + U \\
            \Sigma = \Lambda \Lambda^T + \Psi \\
            \text{Cumulative Variance} = \frac{\sum_{i=1}^{n} \text{Eigenvalue}_i}{\sum_{i=1}^{n} \text{Eigenvalues}} \\
            \text{Second Differences} = \text{diff}(\text{diff}(\text{Cumulative Variance})) \\
            \text{Elbow Point} = \text{argmax}(\text{Second Differences})
        \end{gathered}
        \end{equation}

            \paragraph{Interpretation of the Factors}
            The factor analysis conducted on the dataset identified a crucial factor, chosen at the point where the cumulative variance explained reached 50\% (benchmark). This factor, along with the preceding factors, collectively explains approximately half of the total variance in the data. The analysis successfully achieved dimensionality reduction, simplifying the interpretation of the complex dataset. While the 50\% cumulative variance is a meaningful balance between model simplicity and explanatory power, further exploration of additional factors or alternative models is advisable for a more detailed understanding of the underlying structure. The interpretation of factors is context-dependent, emphasizing the significance of findings concerning the specific goals and nature of the variables in the analysis. The factor analysis performed here was the combination of all the variables and the data points of all the participants involved in both groups and also included some missing variables. Therefore, further analysis is required to break down the data points in terms of specific context as well as scenario-wise for detailed exploration. Overall, the factor analysis provided valuable insights into the relationships among observed variables, facilitating a concise and interpretable representation of the dataset as shown in \cref{tab: fact_anal}. A factor loading importance threshold of 0.4 was chosen which was identified iteratively to capsulate the valuable factors involved in the experiment.
            
             \begin{table*}[htbp]
                \fontsize{5.5pt}{5.5pt}\selectfont
                \caption{Factor Analysis}
                \centering
                    \begin{tabular}{|l||l||l||l||l|}
                    \hline
                    \multicolumn{1}{|c||}{\textbf{\multirow{3}{*}[2ex]{Factor1}}} & \multicolumn{1}{c||}{\textbf{\multirow{3}{*}[2ex]{Factor2}}} & \multicolumn{1}{c||}{\textbf{\multirow{3}{*}[2ex]{Factor3}}} & \multicolumn{1}{c||}{\textbf{\multirow{3}{*}[2ex]{Factor4}}} & \multicolumn{1}{c|}{\textbf{\multirow{3}{*}[2ex]{Factor5}}} \\ 
                    \multicolumn{1}{|c||}{\textbf{(Situational Awareness \& Performance)}} & \multicolumn{1}{c||}{\textbf{(System Status \& Task Impact)}} & \multicolumn{1}{c||}{\textbf{(Human-AI Interaction \& Trust)}} & \multicolumn{1}{c||}{\textbf{(Oculomotor Behavior: Cognitive Load \& Attention}} & \multicolumn{1}{c|}{\textbf{(AI and DRL Perception)}} \\ \hline
                                                 &                       &                               &                     &                                  \\

                        Recovery\_time                         & Recovery\_status                      & Accuracy                              & Velocity\_overflow                    & EDA                                 \\
                        Reaction\_time                         & No\_of\_alarms                        & AI\_vs\_human\_error                  & Fixation\_duration\_overflow          & AI\_trust                           \\
                        Response\_time                         & Consequence                           & AI\_ack                               & Saccade\_duration\_overflow           & AI\_additional\_load                \\
                        Accuracy                               & Overall\_performance                  & AI\_support                           & Saccade\_amplitude\_overflow          & DRL\_importance                     \\
                        No\_of\_alarms                         & TLX\_index                            & Trust\_increase\_due\_to\_DRL         &                                       & Trust\_increase\_due\_to\_DRL       \\
                        SPAM\_index                            &                                       &                                       &                                       &                                      \\
                        Pupil\_diameter\_overflow              &                                       &                                       &                                       &                                      \\
                        Velocity\_critical                     &                                       &                                       &                                       &                                      \\
                        Pulse\_rate                            &                                       &                                       &                                       &                                      \\ \hline
                    \end{tabular}
                \label{tab: fact_anal}
            \end{table*}

    \subsection{Within Participants (GroupAI)}

        \subsubsection{Heat Map}
\begin{figure}[!t]
            \centering
            \subfloat[]{\includegraphics[width=0.95\linewidth]{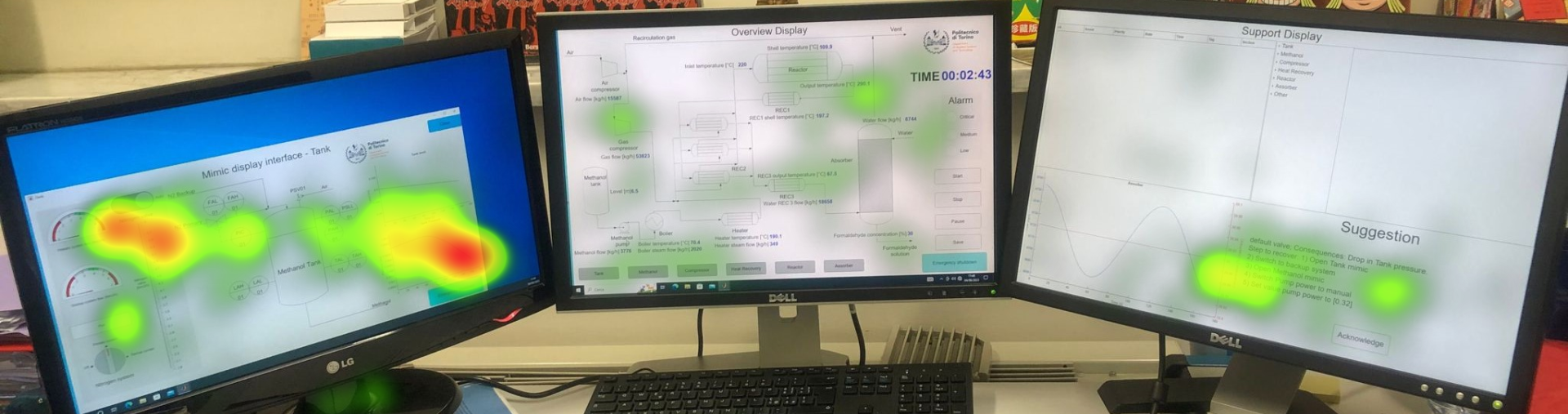}}
                \\
            \subfloat[]{\includegraphics[width=0.95\linewidth]{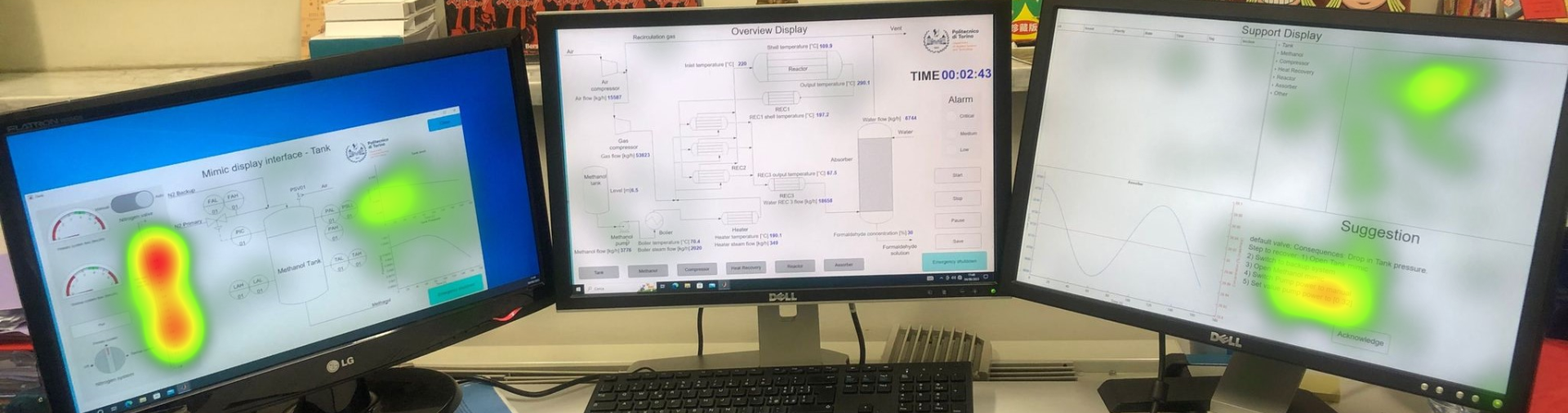}}
                \\
            \subfloat[]{ \includegraphics[width=0.95\linewidth]{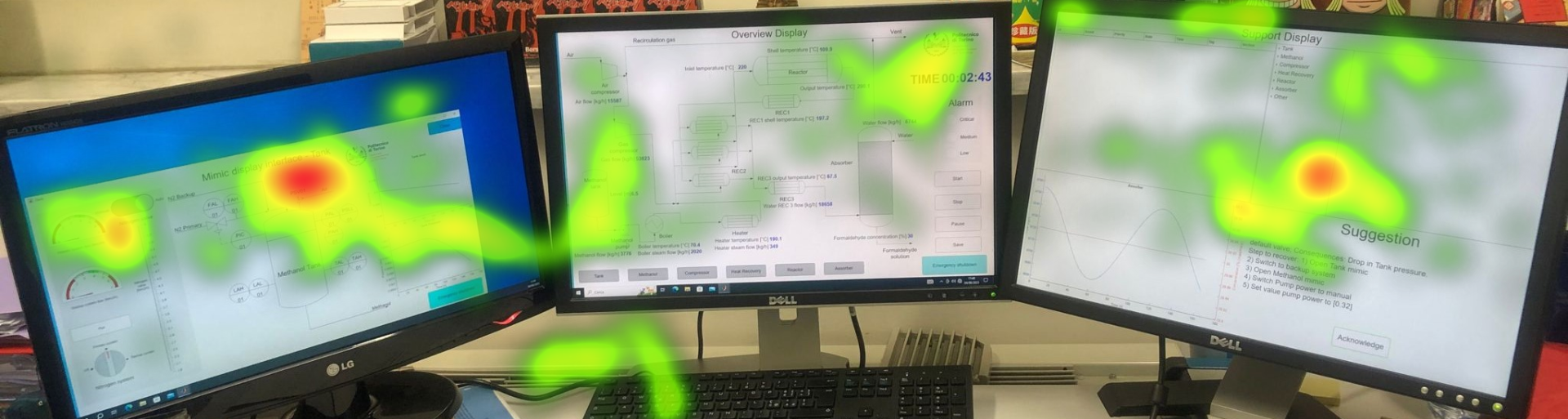}}
                \\
            \subfloat[]{\includegraphics[width=0.95\linewidth]{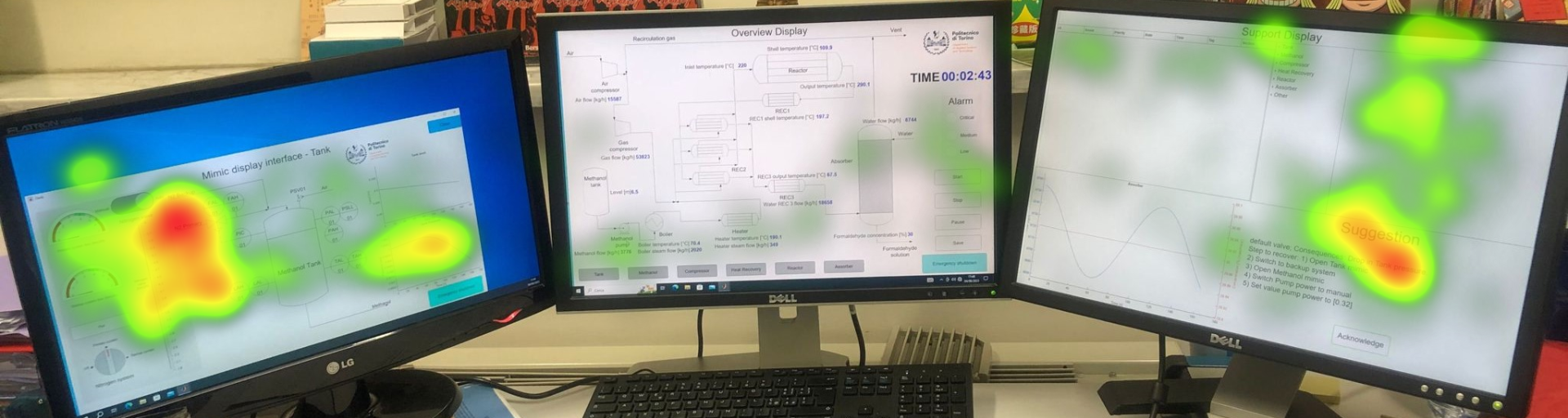}}
            \caption{Heat maps for (a) P21, (b) P24, (c) P30, and (d) P37 within the Time of Interest (TOI) of critical alarms.}
            \label{fig:heat-map-part}
        \end{figure}

        As depicted in the diverse heatmaps representing different participants in \cref{fig:heat-map-part}, distinct patterns emerge. Certain participants (a) exclusively adhere to the AI procedure, others (b) concurrently emphasize both AI and DRL, and (c) a subset concentrates solely on screen procedures. Meanwhile, (d) some validate their approaches through a combination of AI and screen procedures.

        \subsubsection{Selection of Participants}
        A subset of participants from GroupAI was chosen based on how they preferred to understand and adhere to the intervention strategy as shown in \cref{tab: part_sel}. This interpretation of the preference was extracted from the heatmap and other variables such as AI vs human response. Two participants were compared with each other on various factors to better understand the behavior patterns and performance. For each comparison, a participant was included who preferred to use both the AI + DRL support to be able to understand its characteristics with the other participant's behavioral pattern.
        
        \begin{table}[htbp]
            \caption{Selection of participants based on their preference of suggestion tool used during the experiment}
            \centering
            \begin{tabular}{|l||l|}
            \hline
            \multicolumn{1}{|c||}{\textbf{Participants}} & \multicolumn{1}{c|}{\textbf{Preference}} \\ \hline
            P21                                         & AI                                       \\
            P24                                         & AI + DRL                                  \\
            P30                                         & Procedure                                \\
            P32                                         & AI + DRL                                  \\
            P37                                         & AI + Procedure                           \\
            P96                                         & AI + DRL                                 \\ \hline
            \end{tabular}
            \label{tab: part_sel}
        \end{table}

        \vspace{0.1in}
        \subsubsection{P21 vs P24}
\begin{figure}[!t]
                \centering
                \subfloat[]{\includegraphics[width=0.48\linewidth]{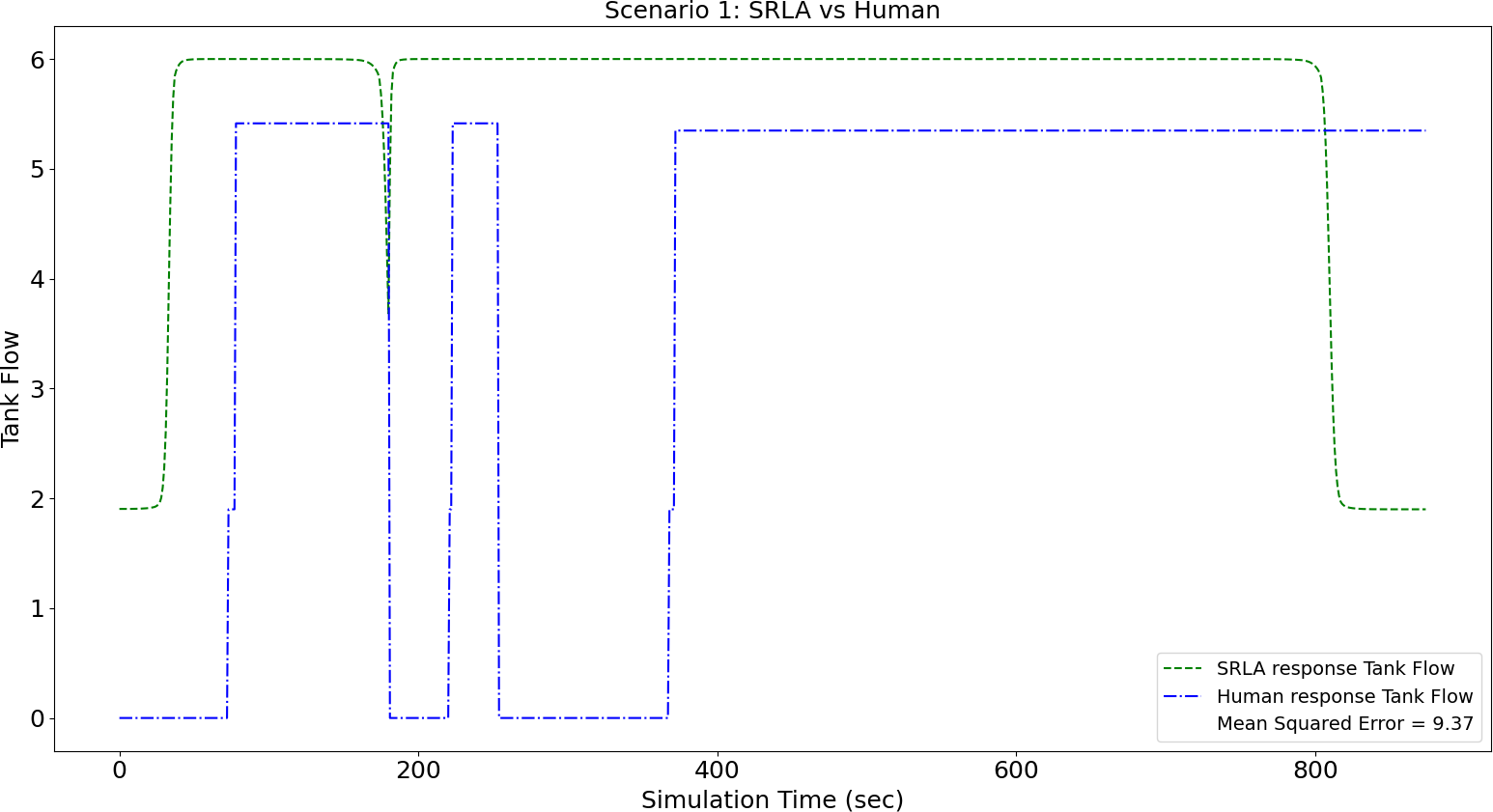}
                            \includegraphics[width=0.48\linewidth]{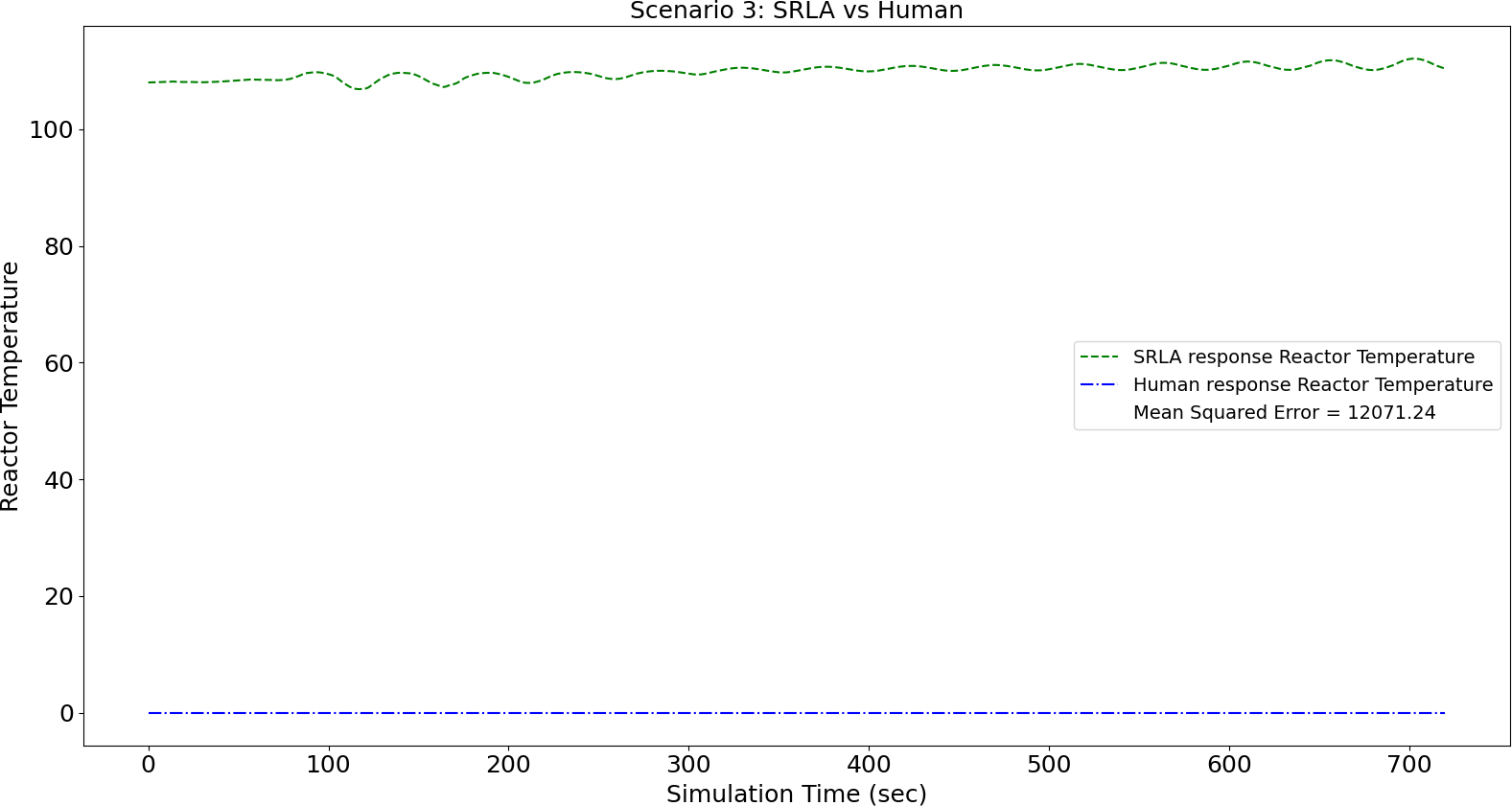}}
                    \\
                \subfloat[]{\includegraphics[width=0.48\linewidth]{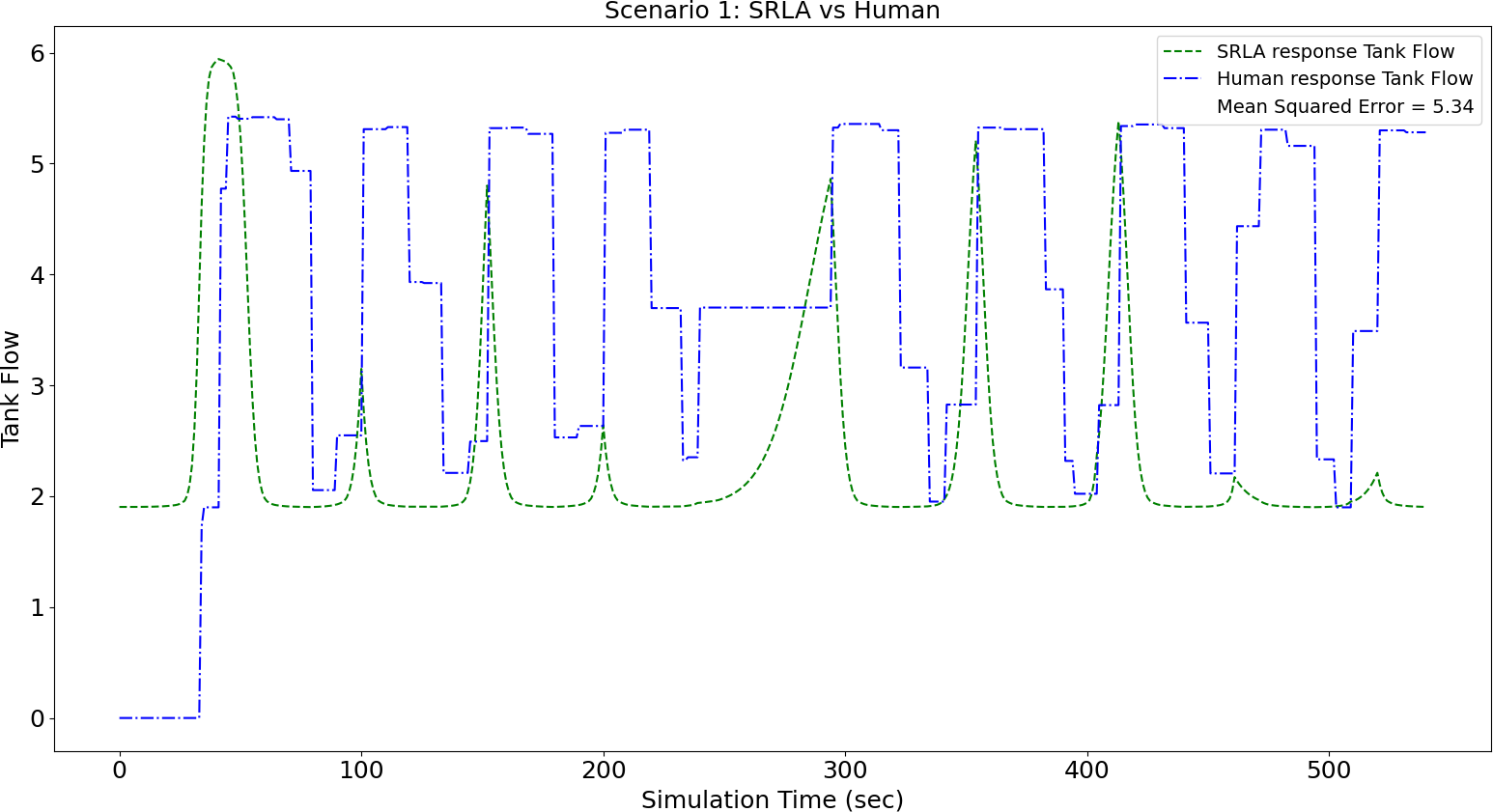}
                            \includegraphics[width=0.48\linewidth]{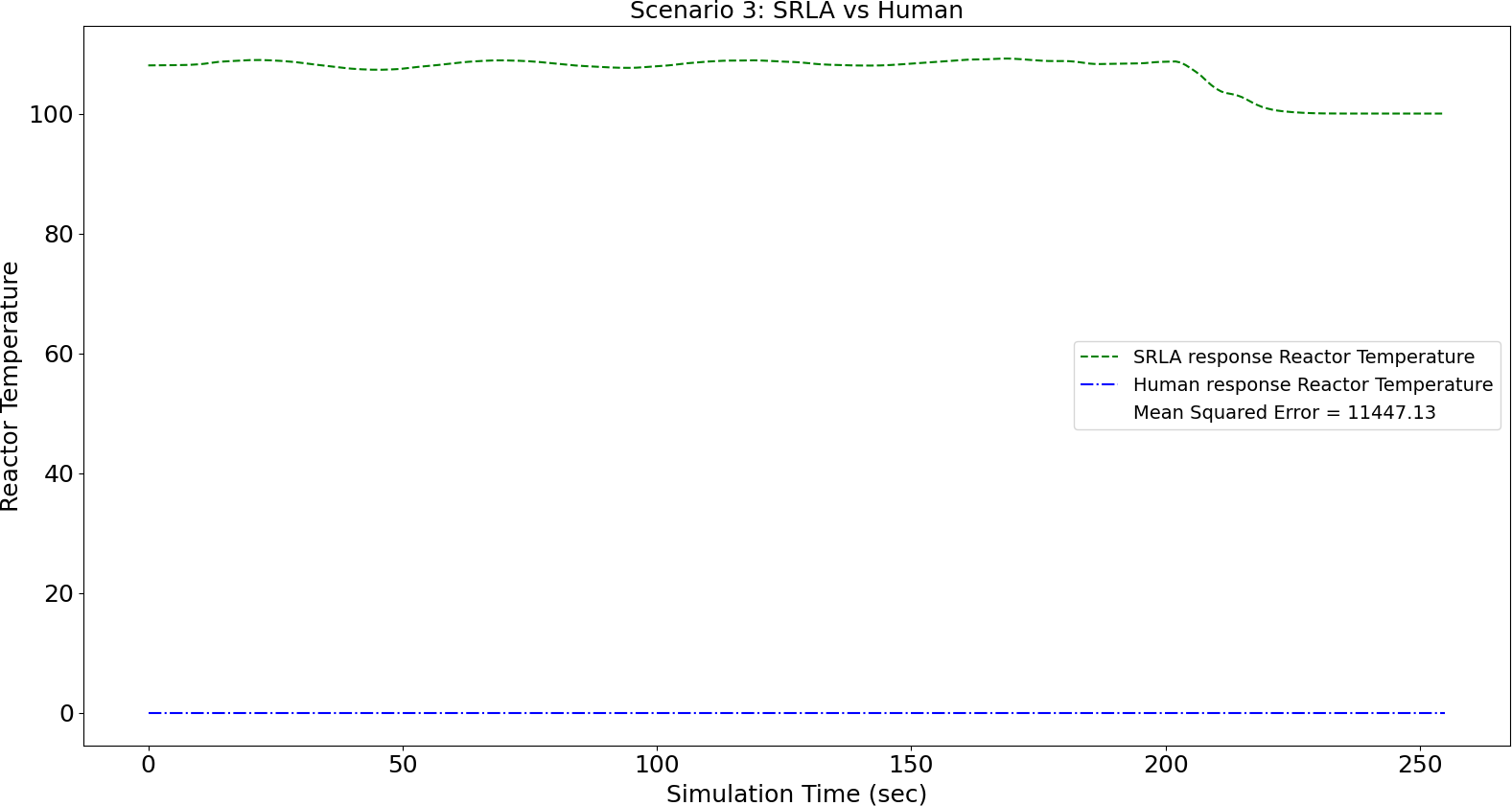}}
                \caption{SRLA vs human response of participant (a) P21 and (b) P24 within the Time of Interest (TOI) of critical alarms (scenario 1) and alarm overflow (scenario 3).}
                \label{fig: srla_vs_human_2124}
            \end{figure}

        This section comprises the comparative evaluation of P21 (preference: AI) and P24 (preference: AI and SRLA).
            \paragraph{AI (SRLA) vs Human Response}
            The comparison between the Specialized Reinforcement Learning Agent (SRLA) suggestion and the human response can be seen in \cref{fig: srla_vs_human_2124}, which shows P24 follows the SRLA suggestions better than P21, however, for the scenario of alarm overflow both have similar performance.
            
            \paragraph{Radar Plot}
            Participant P21 demonstrates superior physiological responses, better AI interaction, higher overall performance, quicker recovery from errors, and greater task accuracy compared to P24. P24, on the other hand, exhibits a higher cognitive workload, better situational awareness, and more significant consequences for actions. The comparison suggests that P21 excels in performance-related metrics, while P24 may have strengths in cognitive aspects and situational awareness. 
            
            \begin{figure}[!t]
                \centering
                    \includegraphics[width=0.95\linewidth]{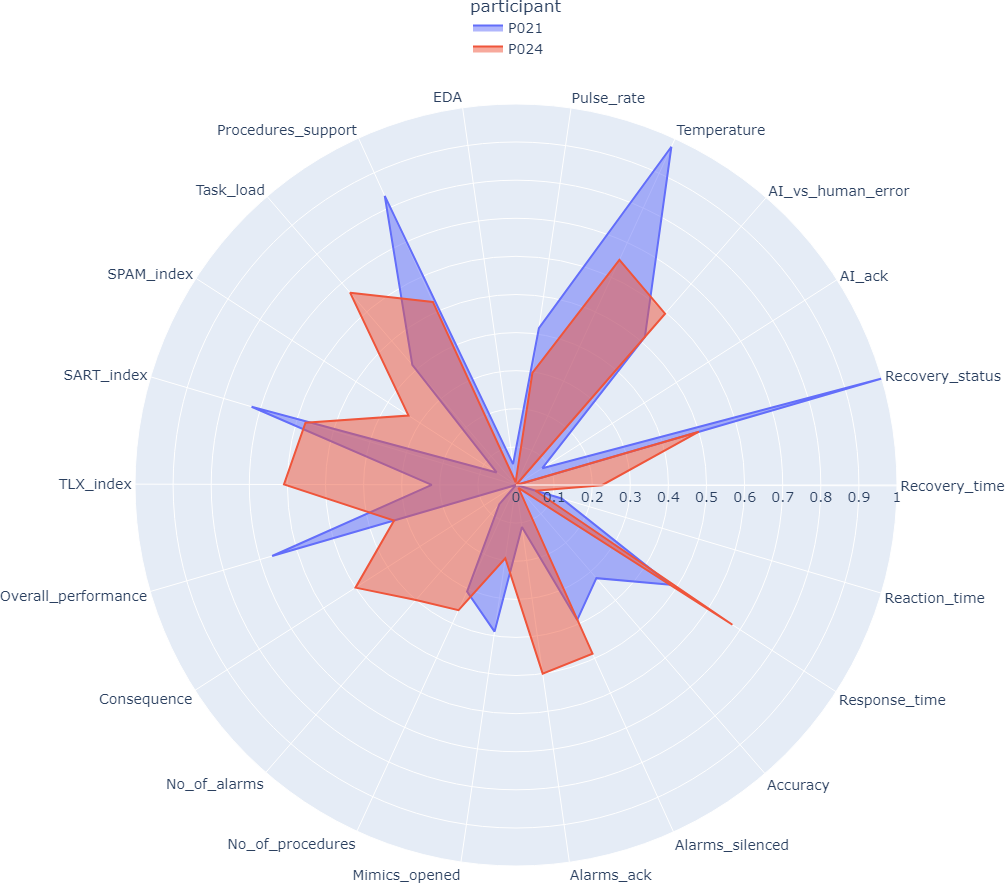}
                    \caption{Radar plot between participant 21 and 24}
                    \label{fig: radar_2124}
            \end{figure}            

        \vspace{0.1in}
        \subsubsection{P30 vs P32}
        This section comprises the comparative evaluation of P30 (preference: screen procedures) and P32 (preference: AI and SRLA).

\begin{figure}[!t]
                \centering
                \subfloat[]{\includegraphics[width=0.48\linewidth]{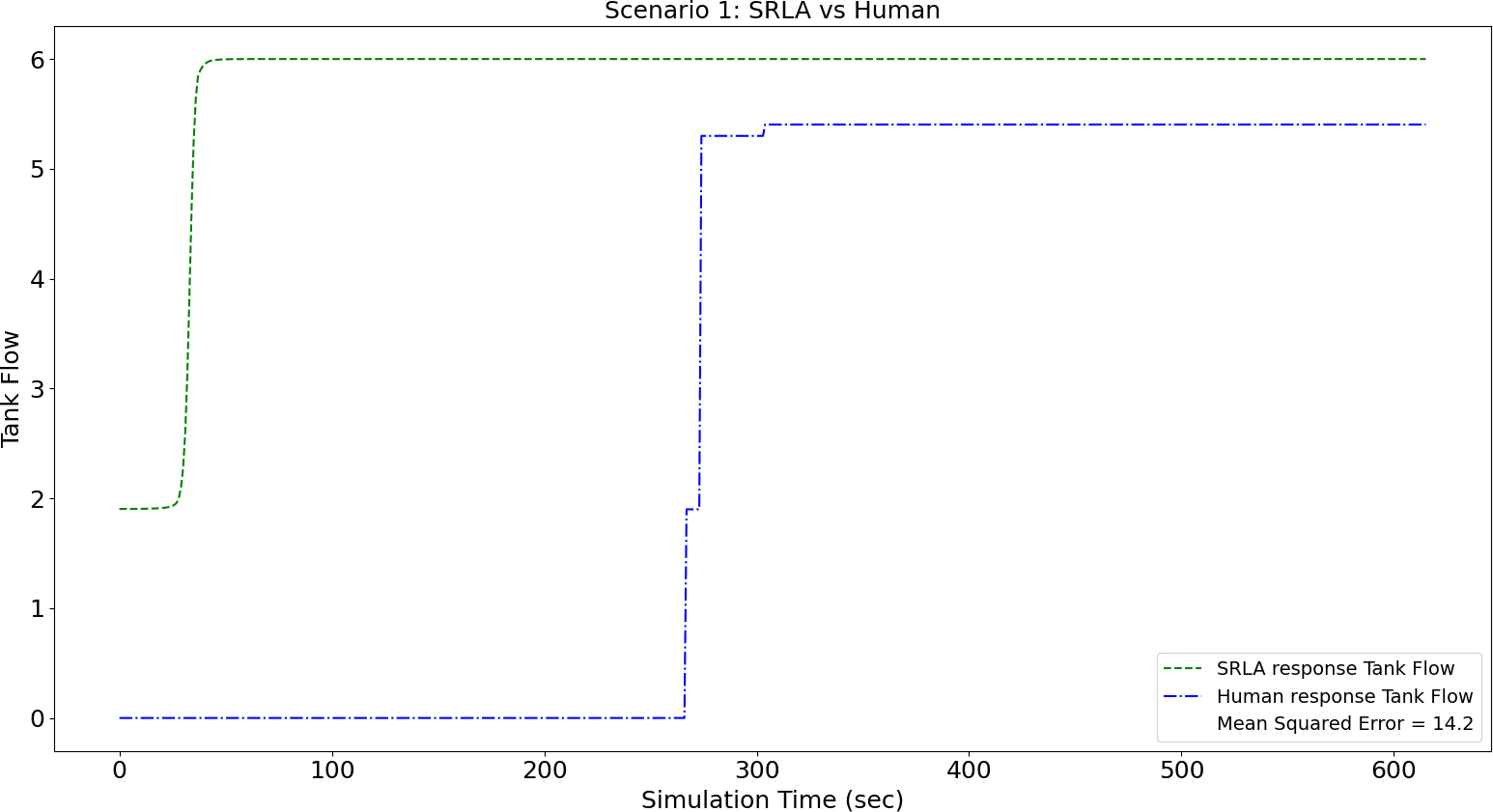}
                            \includegraphics[width=0.48\linewidth]{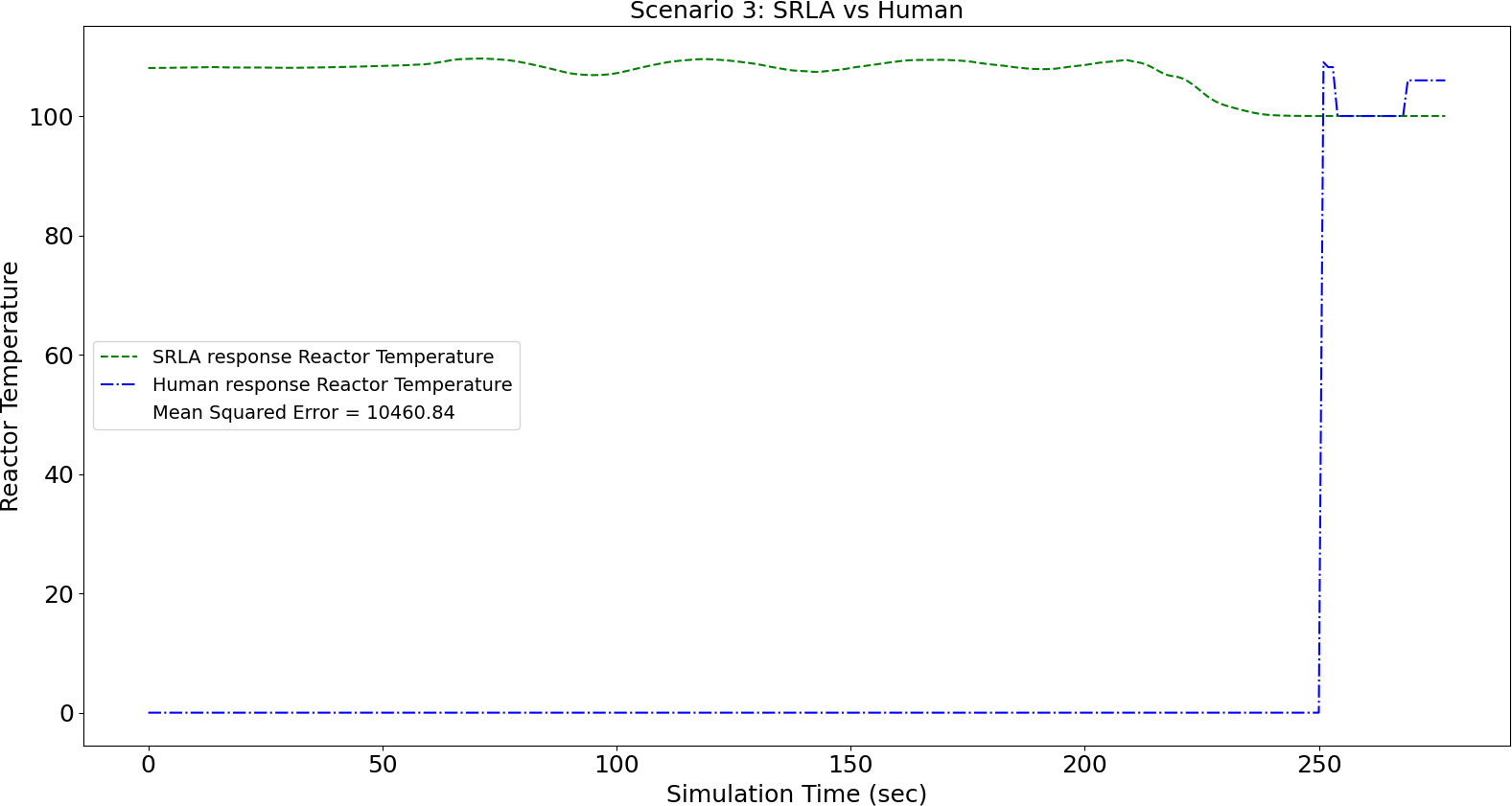}}
                    \\
                \subfloat[]{\includegraphics[width=0.48\linewidth]{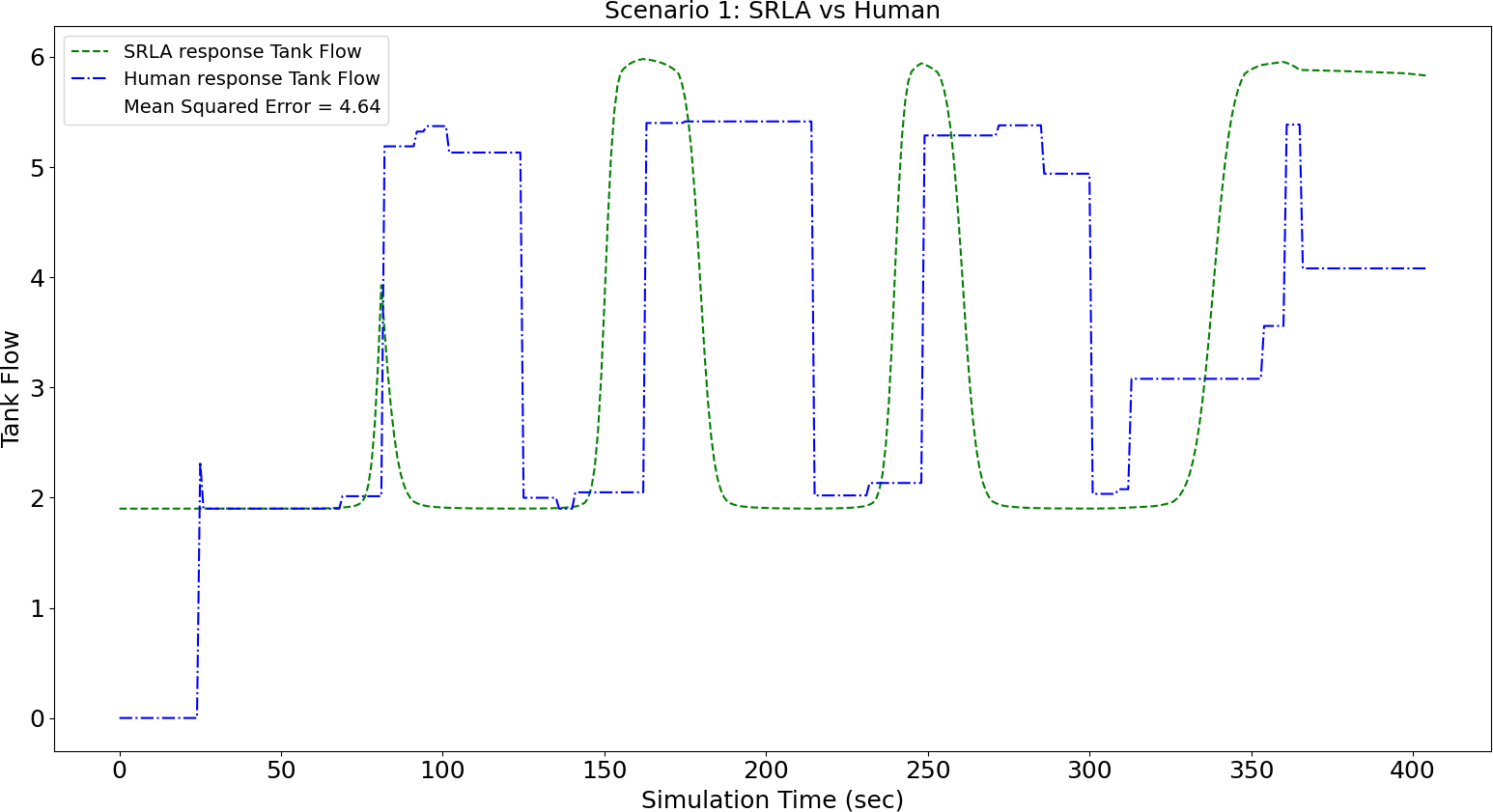}
                            \includegraphics[width=0.48\linewidth]{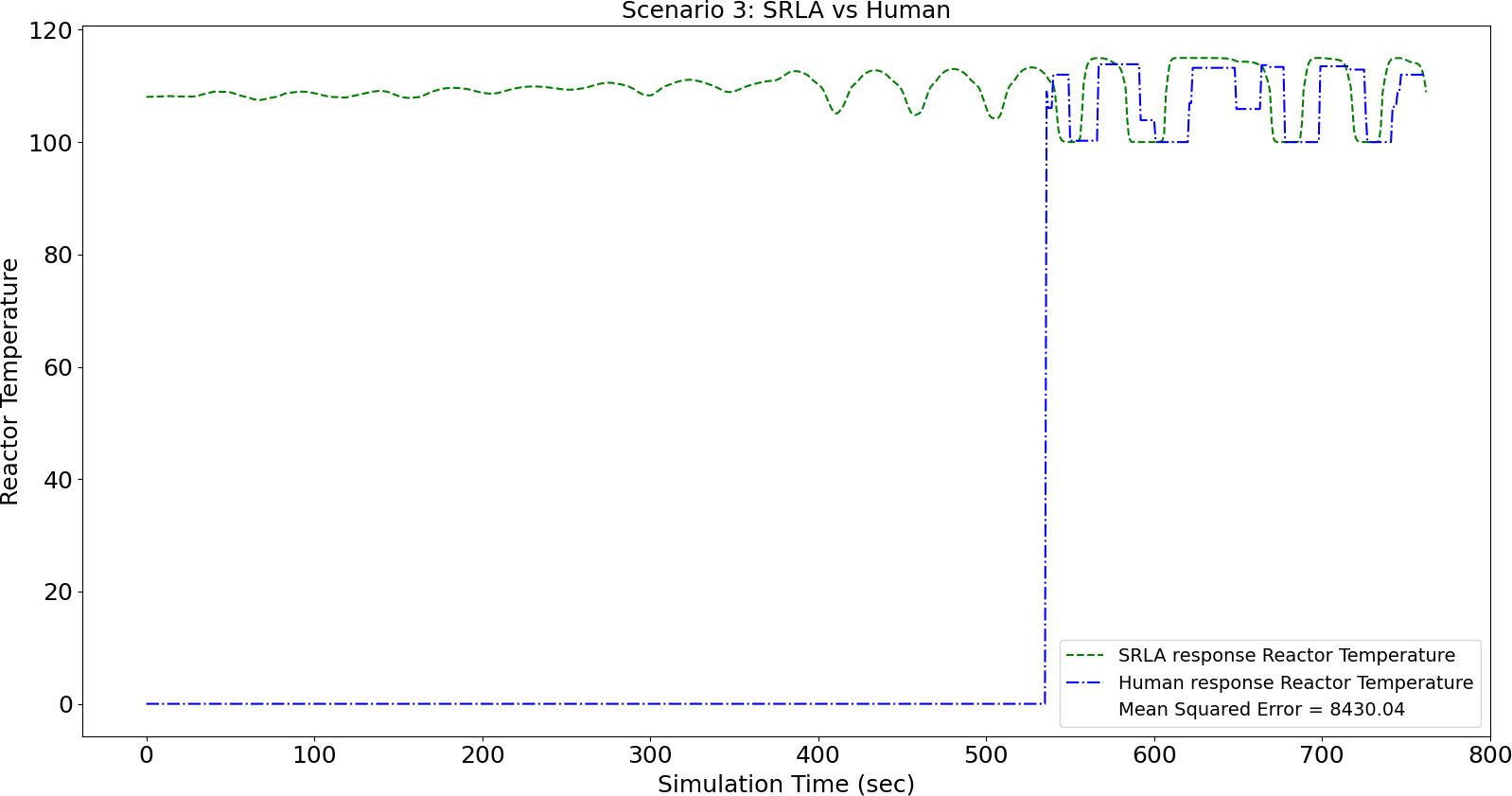}}
                \caption{SRLA vs human response of participant (a) P30 and (b) P32 within the Time of Interest (TOI) of critical alarms (scenario 1) and alarm overflow (scenario 3).}
                \label{fig: srla_vs_human_3032}
            \end{figure}

            \paragraph{AI (SRLA) vs Human Response}
            The comparison between the SRLA suggestion and the human response can be seen in \cref{fig: srla_vs_human_3032}, which shows that P32 follows the SRLA suggestions better than P30 for both the scenario of critical alarm and alarm overflow.
            
            \paragraph{Radar Plot}
            Participant P30 generally experiences higher consequences, workload, and positive interaction with AI, however, has higher errors following the AI suggestion. P30 also ranks procedures support and the need for AI validation higher as evidenced by the preferred choice of intervention support. P30 also encounters higher recovery, reaction, and response time leading to more alarms and the opening of more mimics, indicating higher evaluation time for following the screen procedures support. In contrast, P32 excels in situational awareness, quicker transitions between points of interest sustained attention at specific locations, and more deliberate or careful exploration of the visual environment (based on eye movement dynamics). P32 relies more on AI support and rates highly its help demonstrating quicker recovery and higher importance of deep reinforcement learning, however, experiences a higher load in doing so as also evident from the rating of AI as an additional load.
            
            \begin{figure}[!t]
                \centering
                    \includegraphics[width=\linewidth]{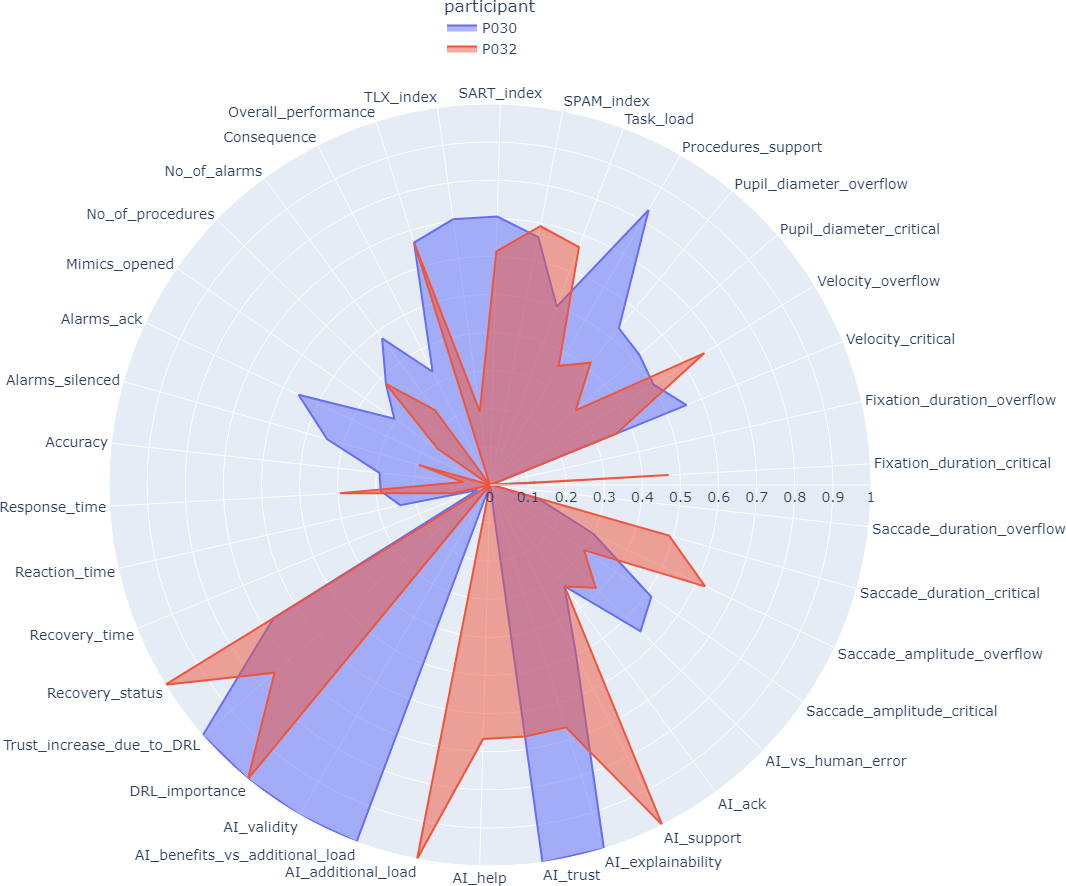}
                    \caption{Radar plot between participant 30 and 32}
                    \label{fig: radar_3032}
            \end{figure}  
            
        \vspace{0.1in}
        \subsubsection{P37 vs P96}
        This section comprises the comparative evaluation of P37 (preference: AI and Screen Procedures) and P96 (preference: AI and SRLA).

\begin{figure}[!t]
                \centering
                \subfloat[]{\includegraphics[width=0.48\linewidth]{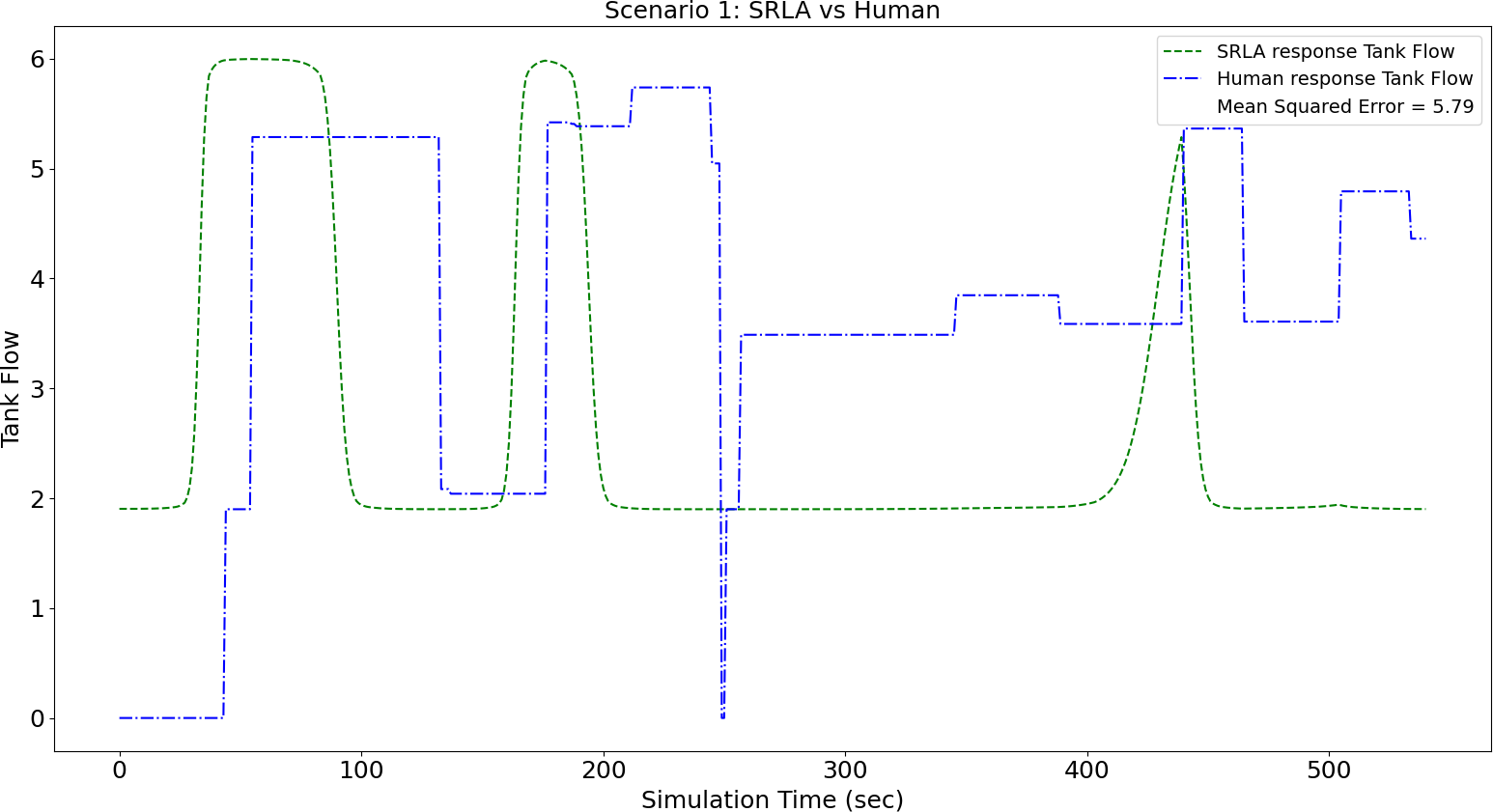}
                            \includegraphics[width=0.48\linewidth]{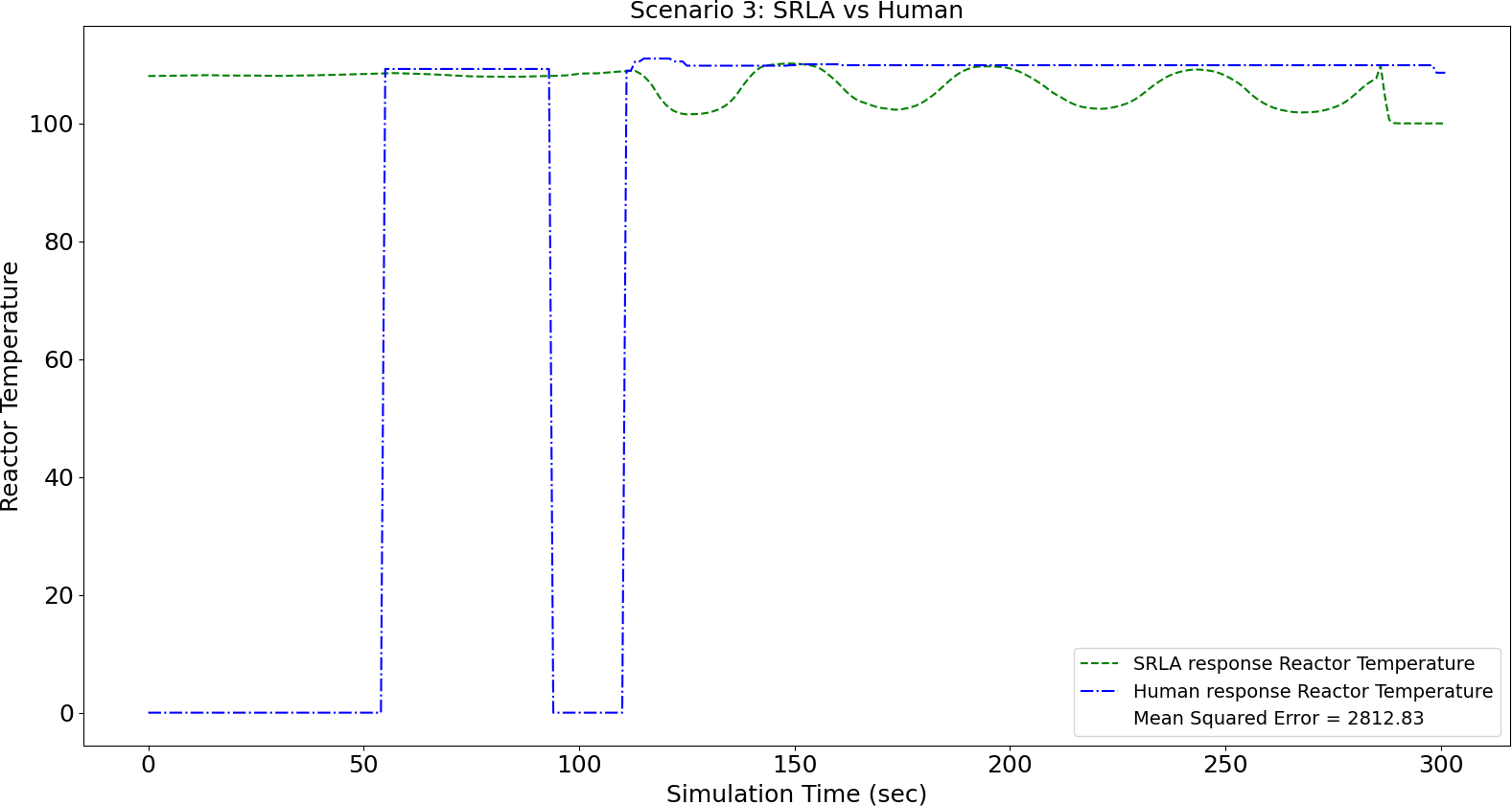}}
                    \\
                \subfloat[]{\includegraphics[width=0.48\linewidth]{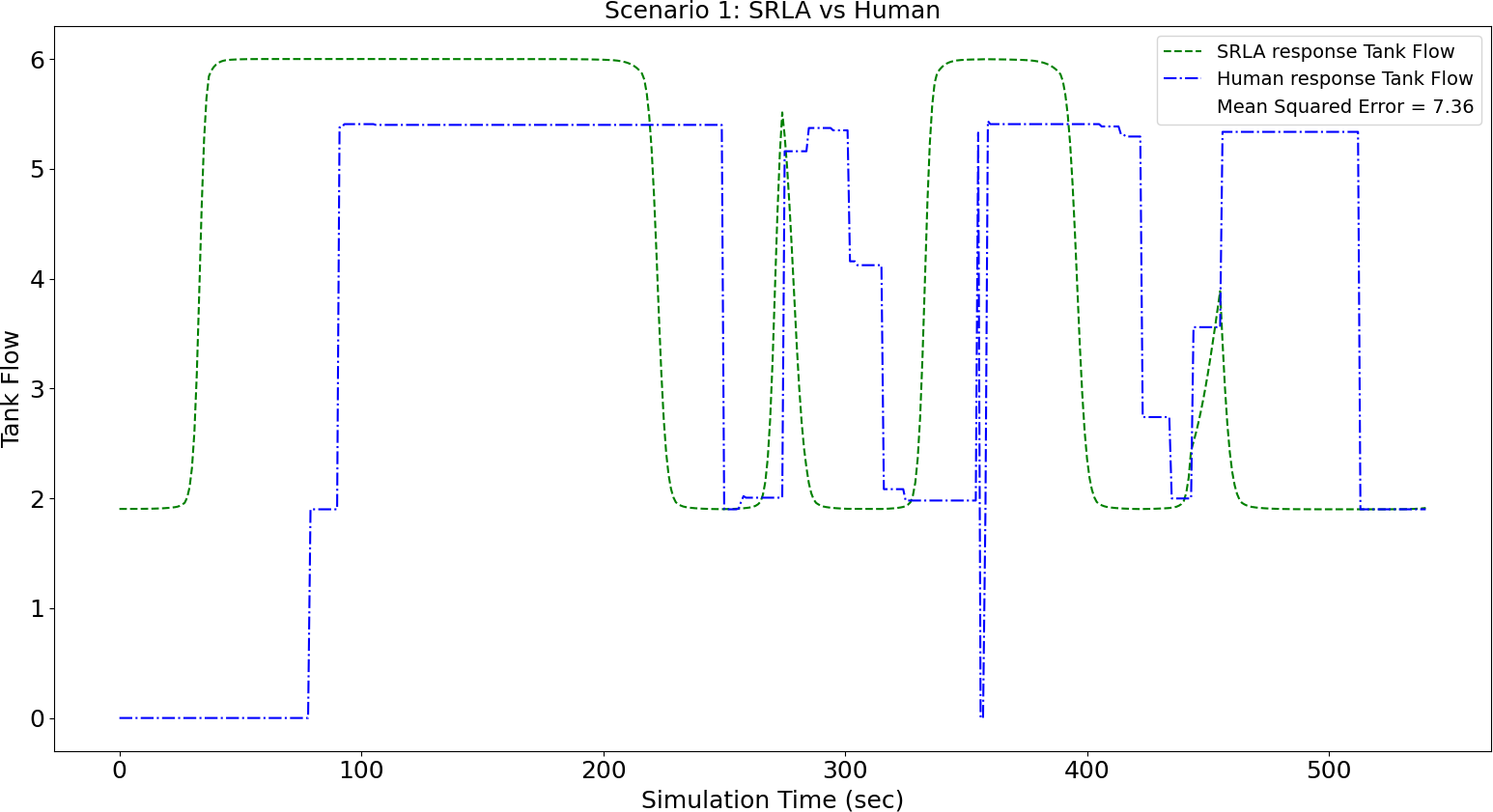}
                            \includegraphics[width=0.48\linewidth]{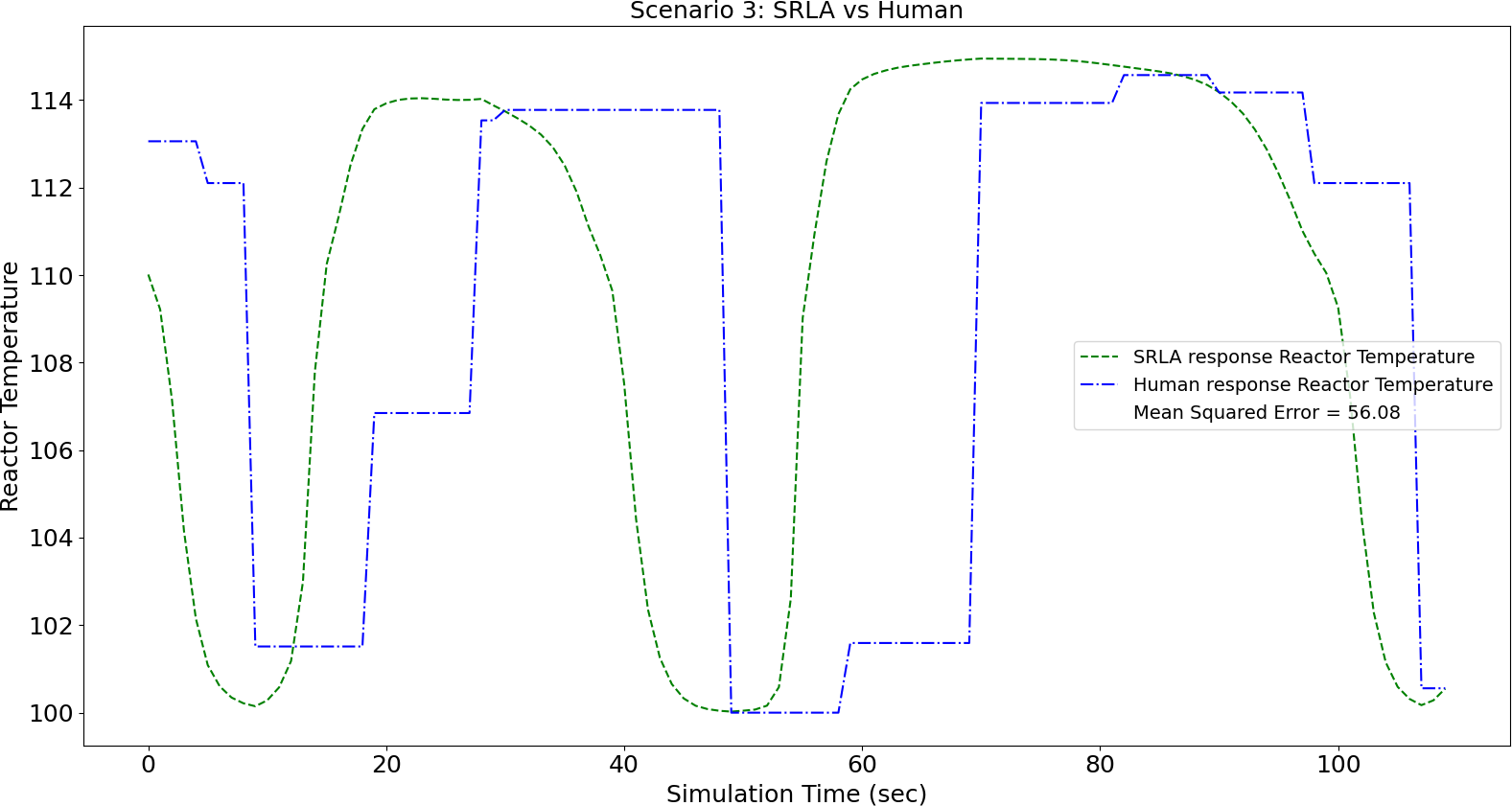}}
                \caption{SRLA vs human response of participant (a) P37 and (b) P96 within the Time of Interest (TOI) of critical alarms (scenario 1) and alarm overflow (scenario 3).}
                \label{fig: srla_vs_human_3796}
            \end{figure}

            \paragraph{SRLA vs Human Response}
            The comparison between the SRLA suggestion and the human response can be seen in \cref{fig: srla_vs_human_3796}, which shows a comparable performance of both the participants, however, P96 tends to respond more accurately to the dynamic changes of the SRLA suggestions.
            
            \paragraph{Radar Plot}
            Participant P37 encounters more challenges with higher AI vs human error, longer recovery time, increased alarms, opened procedures, severe consequences, higher task load, EDA, and temperature. In contrast, Participant P96 experiences a more positive interaction with AI, including higher AI acknowledgment, support, explainability, better recovery status, response time, accuracy, and overall performance, but also higher perceptions of AI as an additional load. P96 also reports a balanced perceived task load and higher situational awareness. 
            
            \begin{figure}[!t]
                \centering
                    \includegraphics[width=\linewidth]{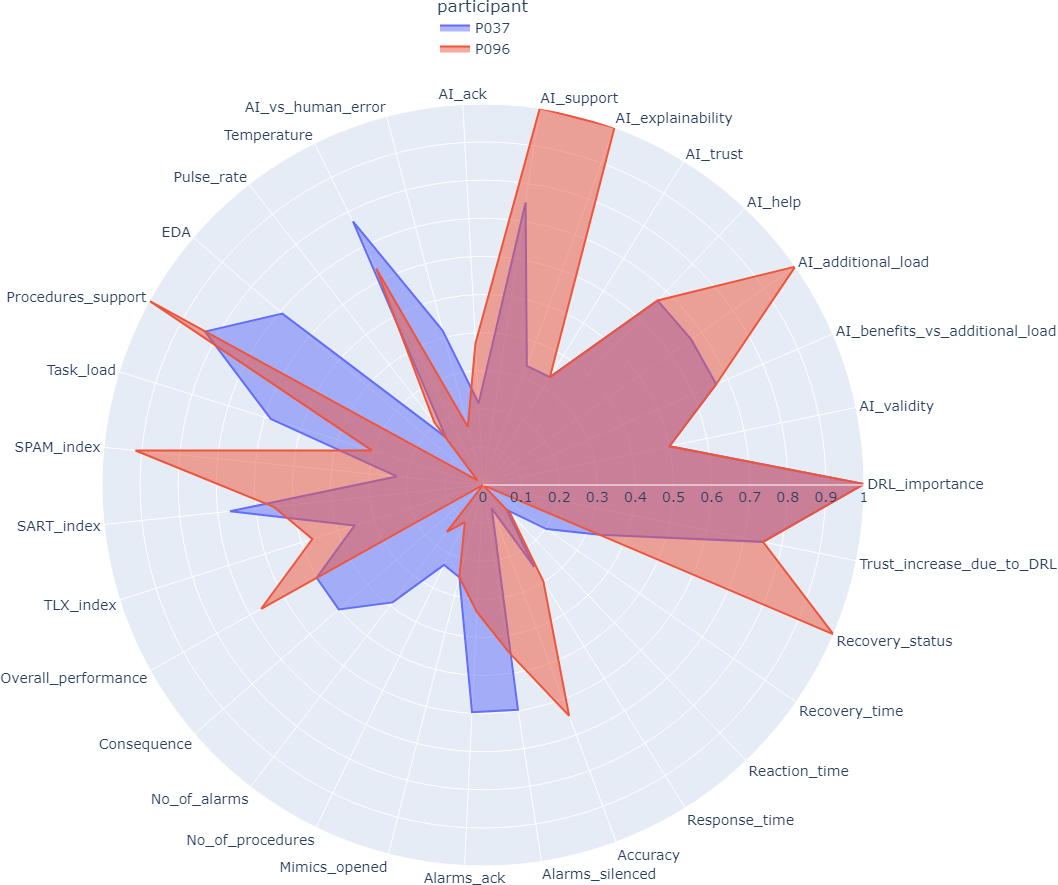}
                    \caption{Radar plot between participant 37 and 96}
                    \label{fig: radar_3796}
            \end{figure}

\section{Analysis II: Human Failure Prediction in Real-Time using Process Variables and Human-Machine Interaction}
\label{sec:hmm_pred}

\begin{figure}[!t]
    \centering
    \subfloat{\includegraphics[width=0.48\linewidth]{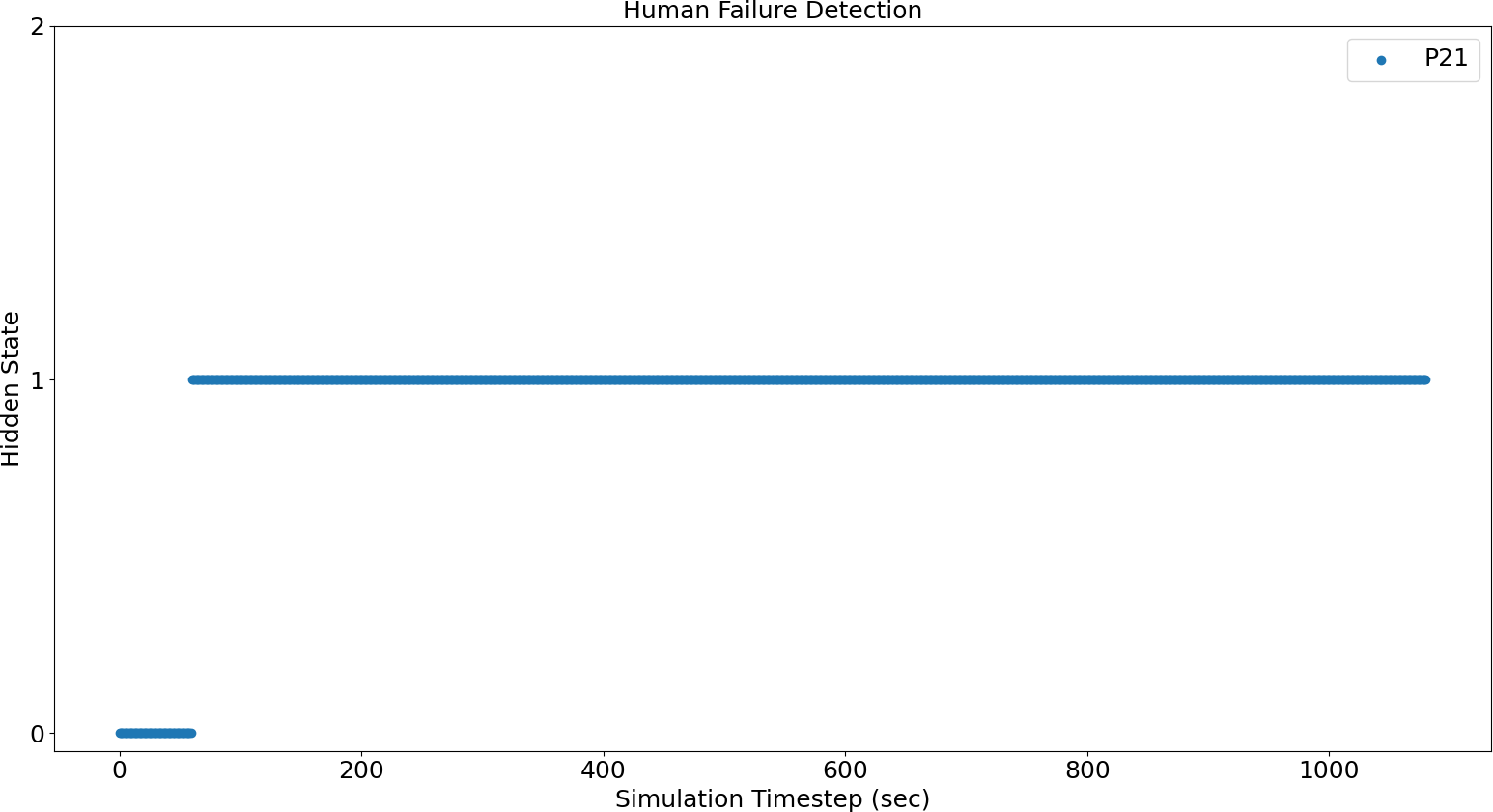}}
    \subfloat{\includegraphics[width=0.48\linewidth]{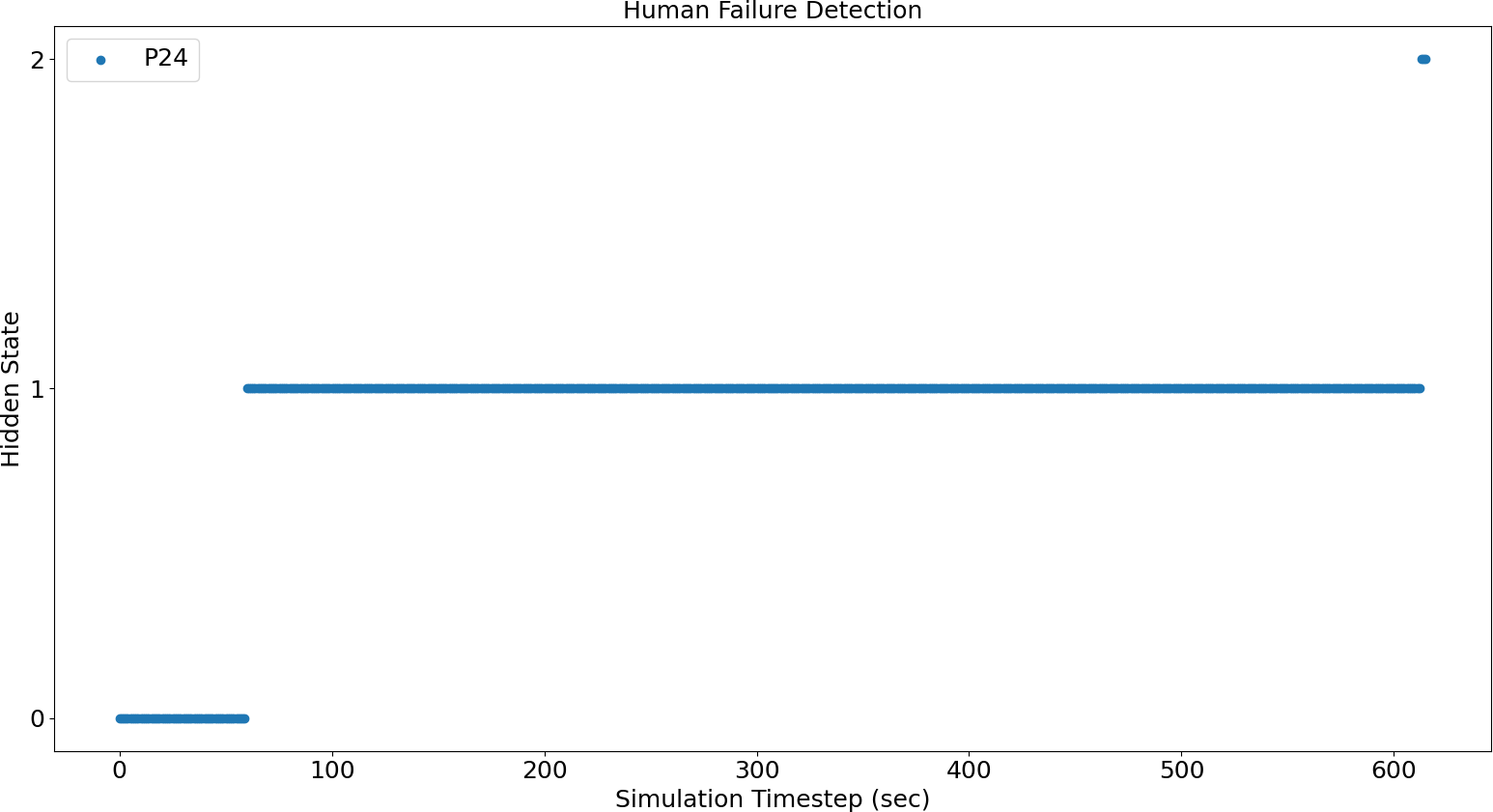}}
        \\
    \subfloat{ \includegraphics[width=0.48\linewidth]{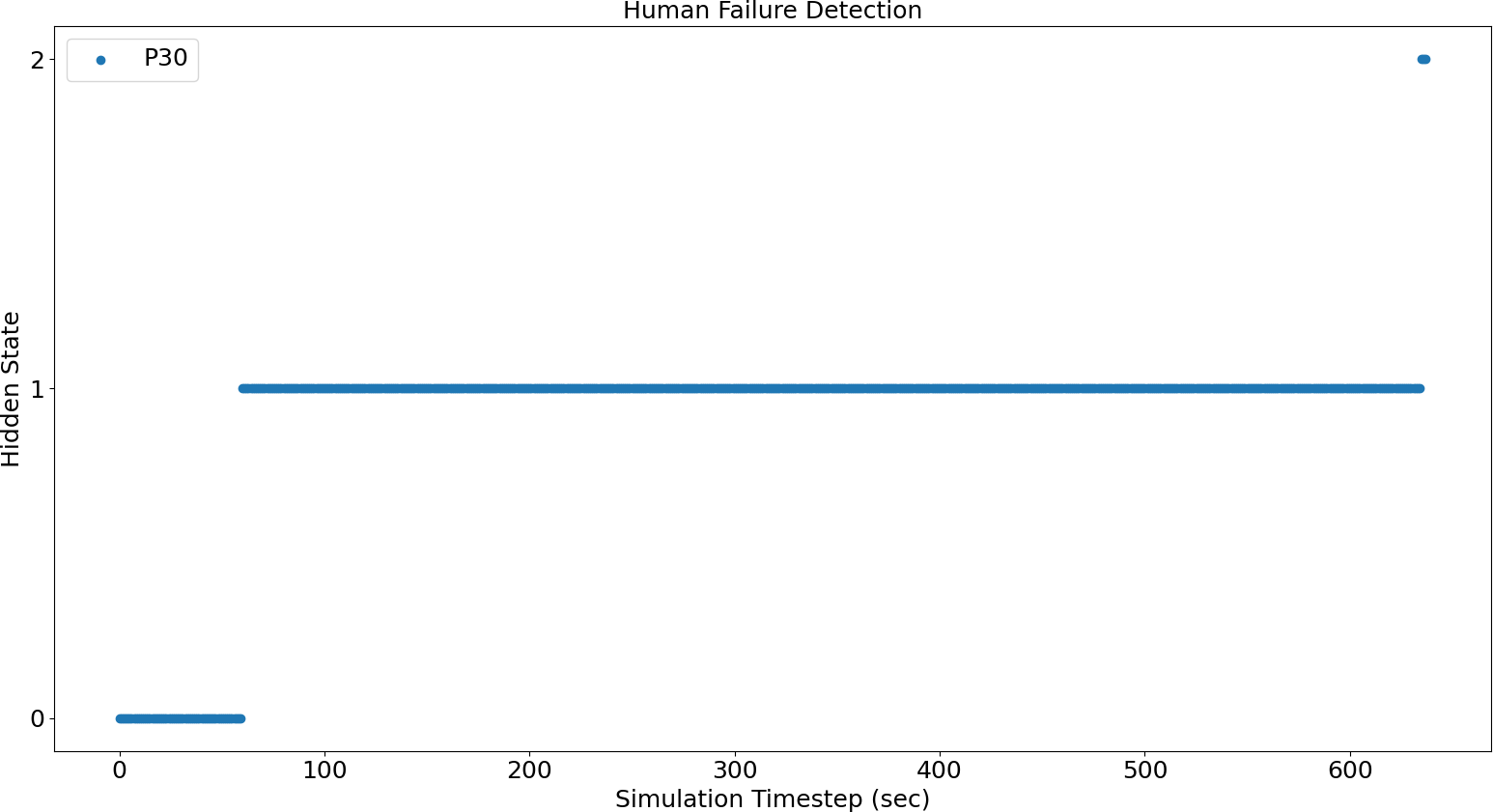}}
    \subfloat{\includegraphics[width=0.48\linewidth]{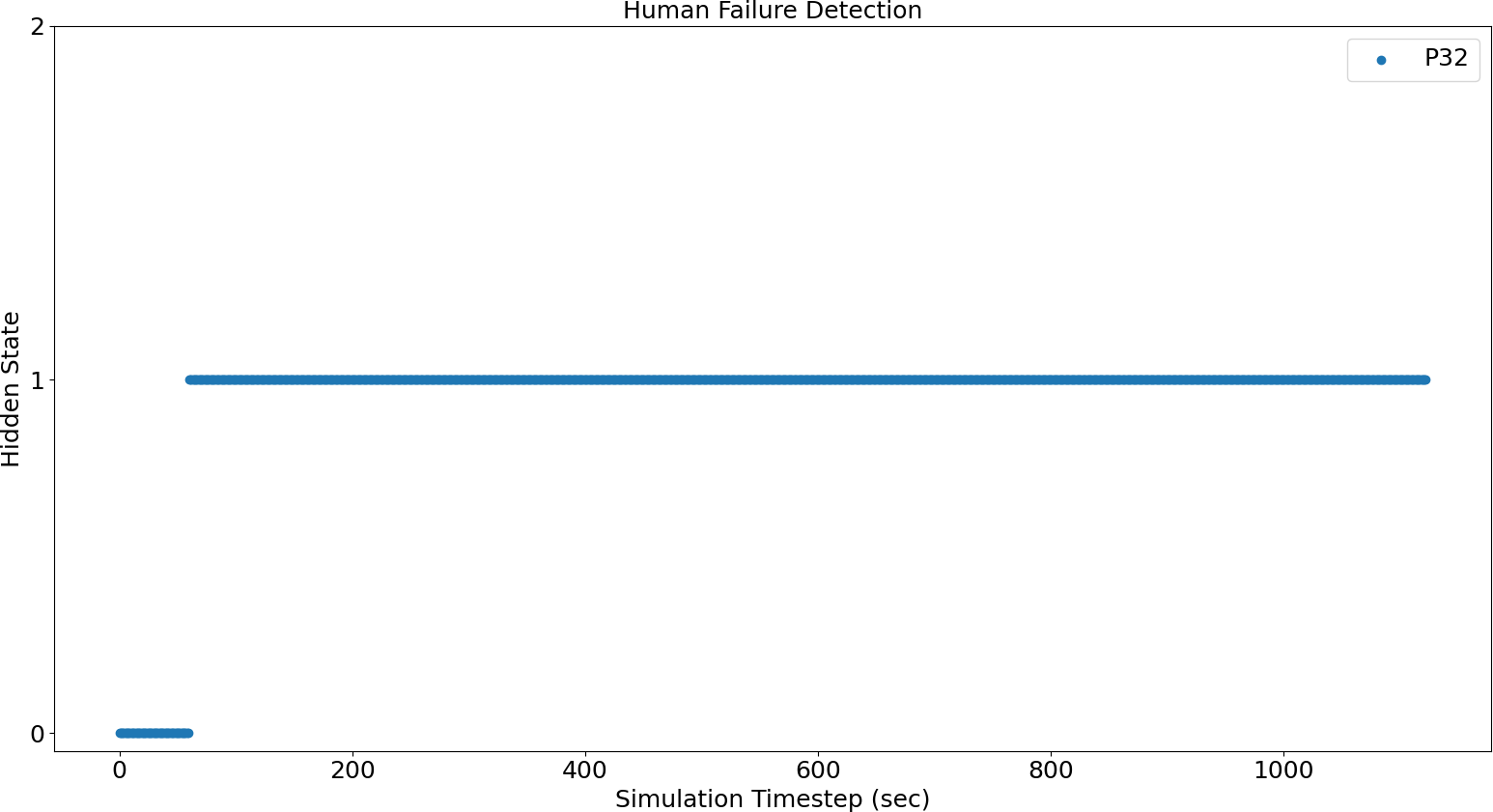}}
        \\
    \subfloat{\includegraphics[width=0.48\linewidth]{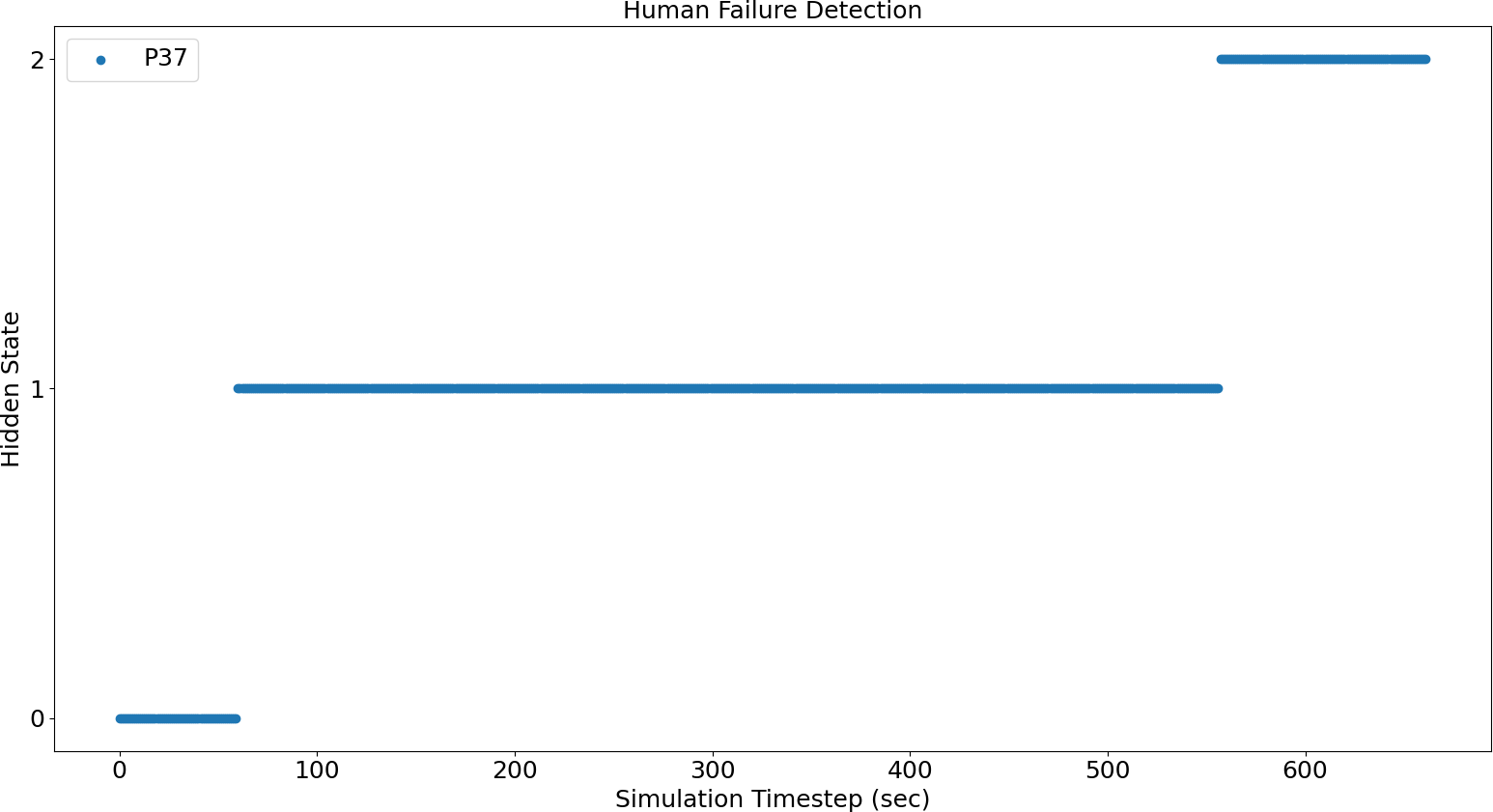}}
    \subfloat{\includegraphics[width=0.48\linewidth]{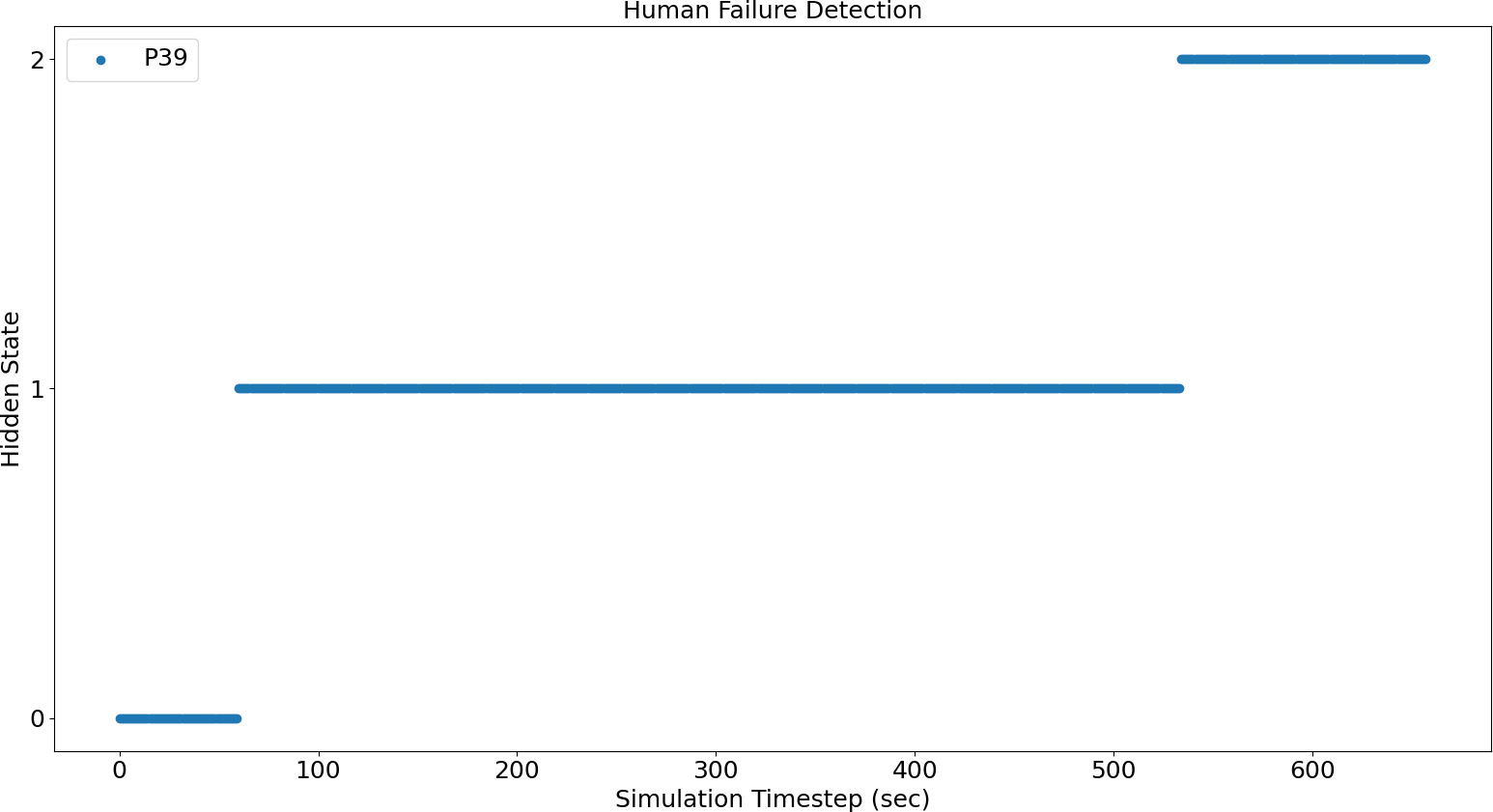}}
        \\
    \subfloat{\includegraphics[width=0.48\linewidth]{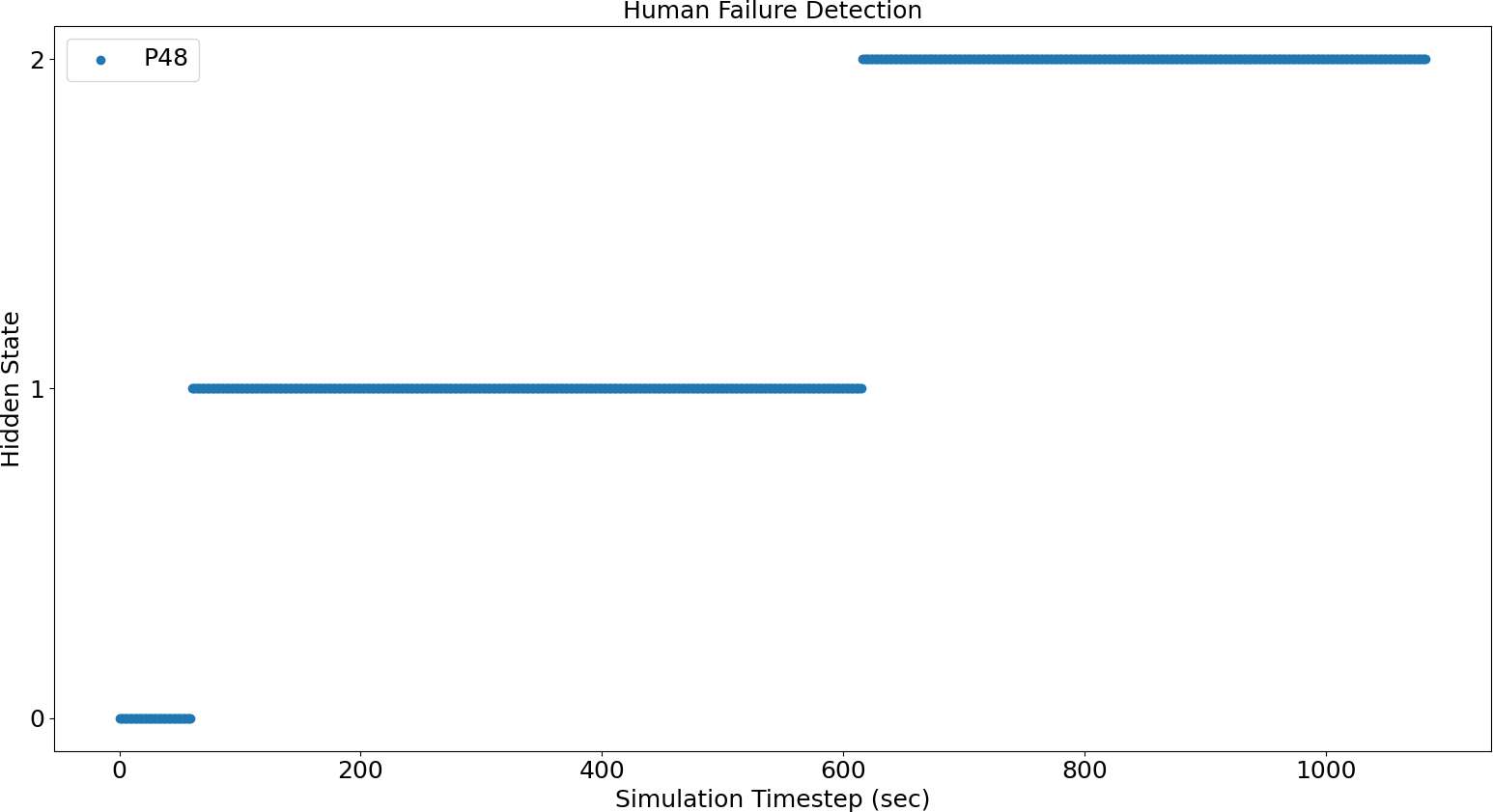}}
    \subfloat{\includegraphics[width=0.48\linewidth]{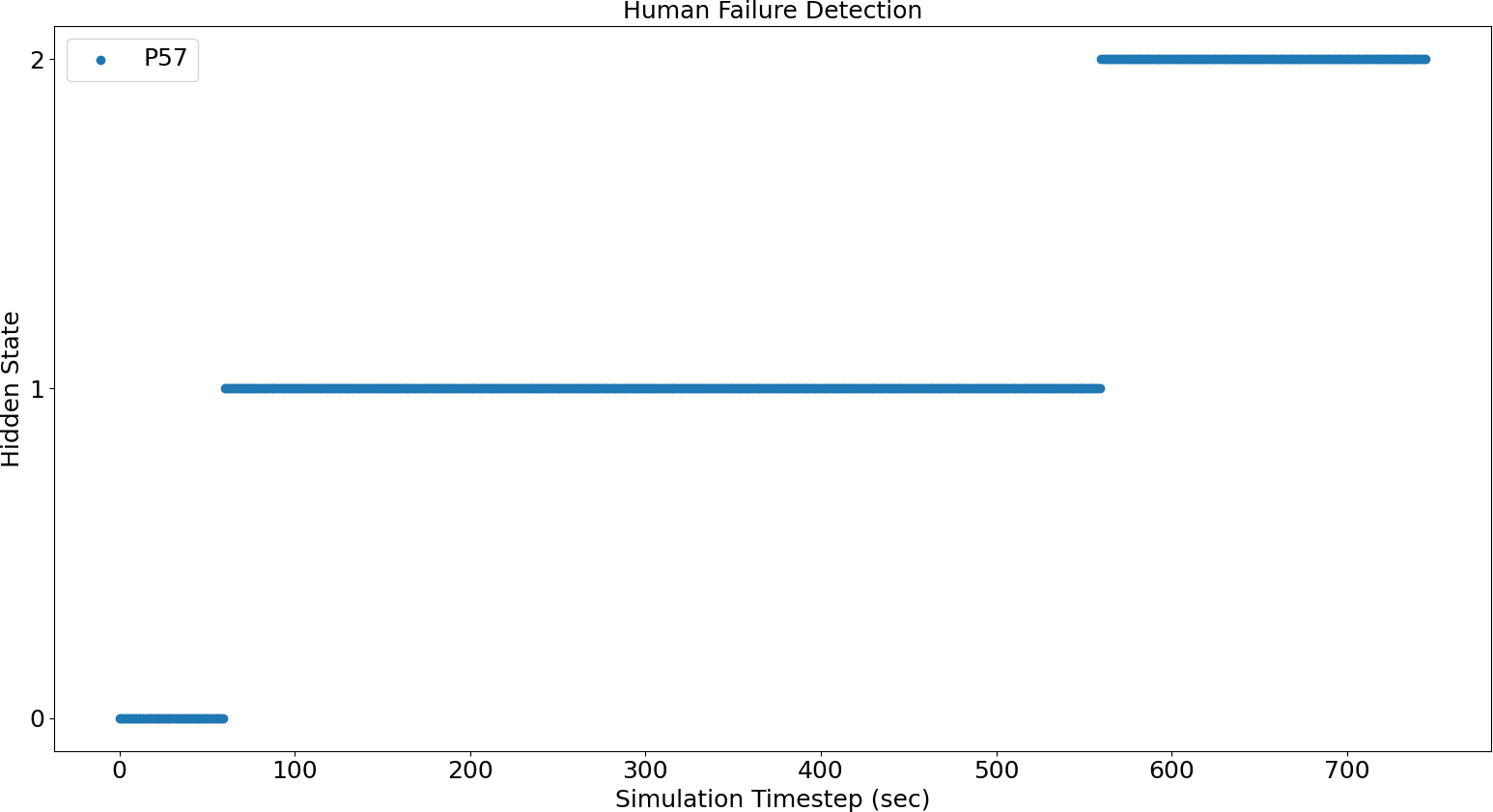}}
        \\
    \subfloat{\includegraphics[width=0.48\linewidth]{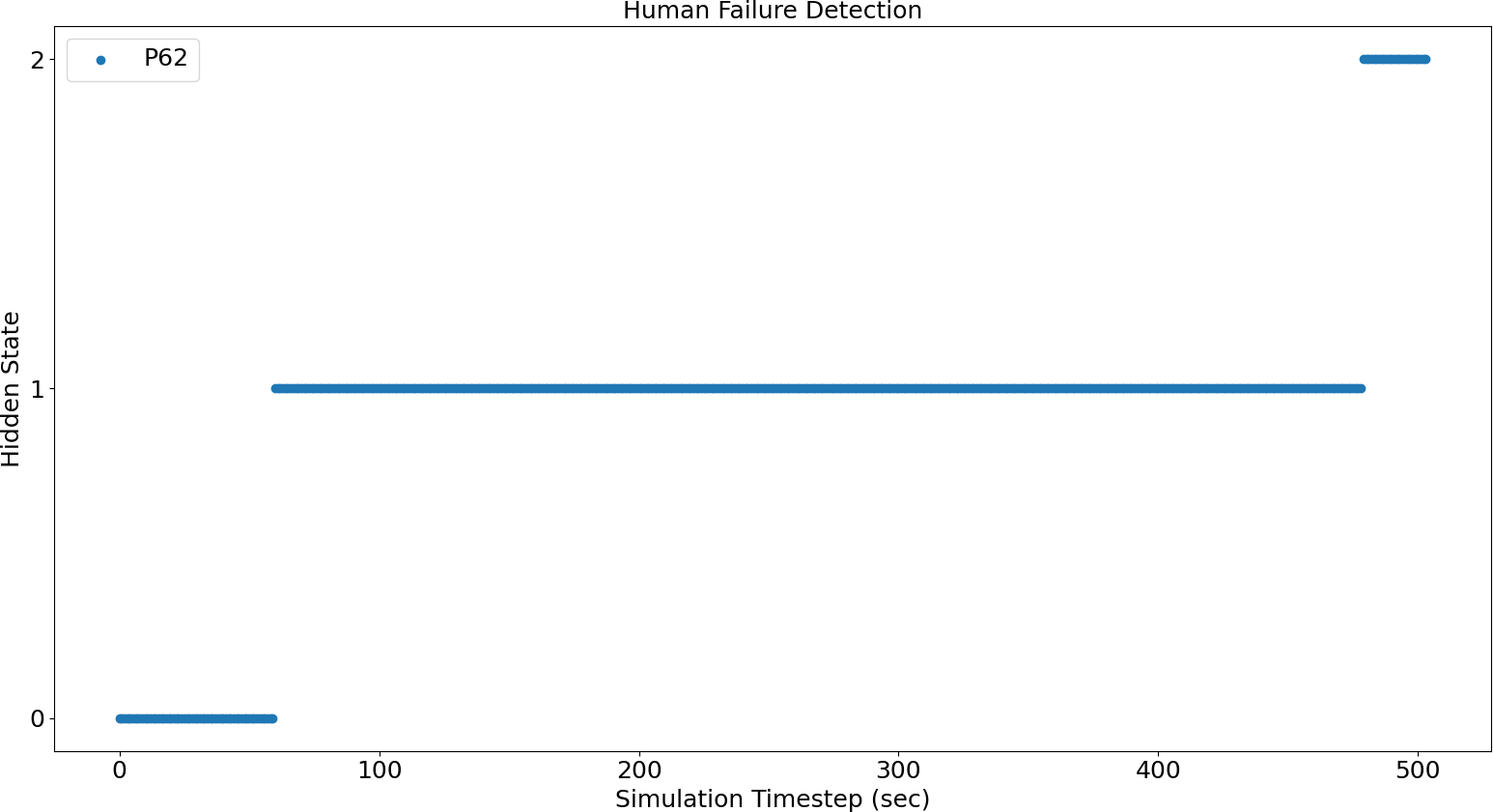}}
    \subfloat{\includegraphics[width=0.48\linewidth]{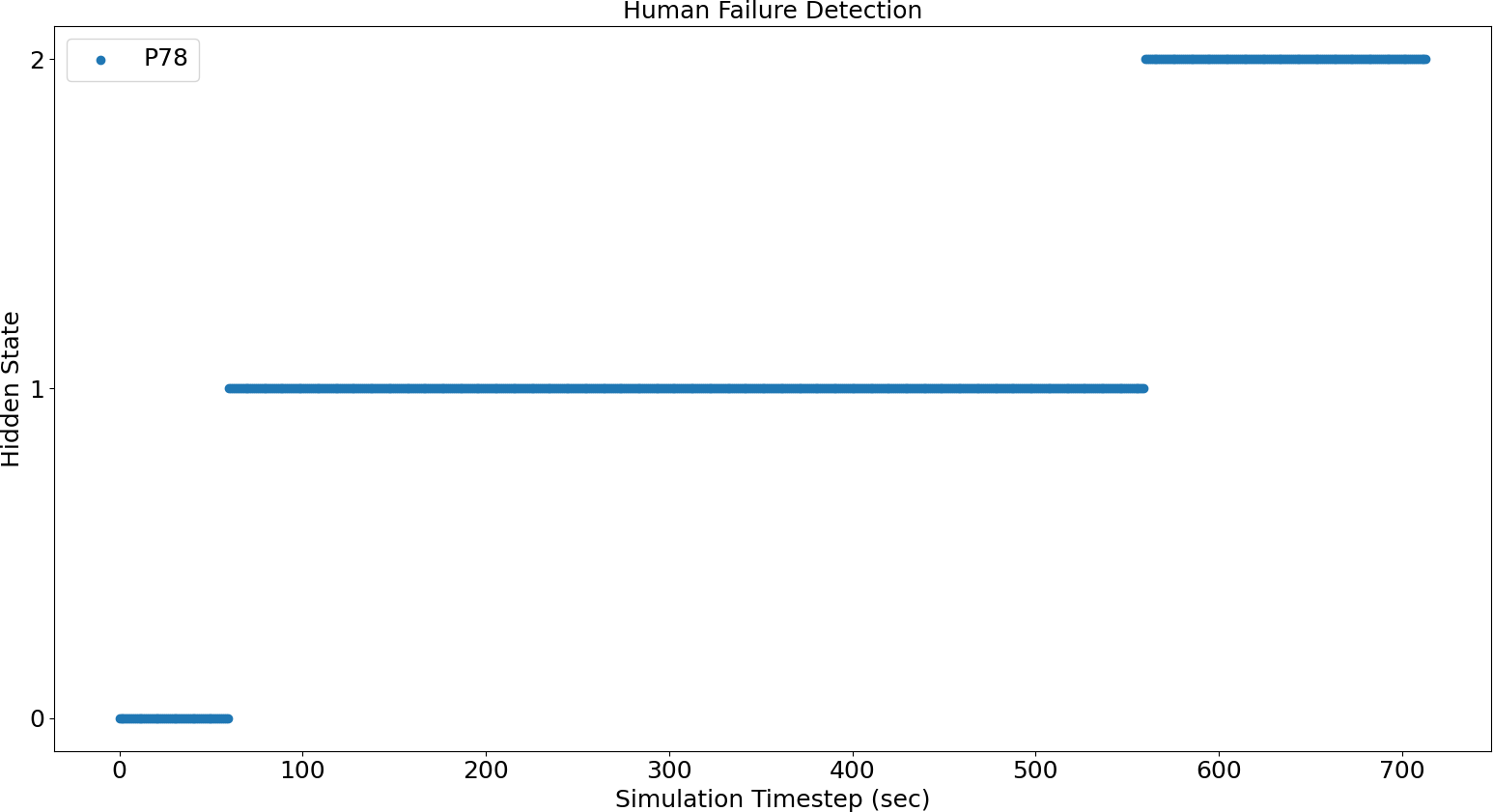}}
        \\
    \subfloat{\includegraphics[width=0.48\linewidth]{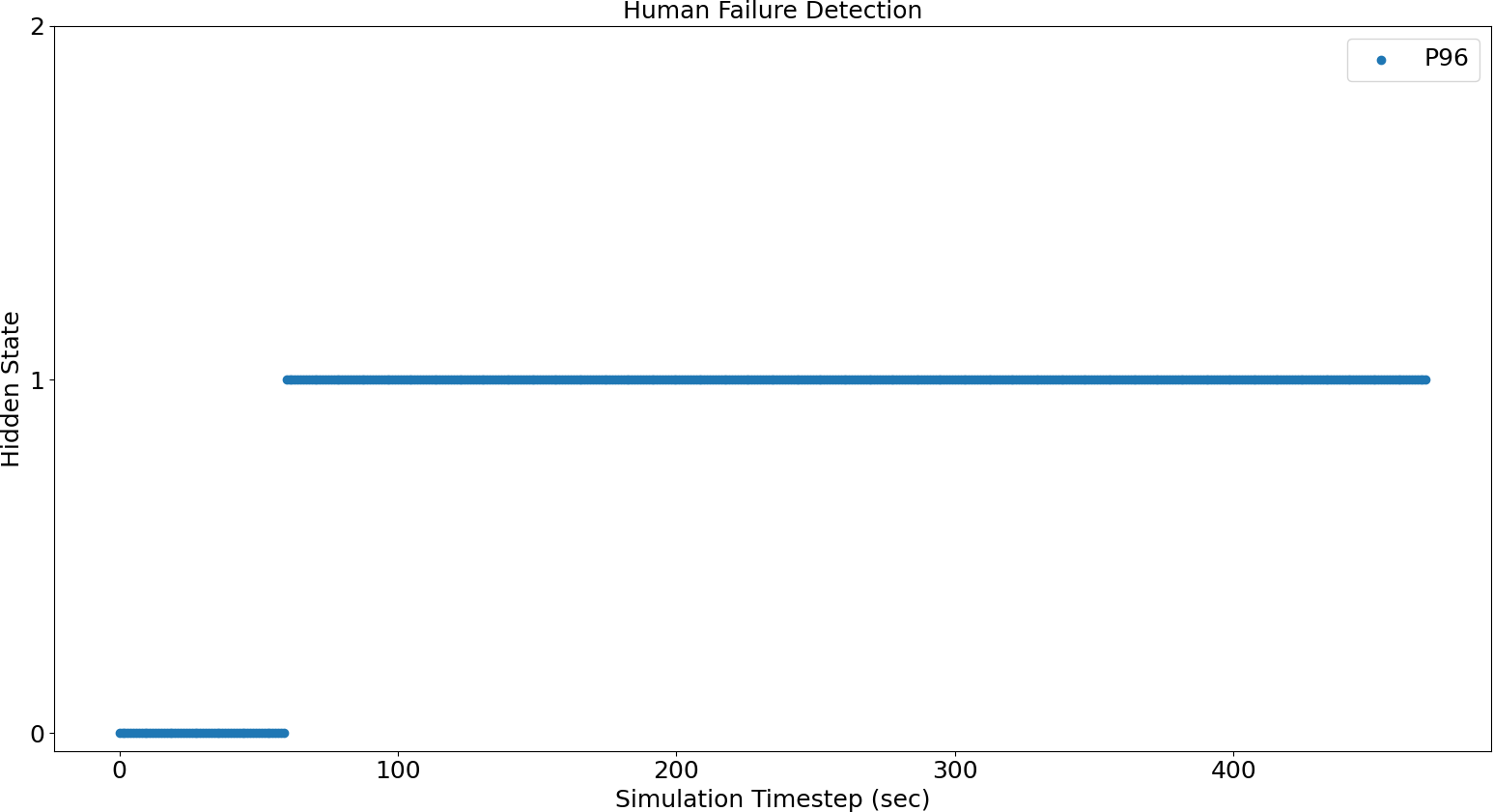}}
    \subfloat{\includegraphics[width=0.48\linewidth]{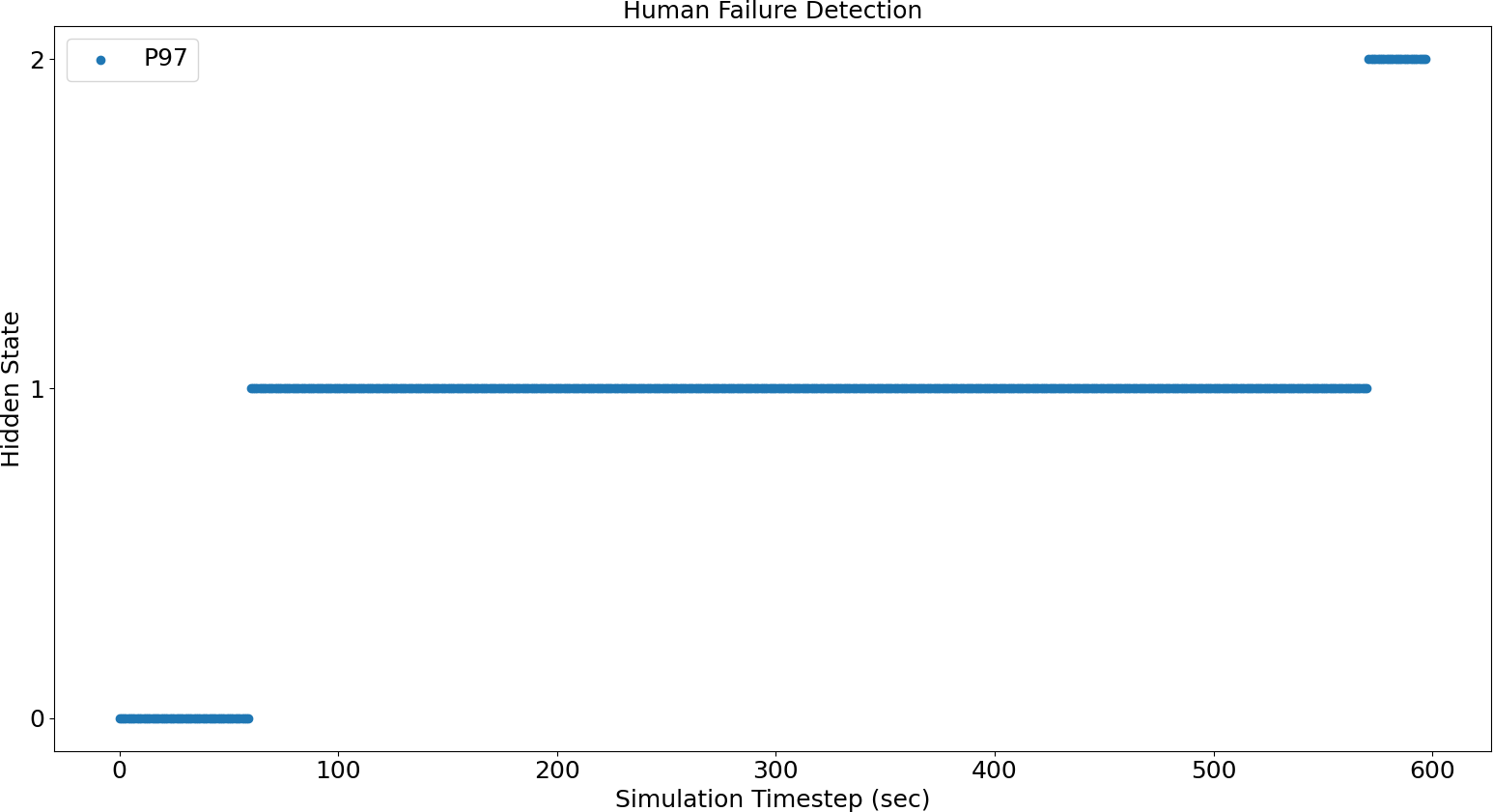}}
    \caption{Sample of participant's failure prediction based on the real-time simulator log time-series data.}
    \label{fig: human-fail-det}
\end{figure}

The data used for the Hidden Markov Model (HMM) included the variables that can be accessed in real-time, such as process variables, alarm logs, and human-machine interactions such as the number of procedures opened, the number of times the manual switch was controlled, etc. Optuna \cite{optuna_2019} was used for hyperparameter tuning for the HMM modeling and the best chosen hyperparameters are summarized in \cref{tab:hyp_hmm}. With 95.8\% accuracy for the situation of alarm overflow (scenario 3), the HMM was able to predict the hidden state (state 2) for which the human would fail the situation using the raw data from the process logs as shown in \cref{fig: human-fail-det}. The plots reveal that in cases where participants encountered failure in the scenario, the HMM demonstrated the capability to forecast the failure well in advance. The HMM can effectively intervene in human actions by issuing timely alarms about potential consequences and highlighting alternative courses of action. 

\begin{table}[htbp]
\centering
    \caption{Hyperparameters for HMM modeling}
    \begin{tabular}{|c||c|}
      \hline
      \textbf{Hyperparameter} & \textbf{Value} \\
      \hline
      \texttt{n\_states} & 3 \\
      \texttt{model\_type} & 'gmmhmm' \\
      \texttt{n\_mix} & 3 \\
      \texttt{covariance\_type} & 'tied' \\
      \texttt{is\_lr} & True \\
      \texttt{is\_scalar} & True \\
      \texttt{is\_pca} & True \\
      \texttt{n\_decomp} & 4 \\
      \hline
    \end{tabular}
    \label{tab:hyp_hmm}
\end{table}

    \subsection{Factor Loadings on the Principal Component Analysis (PCA)}
    The reason for this analysis is to understand the importance of feature sets or data sources used for the prediction of the system state based on the state of the process as well as the state of the human and human-machine interaction. The Principal Component Analysis (PCA) is a method for reducing dimensionality in multivariate data. It encompasses the computation of factor loadings (\(f_{ij}\)), representing the coefficients in the linear combination of the original variables (\(X_i\)) to construct each principal component (\(PC_j\)). The factor loadings for the initial principal component are determined by the elements of the eigenvector (\(a_{kj}\)), acquired through the eigenvalue problem for the covariance or correlation matrix of the original variables.

    The principal components (\(PC_j\)) are novel variables formed as linear combinations of the original variables (\(X_i\)), with the factor loadings (\(a_{ij}\)) serving as coefficients in these combinations. 
    
    Factor loadings:
    \[ f_{ij} = \sqrt{\frac{1}{p} \sum_{k=1}^{p} a_{kj}^2} \]
    
    Principal components:
    \[ PC_j = \sum_{i=1}^{p} a_{ij}X_i \]
    
\begin{figure}[!t]
        \centering
        \subfloat[]{\includegraphics[width=0.95\linewidth]{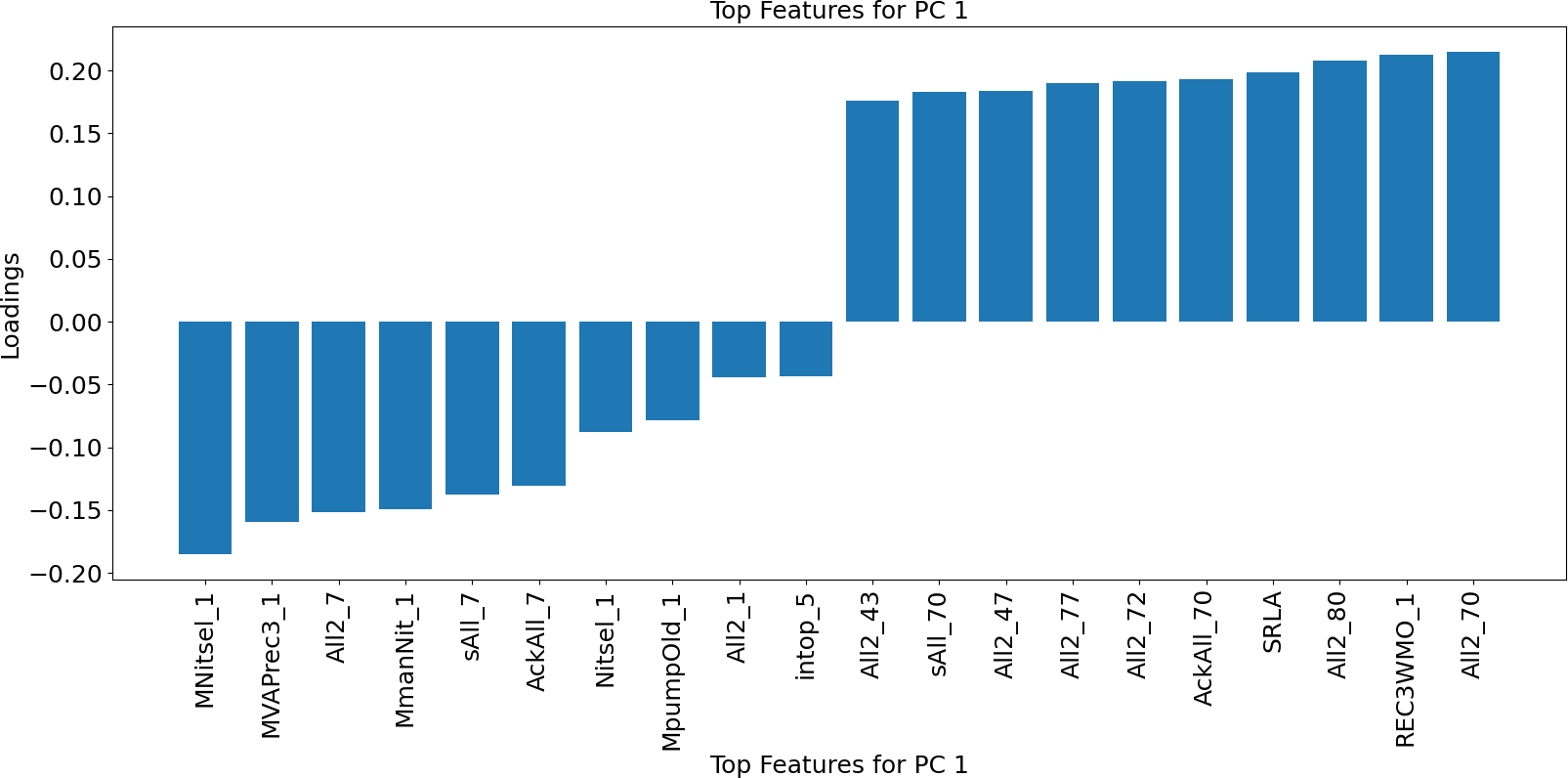}}
            \\
        \subfloat[]{\includegraphics[width=0.95\linewidth]{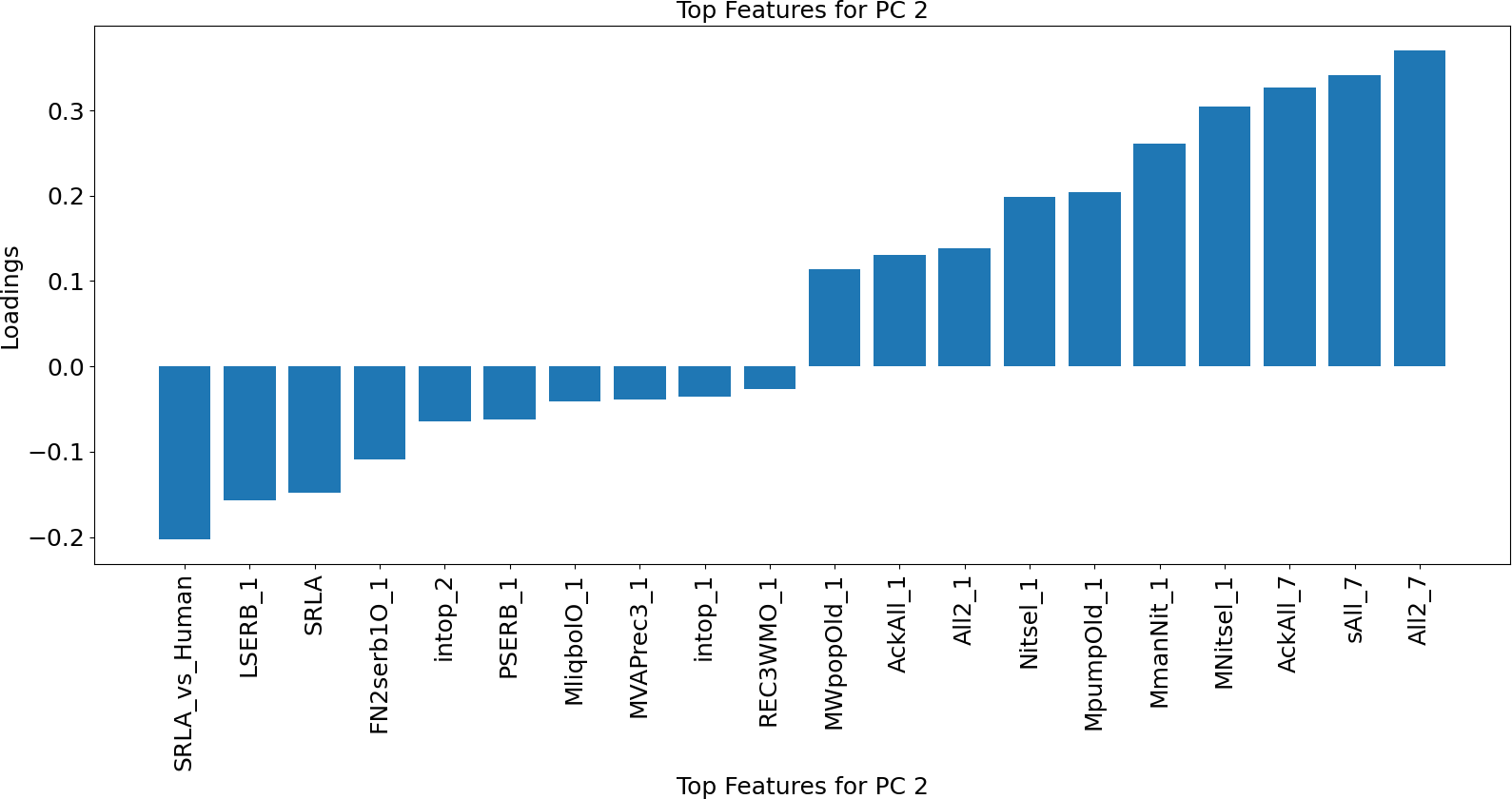}}
            \\
        \subfloat[]{ \includegraphics[width=0.95\linewidth]{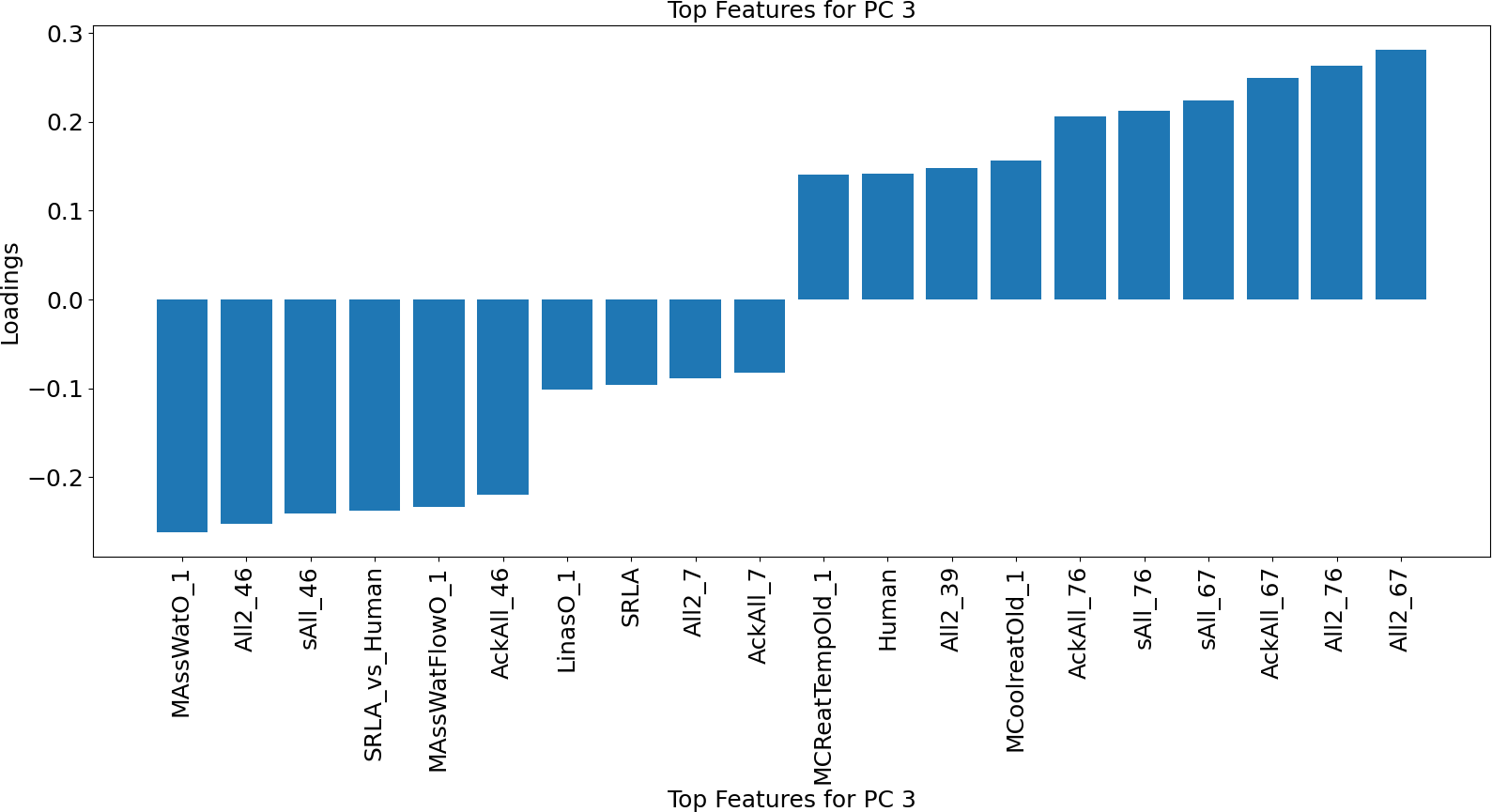}}
            \\
        \subfloat[]{\includegraphics[width=0.95\linewidth]{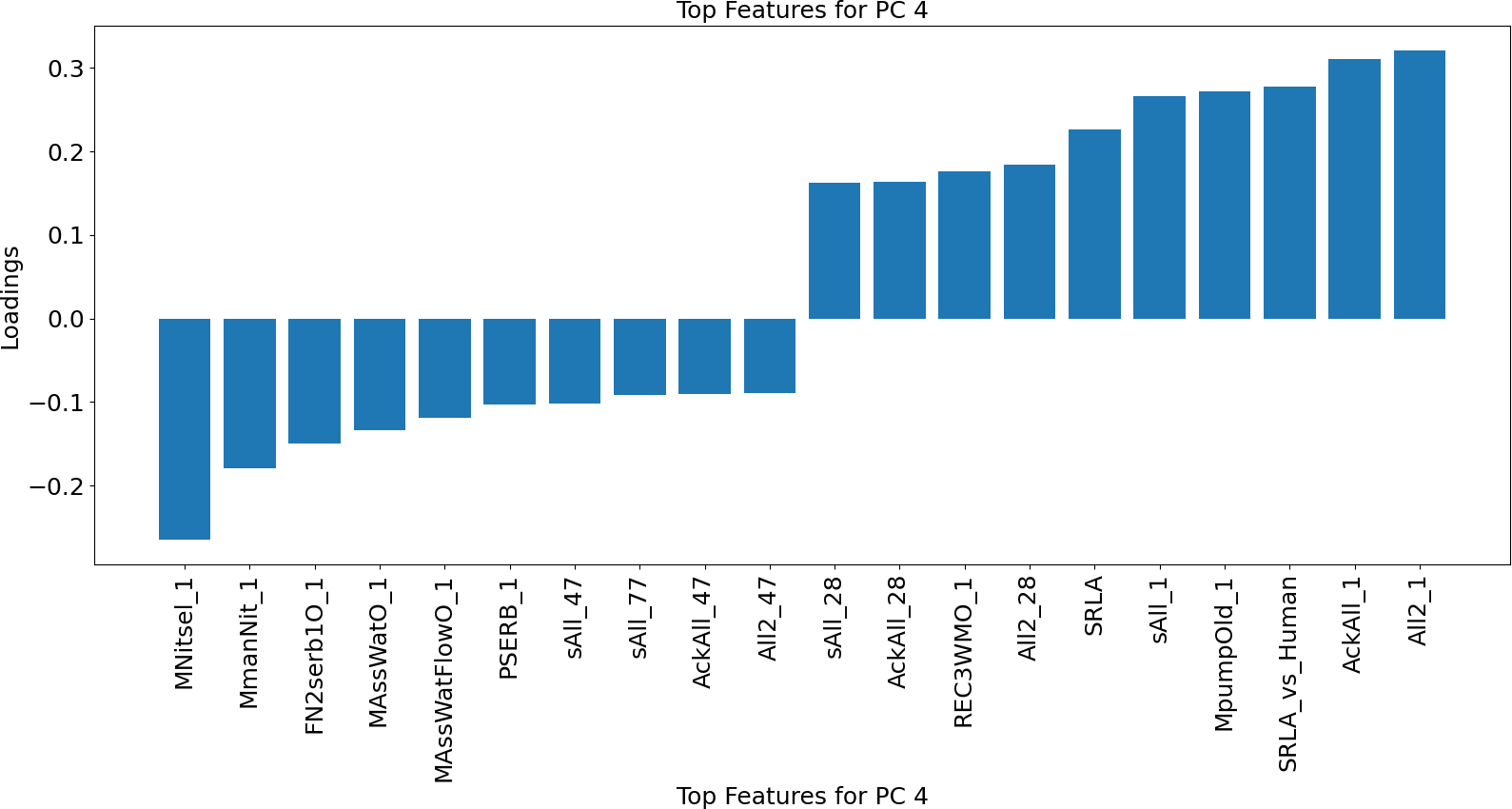}}
        \caption{Top 10 factor loadings in both directions on the Principal Component Analysis (PCA).}
        \label{fig:pca_comp_fact}
    \end{figure}

    In these equations, \(p\) denotes the number of variables in the original dataset, These formulae aid in the analysis and encapsulation of essential information within a dataset through a condensed set of variables. The top 10 factor loadings in both directions on the chosen 4 Principal Components are shown in \cref{fig:pca_comp_fact}.
    
    It can be seen from the description of these features as provided in \textit{Appendix}, that the most important factor loadings for the principal components are the combination of process, alarms, and HMI variables.

    \subsection{Intervention Strategies and Applications: Towards a General SRLA-Based Support System}
    In the proposed framework, a Dynamic Influence Diagram (DID) is used for process-level abnormality detection and the Hidden Markov Model (HMM) is for human-level abnormality detection. The Specialized Reinforcement Learning Agent (SRLA) is used in this loop to identify the best possible intervention strategies and suggest them to the human operator who is responsible for the final decision. Based on these intervention strategies we propose several applications for which such a system can be implemented in the real world safety-critical process industries. The following proposed applications in the real world are based on the interviews with the safety-critical industry experts, control room operators, and decision-makers:
    
        \subsubsection{Training and Tuning}
        Such a framework can be implemented during the training of new operators as well as training the experienced operators when there is a major change in the system. It can help the operator focus on the pruned procedures and what is needed to be followed in such a situation which can later be generalized by their intuition. Furthermore, it can also be used for tuning the system's parameters \cite{Hsieh2012} by observing the predicted hidden states and control suggestions. 
        
        \subsubsection{Decision Support to Operator}
        Such a system has already been tested and provided in this research, where the framework is used to provide suggestions to the operator in a simulated environment, however, in real-world the conditions can be more complex, and therefore, it may be challenging to directly implement such a system into safety-critical industries \cite{lee2007development}. Nevertheless, even in scenarios where the recommendation seeks the operator's confirmation for automating the process, the ultimate decision remains in the hands of the human. This approach ensures the implementation of a higher level of safety. Therefore, a tradeoff between increased task load and aid in decision support is to be made as also evidenced by the findings in the experimental results. Furthermore, it can also help the operator to self asses their operating state and control decisions. 
    
        \subsubsection{Decision Support to Supervisor}
        An alternative application could involve offering decision support directly to supervisors instead of burdening operators with the additional workload. However, ethical considerations regarding the disclosure of information about the operator's state must be carefully taken into account. In such a configuration, the supervisor can assess the overall system performance, determine necessary corrective measures, and disseminate relevant information among the operators.
    
        \subsubsection{SRLA-Based Validation}
        Specialized Reinforcement Learning Agent (SRLA) can also be trained and used to validate the control actions taken by the operator in abnormal as well as normal operating conditions and prompt for any predicted consequences as also a similar setup developed in \cite{lee2007development}.
    
        \subsubsection{SRLA Control}
        On the opposite end of the spectrum, an approach involving solely providing suggestions to the operator is full process automation, particularly in instances where the HMM predicts elevated workload, reduced situational awareness, or an imminent failure. However, the automation would only be performed after receiving confirmation from the operator. This form of automation proves beneficial in situations where the agent's speed in performing control actions surpasses that of a human, as illustrated in our experiment's scenario 3. It's crucial to acknowledge that incorporating human latency as an input to SRLA is essential for delivering more robust control suggestions. Furthermore, SRLA can also determine which processes to automate and which to recommend for manual intervention. 
        
\section{Discussion}
While comparing GroupN with GroupAI the general observation that was made was that GroupAI had a better overall performance and reduced task load, however, the participants in GroupAI also had lower situational awareness as compared to GroupN. The comparison among participants in GroupAI revealed that the participants who followed the suggestions provided by the Specialized Reinforcement Learning Agent (SRLA) more closely resulted in better situational awareness and performance, however, a higher workload was observed. The differences in these experiences highlight the importance of evaluating various factors in human-AI interactions for a comprehensive understanding.

In situations where a high level of well-informed situational awareness is crucial, artificial intelligence may not be the optimal solution regardless of the workload. However, in emergency scenarios and similar cases, the combination of AI, along with prior knowledge and training, can prove beneficial.

From the correlation analysis in \cref{sec:corr_anal}, it was generally observed that:

\[
\text{Rating of support} \propto \frac{1}{\text{Task load}} \propto \text{Situational awareness}
\]  

Expressing that as participants perceive greater support from various components, their task load decreases, and situational awareness improves. Hence, for optimal results, the decision support system should be universally applicable, benefiting each individual. Additionally, providing operators with thorough training on such an interface is likely to enhance overall performance and situational awareness while reducing task load. Furthermore, it can be observed that the process, alarm, and HMI logs are correlated with situational awareness, task load, cognitive load, attention dynamics, and physiological responses. Therefore, these correlations can help base future research on identifying such states using the Hidden Markov Model (HMM) on the real-time data and its implementation in the real world. Further in-depth study is needed to interpret the hidden states predicted by the HMM and its relation to the physiological state of the operator using ground truth data from various sources such as an eye tracker, smartwatch, questionnaires, and an electroencephalogram (EEG). 

\section{Limitations and Future Work}
This study focuses on a specific simulated chemical plant. Future studies will aim to replicate the experiment in different industrial settings to assess the broader applicability of the AI-based decision support system.

Moreover, the study's participant pool was relatively small as they were non-expert students with limited knowledge of chemical engineering and the process control room. To enhance the robustness of the findings, including a larger and more diverse group of participants and real-world operators in future research is imperative.

In future research, the proposed framework will be employed with extended experimental study including human participants. Through which the effectiveness of human-state prediction can be evaluated. Furthermore, the possibility of creating a human digital twin will be explored to be able to run the entire experiment in simulation without the need of real human participants and the digital twin will be used as their replacement to be able to have more flexibility in terms of experimental validations. Furthermore, in current research only human failure was predicted, however, the correlation between that failure and its association with task load and situational awareness is yet to be explored.

The integration of real-time feedback mechanisms to continuously adapt and improve the AI system's performance based on operator interactions and feedback is aimed for further exploration. Conducting comparative studies to evaluate the AI system's performance against traditional decision support systems or other AI-based approaches in similar industrial contexts will further validate the proposed framework.

Moreover, current research does not investigate the ethical and social implications of integrating AI systems into safety-critical industries, including considerations of trust, accountability, and transparency that would be the request for future research.

\section{Conclusion}
This study introduces an AI-based recommendation system using dynamic influence diagrams and reinforcement learning to tackle information overload in complex industrial environments, with a focus on chemical process control rooms. Preliminary results indicate the system's potential to reduce operator workload and enhance situational awareness, especially in situations of information overload and for less experienced operators.

Feedback from NASA Task Load Index (TLX) and Situation Awareness Rating Technique (SART) questionnaires, along with eye-tracking data, suggests a decrease in perceived workload and increased situational awareness when the recommendation system is active. Additionally, a reduction in operators' heart rates while using the system implies a potential reduction in the stress associated with managing process deviations.

While these results are promising, further research with larger participant samples is needed to confirm these findings and optimize the system for broader application in real-world industrial settings. This research contributes to the advancement of AI decision-support tools in safety-critical industries, paving the way for improved process safety and more efficient decision-making.

\section*{Acknowledgments}
This publication is the result of the research done along the Collaborative Intelligence for Safety-Critical systems (CISC) project that has received funding from the European Union’s Horizon 2020 Research and Innovation Programme under the Marie Skłodowska-Curie grant agreement no. 955901.

\bibliography{SRLA_DSS.bib}
\bibliographystyle{ieeetr}

{\appendix[Process, Alarm, and HMI Logs Description]}

\setcounter{table}{0}

\begin{longtblr}[
  \label = {tab:logs_descr},
]{
  colspec = {|c||p{4.5cm}|},
  rowhead = 1,
  rowfoot = 1,
  row{1} = {font=\bfseries},
  hline{1} = {0.8pt}, 
  hline{2,Z} = {0.8pt}, 
  vline{1,Z} = {0.8pt}, 
}
Var Name & Description \\
LSERB\_1 & Level \\
All2\_XX & Alarm activation \\
sAll\_XX & Alarm sound silence \\
PSERB\_1 & Pressure \\
Human & Control adjusted by human \\
Nitsel\_1 & Nitrogen system selector \\
AckAll & Alarm action \\
SRLA\_vs\_Human & Error between the suggestion and chosen control \\
SRLA & Control suggested by SRLA \\
MCReatTempOld\_1 & Manual Reactor cooling value \\
MmanNit\_1 & Manual primary flow value \\
LinasO\_1 & Water IN flow \\
MVAPrec3\_1 & Flow Steam REC3 \\
MAssWatO\_1 & Manual Absorber water selector \\
intop\_1 & Open interface \\
REC3WMO\_1 & Manual REC3 selector \\
MWpopOld\_1 & Manual Pump power value \\
MpumpOld\_1 & Manual Pump selector \\
intop\_5 & Open interface \\
FN2serb1O\_1 & Primary Nitrogen Flow \\
MCoolreatOld\_1 & Manual Reactor Cooling selector \\
intop\_2 & Open interface \\
MliqbolO\_1 & Methanol Mass in the Boiler \\
MNitsel\_1 & Manual Primary nitrogen selector \\
MAssWatFlowO\_1 & Manual Absorber water value \\
\end{longtblr}


\newpage

\vfill

\end{document}